\documentclass[10pt,twocolumn,letterpaper]{article}

\newif\ifreview 
\newif\ifarxiv \newcommand{\arxiv}{\arxivtrue}
\newif\ifcamera 

\arxiv

\ifcamera \usepackage{cvpr} \fi             
\ifreview \usepackage[review]{cvpr} \fi      
\ifarxiv \usepackage[pagenumbers]{cvpr} \fi 

\usepackage{pifont}
\definecolor{tabfirst}{rgb}{1, 0.7, 0.7}
\definecolor{tabsecond}{rgb}{1, 0.85, 0.7}
\definecolor{tabthird}{rgb}{1, 1, 0.7}
\usepackage{makecell}
\usepackage{animate}
\usepackage{graphicx}	
\usepackage{amsmath}	
\usepackage{amssymb}	
\usepackage{amsthm}
\usepackage{booktabs}
\usepackage{cuted}
\usepackage{times}
\usepackage{epsfig}
\usepackage{caption}
\usepackage{float}
\usepackage{placeins}
\usepackage{color, colortbl}
\usepackage{stfloats}
\usepackage{enumitem}
\usepackage{tabularx}
\usepackage{xstring}
\usepackage{multirow}
\usepackage{xspace}
\usepackage{url}
\usepackage{subcaption}
\usepackage[hang,flushmargin]{footmisc}
\usepackage{kotex}
\usepackage{arydshln}
\usepackage{adjustbox}
\usepackage{wrapfig}
\usepackage{algorithm,algorithmicx,algpseudocode}
\usepackage{listings}
\usepackage{bm}

\newlength\paramargin
\newlength\abovetabcapmargin
\newlength\belowtabcapmargin
\newlength\abovefigcapmargin
\newlength\belowfigcapmargin
\newlength\aboveeqmargin
\newlength\beloweqmargin

\usepackage{amsmath}

\def\kb{{\mathbf{k}}}

\def\ob{{\mathbf{o}}}

\def\q{{\mathbf{q}}}

\def\vb{{\mathbf{v}}}

\def\x{{\mathbf{x}}}

\def\z{{\mathbf{z}}}

\def\A{{\mathbf{A}}}
\def\R{{\mathbf{R}}}

\def\phib{{\bm{\phi}}}

\usepackage{thmtools,thm-restate}

\def\Rd{{\mathbb{R}}}

\definecolor{cvprblue}{rgb}{0.21,0.49,0.74}
\usepackage[pagebackref,breaklinks,colorlinks,allcolors=cvprblue]{hyperref}

\def\paperID{12409}
\def\confName{CVPR}
\def\confYear{2026}

\title{ReDirector: Creating Any-Length Video Retakes with Rotary Camera Encoding}

\author{
    Byeongjun Park\textsuperscript{1}  \qquad
    Byung-Hoon Kim\textsuperscript{1,2} \qquad
    Hyungjin Chung\textsuperscript{1}\thanks{Corresponding authors} \qquad
    Jong Chul Ye\textsuperscript{3}\footnotemark[2] \vspace{0.5em} \\
    \textsuperscript{1} EverEx \qquad
    \textsuperscript{2} Yonsei University \qquad
    \textsuperscript{3} KAIST
}

\begin{document}
\maketitle
\begin{strip}
    \centering
    \ifreview \vspace{-1.5em} \else \vspace{-3.5em} \fi
    \resizebox{\textwidth}{!}{
    \includegraphics[width=\linewidth]{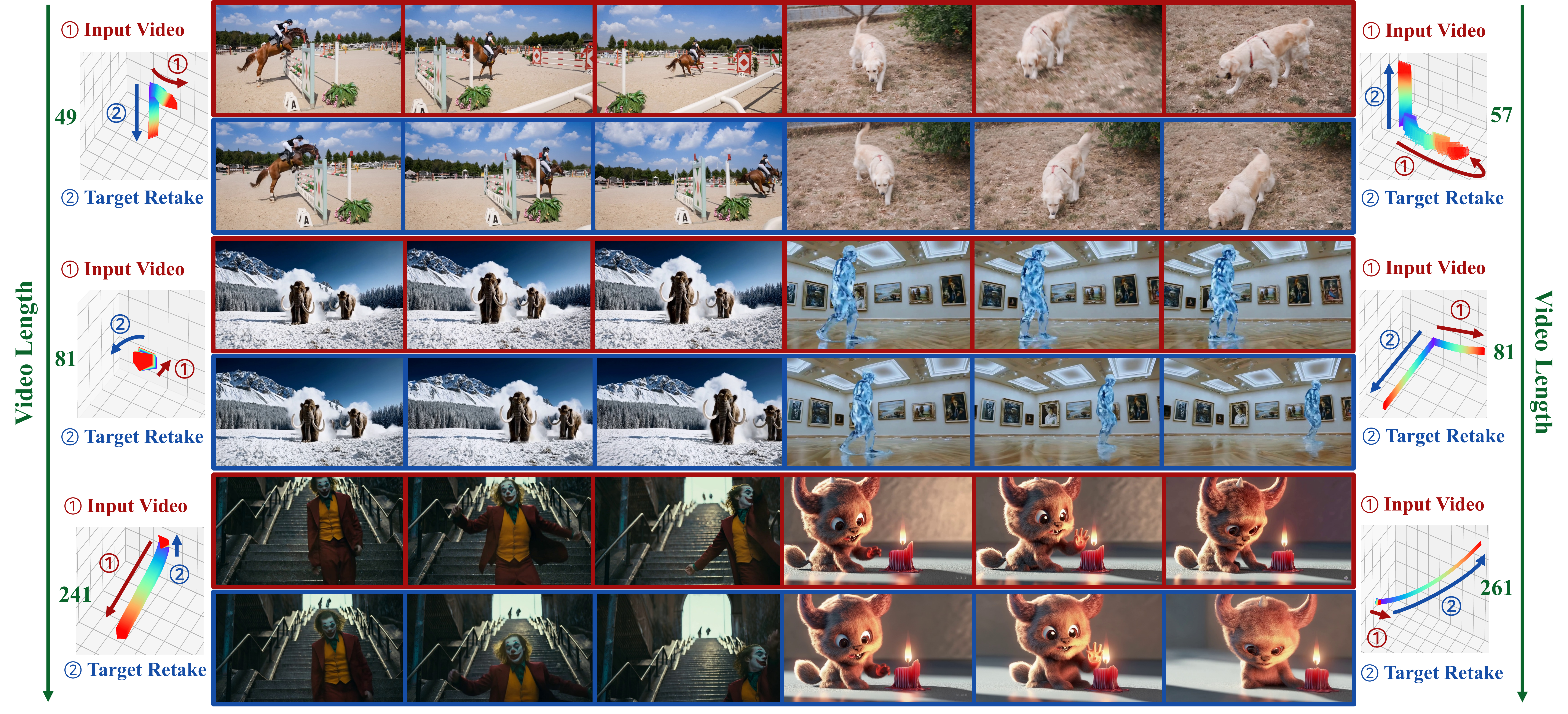}
    }
    \vspace{\abovefigcapmargin}
    \vspace{-0.45cm}
    \captionof{figure}{\textbf{ReDirector for video retake generation.} Given any-length input videos (①), ReDirector generates realistic retakes (②) along the target camera trajectories even with dynamic camera motion in the input video. ReDirector is capable of accurately localizing dynamic objects while preserving static backgrounds throughout the sequence, leading to multi-view consistent retakes spanning hundreds of frames.}
    \label{fig:teasure}
    \vspace{-0.1cm}
\end{strip}

\begin{abstract}
We present ReDirector, a novel camera-controlled video retake generation method for dynamically captured variable-length videos. In particular, we rectify a common misuse of RoPE in previous works by aligning the spatiotemporal positions of the input video and the target retake. Moreover, we introduce Rotary Camera Encoding (RoCE), a camera-conditioned RoPE phase shift that captures and integrates multi-view relationships within and across the input and target videos. By integrating camera conditions into RoPE, our method generalizes to out-of-distribution camera trajectories and video lengths, yielding improved dynamic object localization and static background preservation. Extensive experiments further demonstrate significant improvements in camera controllability, geometric consistency, and video quality across various trajectories and lengths.
\end{abstract}    
\section{Introduction}
\label{sec:intro}

\ifreview
\else
    \begingroup
        \renewcommand\thefootnote{}
        \footnotetext{Project page: \href{https://byeongjun-park.github.io/ReDirector/}{https://byeongjun-park.github.io/ReDirector/}}
        \addtocounter{footnote}{0}
    \endgroup
\fi

Camera-controlled video generation has attracted broad interest because it enables precise camera control over scene progression and visual dynamics, such as in filmmaking and virtual production~\cite{ma2025controllable}. In this context, video retake generation has made significant progress by using video-to-video generative models to redirect the camera trajectories of existing videos, enabling video captures at physically impractical viewpoints and stabilizing shaky footage, thereby reducing production costs and providing immersion~\cite{Bai_2025_ICCV, luo2025camclonemaster}.

Recent warping-based frameworks~\cite{Yu_2025_ICCV, Jeong_2025_ICCV, zhang2024recapture, xiao2025trajectory, chen2025cognvs, lu2025see4d} achieve high-quality video retake generation by leveraging video generative models~\cite{yang2024cogvideox, wan2025wan, kong2024hunyuanvideo} with two conditions, an input video and target camera trajectories. These frameworks resort to explicit geometric transformations (\ie, per-frame warping) using video depth estimation models~\cite{hu2025depthcrafter, chen2025video} and point tracking models~\cite{xiao2024spatialtracker, karaev2024cotracker}, after which the video generative model refines and inpaints warped frames. However, the external geometry estimators frequently degrade under dynamic camera motion and complex structures, allowing warping artifacts to propagate directly into the video generative model, where they are inadequately addressed.

To address these limitations, another line of works~\cite{van2024generative, Bai_2025_ICCV} extend camera-controlled text-/image-to-video generation methods~\cite{he2024cameractrl, bahmani2025ac3d, wang2024motionctrl, Go_2025_CVPR} that utilize camera parameters to {\em implicitly} synthesize video frames without a warping process. In practice, the representations of the input video and the target camera trajectory are combined with those of the target video via concatenation or addition along channel or token axes, fine-tuning video generative models to internalize the multi-view geometry between input and target videos. 
While such designs may be sufficient for general conditioning in video-to-video generation tasks~\cite{Jiang_2025_ICCV}, critical drawbacks remain. Namely, the existing methods, including the previous warping-based approaches, assume fixed-length inputs with minimal camera motion, and hence degrade quickly outside these regimes.
A key open challenge, therefore, is to handle variable-length inputs with dynamic camera motion by seamlessly integrating the two control signals to encode multi-view relationships across both videos and along their camera trajectories.

In this work, we introduce ReDirector, a novel camera-controlled video retake generation method for dynamically captured variable-length input videos, as shown in~\cref{fig:teasure}. We leverage Rotary Position Embedding (RoPE)~\cite{su2024roformer}, a relative positional encoding known to generalize across sequence lengths~\cite{barbero2024round, heo2024rotary, yang2024cogvideox, wan2025wan}. Our key idea is to employ the same RoPE to both input and target videos to encode the length-agnostic relative positions, then inject camera conditions as physically grounded positional signals that encode multi-view geometry within and across the two videos. This design views shared RoPE indices as tightly aligned spatiotemporal positions, and the camera condition encourages discrimination between the input and target videos at the corresponding positions.
More concretely, we present Rotary Camera Encoding (RoCE), which integrates camera pose parameters into RoPE through a camera-conditioned phase shift. RoCE is initialized with zero phase, disabling camera conditioning at the start, and gradually learns nonzero phases during fine-tuning, ensuring stable incorporation of the camera signal. Applied across multiple RoPE frequencies with head-specific phase shifts, RoCE enhances geometric consistency by strengthening attention at the same physical positions across views, enabling multi-view-consistent retakes even over long temporal gaps.

In parallel, amid growing interest in camera conditions as positional encoding~\cite{park2024bridging, bai2025positional}, recent geometry-aware attention methods~\cite{miyato2024gta, kong2024eschernet, li2025cameras} use explicit geometric transforms directly into attention layers, which are trained from scratch on static scenes. Given their design, these methods are ill-suited for fine-tuning video generative models. Instead, we replace explicit geometric transforms with our camera-conditioned phase shifts, which apply a $SO(2)$ phase shift to values before attention weighting and the inverse phase shift after the value aggregation. This preserves geometry awareness in the attention layers, remains compatible with fine-tuning of video generative models, and improves the disentanglement of dynamic objects from static backgrounds, resulting in physically plausible video retakes.

Through extensive experiments on video examples from the DAVIS dataset~\cite{pont20172017} and trajectories from ReCamMaster~\cite{Bai_2025_ICCV}, our method consistently outperforms previous methods in camera control, geometric consistency, and visual quality across diverse target trajectories and video lengths. Evaluations on the iPhone dataset~\cite{gao2022dynamic} further validate the strong generalizability of our method to out-of-distribution camera trajectories, video lengths, and resolutions.
\section{Related Works}
\label{sec:related_works}

\subsection{Video Retake Generation}

Recent progress in diffusion transformers~\cite{peebles2023scalable, park2023denoising, park2024switch, ham2025diffusion} and video generative models~\cite{kong2024hunyuanvideo, wan2025wan, yang2024cogvideox, genmo2024mochi} enables the incoporation of camera control signals into text-to-video generation~\cite{Go_2025_CVPR, Go_2025_ICCV} and image-to-video generation~\cite{wang2024motionctrl, he2024cameractrl, bahmani2025ac3d, xu2024camco, 11125650, hou2025trainingfree, Muller_2024_CVPR, bahmani2025vd3d, Ren_2025_CVPR, seo2024genwarp}. These methods exploit camera poses either through explicit geometric transformation (\ie, per-frame warping) or by embedding Plücker rays~\cite{plucker1865xvii} to encourage the model to implicitly learn the multi-view geometry. Building on these advances, video retake generation, which repurposes the input video by redirecting its camera trajectory using video-to-video generative models, has been widely explored in both explicit and implicit forms.

\vspace{\paramargin}
\paragraph{Explicit geometric transformation.}
The dominant approach for video retake generation is warping-based methods. Most pipelines first estimate scene geometry with video depth models~\cite{hu2025depthcrafter, chen2025video} and backproject into 3D point clouds, then reproject each input frame to the target camera trajectories to synthesize a geometrically aligned proxy of the retake. A video generative model subsequently refines and inpaints these proxies using warped frames and the input video, either via fine-tuning~\cite{Yu_2025_ICCV, Jeong_2025_ICCV, zhang2024recapture, xiao2025trajectory, chen2025cognvs, lu2025see4d} or in a training-free manner~\cite{park2025zero4d, you2025nvssolver, yesiltepe2025dynamic}. While effective, these methods depend heavily on accurate video depths, and scale alignment between the estimated input geometry and the target camera trajectory typically relies on hand-crafted heuristics. Moreover, under dynamic camera motion and complex scene structure, warping artifacts from external geometric estimators propagate directly into the generated retakes, substantially degrading quality. The per-frame warping process further undermines the disentanglement of dynamic objects from static backgrounds and prevents encoding of a spatio-temporally aligned scene representation.

\vspace{\paramargin}
\paragraph{Implicit geometric transformation.}

Recently, implicit approaches~\cite{van2024generative, Bai_2025_ICCV} construct large-scale synthetic datasets of input video, target camera trajectory, and target retake triplets, using rendering pipelines such as Kubric~\cite{greff2022kubric} and Unreal Engine 5~\cite{unrealengine5}. 
Instead of explicit warping, these methods incorporate both control signals directly into the video generative models by conditioning on camera extrinsic parameters and VAE-encoded input latent, relying on dataset supervision. This encourages the models to internalize multi-view geometry and to inpaint occlusions without hand-crafted preprocessing. Nevertheless, performance remains sensitive to the training data and can degrade under out-of-distribution trajectories and their video lengths. We extend these implicit methods by seamlessly integrating the two control signals to exhibit robust generalization to out-of-distribution trajectories and sequence lengths.

\subsection{Camera encoding in vision transformers}

Camera parameters provide physically grounded positional information that token indices do not capture. Most prior works~\cite{gao2024cat3d, wu2025cat4d, tang2024lgm, jin2024lvsm, woo2024harmonyview} therefore adopt pixel-aligned (token-level) encodings that attach per-pixel geometry to each token, such as spherical angles or Plücker rays. By concatenating these camera encodings with tokens, the model is conditioned on both intrinsics and extrinsics and can learn fine-grained, token-level multi-view relationships. However, this encoding requires specifying a reference frame and a global world coordinate system, which introduces frame dependence and can hinder generalization, even with the scene scale normalization~\cite{kulhanek2022viewformer, Go_2025_CVPR}. 

In parallel, frame-aligned (camera-level) encodings utilize the camera pose of each frame by leveraging the characteristics of $SE(3)$ and $SO(3)$, typically encoding relative transformations into the attention scores. These signals modulate attention to inject geometry awareness and tend to generalize well across variable sequence lengths; they do not require a consistent global frame and have been shown to improve the novel view synthesis performance~\cite{miyato2024gta, kong2024eschernet}. Nonetheless, camera-level encoding lacks per-pixel granularity, which undermines fine multi-view correspondence and visibility reasoning (\eg, occlusions and thin structures).
Concurrently, PRoPE~\cite{li2025cameras} combines both camera encodings and trains a multi-view transformer from scratch on static scenes. In contrast, we propose a novel camera encoding that integrates with RoPE as phase shifts and is tailored for fine-tuning video generative models on dynamic scenes.
\section{Preliminary}
\label{sec:prelim}

\paragraph{Rotary Position Embedding (RoPE).}
Recent video diffusion models~\cite{yang2024cogvideox, kong2024hunyuanvideo, genmo2024mochi, wan2025wan} are trained to generate arbitrary-length video sequences and multiple resolutions, despite not being explicitly trained for these settings. This generalizability is mostly derived from Rotary Position Embedding (RoPE)~\cite{su2024roformer} used in their attention layers. After the VAE encoder and patchify layers, each attention layer takes latent tokens $\x \in \mathbb{R}^{N \times d}$, where $N=fhw$ is the total number of tokens and $f, h, w$ are the compressed resolution of frame, height, and width, respectively.
Linear projections are then applied to obtain the query and the key vectors, $\q,\kb \in \Rd^{N \times d_{head}}$, where $d_{head}$ is the dimension of a single head of the multi-head attention (MHA). RoPE treats the vectors as a set of complex vectors by grouping elements, \ie, the even-indexed and subsequent odd-indexed channels are the real and 
imaginary parts, respectively
\begin{align}
    \q,\kb \mapsto \bar{\q}, \bar{\kb} \in \mathbb{C}^{N \times (d_{head}/2)}.
\end{align}
Complex rotation can then be applied with a unitary rotation matrix $\mathbf{R}$, which can be effectively implemented via
\begin{equation}
    \bar{\q}' = \bar{\q} \circ \mathbf{R}, \bar{\mathbf{k}}' = \bar{\mathbf{k}} \circ \mathbf{R},
\end{equation}
where $\mathbf{R} \in \mathbb{C}^{N\times (d_{head}/2)}$ is the matrix that implements the rotation with Hadamard product $\circ$, where each element $\mathbf{R}(n, c) = e^{i\theta_c n}$ is defined following Euler's formula with exponentially decaying multiple frequency $\theta_c$ as:
\begin{equation}
    \theta_{c} = 10000^{-\frac{c-1}{d_{head} / 2}}, \text{where}~c \in \{1, 2, \dots, d_{head} / 2\}.
\end{equation}
For $(n, m)$-th query and key vectors, this modulates each element in the attention matrix $\textbf{A}' \in \mathbb{R}^{N \times N} $as follows:
\begin{equation}
    \mathbf{A}'{(n, m)} = \text{Re}[\bar{\q}_{n}'\bar{\mathbf{k}}_{m}'^{*}] = \text{Re}[\bar{\q}_{n}(\bar{\mathbf{k}}_{m}^{*} \circ e^{i\theta_c (n-m)})],
\end{equation}
where $\text{Re}[\cdot]$ extracts the real part and $^{*}$ denotes the complex conjugates. Therefore, relative positions are encoded as integer multiples of per-channel phase offsets determined by the token indices, and multiple frequency allows the attention matrix to cover a wide range of relative positions.

\vspace{\paramargin}
\paragraph{3D RoPE.} In video diffusion models, spatiotemporal relative positions are commonly encoded using 3D RoPE, which consists of three complex rotation matrices for the frame, height, and width axes as follows:
\begin{equation}
    \bar{\mathbf{R}}_{f} = \mathbf{R}_f \otimes \mathbf{1}_h \otimes \mathbf{1}_w \in \mathbb{C}^{f \times h \times w \times (d_{head} / 6)},
\end{equation}
\begin{equation}
    \bar{\mathbf{R}}_{h} = \mathbf{1}_f \otimes \mathbf{R}_h \otimes \mathbf{1}_w \in \mathbb{C}^{f \times h \times w \times (d_{head} / 6)},
\end{equation}
\begin{equation}
    \bar{\mathbf{R}}_{w} = \mathbf{1}_f \otimes \mathbf{1}_h \otimes  \mathbf{R}_w \in \mathbb{C}^{f \times h \times w \times (d_{head} / 6)},
\end{equation}
where $\otimes$ is Kronecker product and $\mathbf{1}_l \in \mathbb{C}^{l \times 1}$ denotes the all-ones matrix and $\mathbf{R}_l \in \mathbb{C}^{l \times (d_{head} / 6)}$ is the complex rotation matrix for each axis $l \in \{f, h, w\}$.
Then, we reshape each $\bar{\mathbf{R}}_{l}$ to $\tilde{\mathbf{R}}_l \in \mathbb{C}^{N \times (d_{head}/2)}$ and apply channel-wise concatenation to create 3D RoPE $\R$ as follows:
\begin{equation}
    \R = \tilde{\mathbf{R}}_f \oplus \tilde{\mathbf{R}}_h \oplus \tilde{\mathbf{R}}_w \in \mathbb{C}^{N \times (d_{head}/2)}.
\end{equation}
\section{Methods}
\label{sec:methods}

\begin{figure*}[t]
    \centering
    \includegraphics[width=\linewidth]{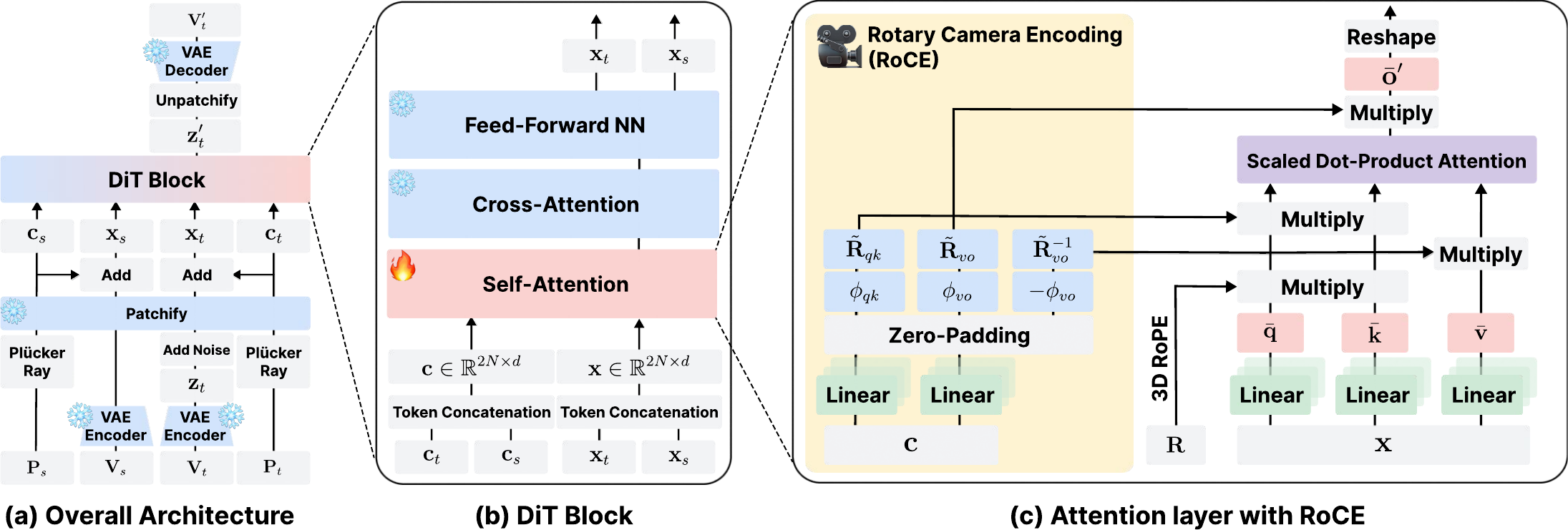}
    \caption{\textbf{Overview of ReDirector.} (a) ReDirector is fine-tuned on Wan-I2V-CamCtrl~\cite{wan2025wan}, which incorporates camera control signals into image-to-video generation. Our goal is to reconstruct a video retake $\mathbf{V'}_t$ conditioned on target camera trajectories $\mathbf{P}_t$, input video $\mathbf{V}_s$, and its poses $\mathbf{P}_s$. (b) Following ReCamMaster~\cite{Bai_2025_ICCV}, we train only self-attention layers while keeping the remaining modules frozen. (c) We insert RoCE into self-attention layers, whose outputs are used as camera-conditioned RoPE phase shifts. First, $\phib_{\q\vb}$ is applied to queries and keys, providing physically grounded rotary position encoding. Second, $\phib_{\vb\ob}$ modulates the value path by applying $-\phib_{\vb\ob}$ before attention weighting and $+\phib_{\vb\ob}$ after value aggregation, enabling geometry-aware attention. For clarity, text prompts in cross-attention are omitted.}
    \vspace{\belowfigcapmargin}
    \label{fig:overview}
\end{figure*}

In this section, we present ReDirector, a camera-controlled video retake generation method for variable-length inputs captured with dynamic camera motion. In~\cref{sec:training}, we explain how we rectify common misuse of RoPE and camera encodings in prior work, enabling arbitrary-length input videos to be retaken with precise camera control. Then, in~\cref{sec:roce}, we introduce RoCE, a physically grounded positional encoding that integrates into RoPE, further improving geometric consistency and achieving better disentanglement of dynamic objects from static backgrounds.

\subsection{Video-to-Video Generative Model}
\label{sec:training}

\paragraph{Architectural design.}
Given an input video $\mathbf{V}_{s}$ and target camera trajectories $\mathbf{P}_{t}$, our objective is to generate a video retake $\mathbf{V}_{t}$ that follows the target trajectories. To this end, we start from the pretrained camera-controlled image-to-video generative model~\cite{wan2025wan} and fine-tune it into a video-to-video generative model.
We specifically revisit and correct common misuse in prior work regarding RoPE and camera encodings. Existing methods make only limited use of positional encodings for the input video, typically applying absolute positional encodings or using only specific axes in 3D RoPE, which forces them to assume fixed-length inputs. Moreover, unlike mainstream camera-controlled text-/image-to-video generation methods that leverage Plücker rays as implicit representations, these methods predominantly rely on explicit geometric transformations, and even implicit variants~\cite{Bai_2025_ICCV, van2024generative} often encode only target camera extrinsic parameters, resulting in coarse camera conditioning and limited token-level multi-view relationships.

In contrast, we apply a shared 3D RoPE to both the input and target videos, aligning their spatiotemporal positions and preserving length-agnostic positional encodings for the input sequence. This significantly improves generalization to longer video sequences, even when such lengths are absent from the training data. We further adopt Plücker rays as camera pose representations to employ token-level camera encodings. For the input video geometry, we use its camera trajectories $\mathbf{P}_{s}$ obtained from the training dataset, or estimate them at inference using ViPE~\cite{huang2025vipe}. By jointly conditioning on the input and target trajectories along with their video frames, the generative model is fine-tuned to represent detailed multi-view geometry under variable-length input videos. This overall setup is illustrated in~\cref{fig:overview}-(a).

\vspace{\paramargin}
\paragraph{Training loss.}
We adopt the rectified flow setting~\cite{esser2024scaling, liu2022flow} and conditional flow matching loss~\cite{lipman2022flow} to train our model $u_\theta$, following ReCamMaster~\cite{Bai_2025_ICCV} to update weights only in the self-attention layers while keeping all other components frozen, as shown in~\cref{fig:overview}-(b).
We define a noise distribution $p_1 \in \mathcal{N}(0, I)$ and a mapping between the data distribution $p_0$ (\ie, target video) and the noise distribution as an ordinary differential equation (ODE):
\begin{equation}
    \mathrm{d}\z_t = u_\theta(\z_t, t)\mathrm{d}t,
\end{equation}
\begin{equation}
    \z_t = t\z_1 + (1-t)\z_0,
\end{equation}
where $\z_0 \sim p_0$ and $\z_1 \sim p_1$, and $t \in [0, 1]$. 
Then, the conditional flow matching loss is defined as:
\begin{equation}
    \mathcal{L}_{\text{CFM}} 
    = \mathbb{E}_{t, p_0, p_1}\!\bigl[\|(\z_1 - \z_0) - u_\theta(\z_t,t)\|^2\bigr].
\end{equation}
After the training, video retakes are generated by solving the ODE from $t=1$ to $t=0$ at the inference stage.

\vspace{\paramargin}
\paragraph{Training strategy.}
To further enhance the performance, we adopt several training strategies. First, we include identity-retake pairs where the input and target videos share the same camera trajectories, $\{\mathbf{V}_s, \mathbf{P}_s\} = \{\mathbf{V}_t, \mathbf{P}_t\}$. This encourages the model to learn tighter alignment between tokens that correspond to the same 3D RoPE and RoCE. In addition, we augment the training dataset with time-reversed versions of the videos, exposing the model to a wider variety of camera trajectories and enabling retake generation from other viewpoints, even at the first frame.

\subsection{Rotary Camera Encoding}
\label{sec:roce}
\begin{figure}[t!]
    \centering
    \includegraphics[width=\linewidth]{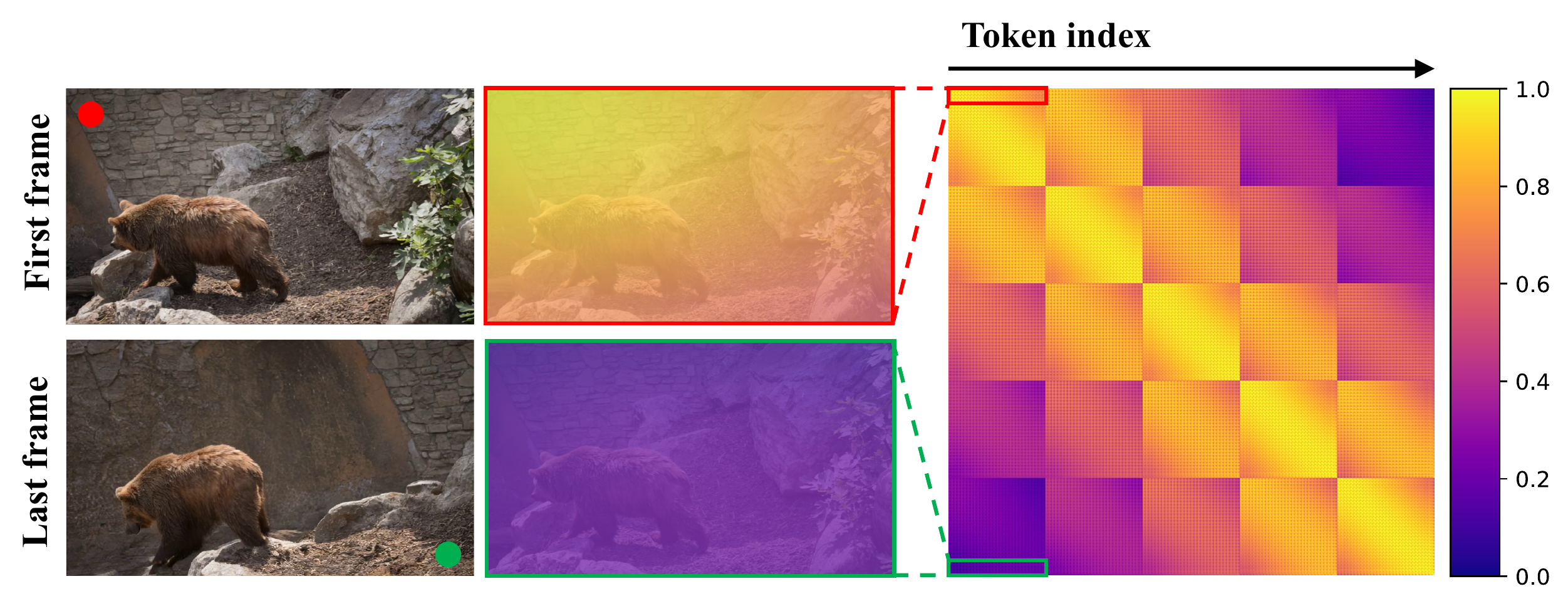}
    \vspace{-0.6cm}
    \caption{\textbf{Attention of RoCE.} We visualize the attention from the colored dot in the leftmost column to the first frame (left), and across five uniformly sampled frames (right). Within each frame, attention varies with pixel coordinates, whereas differences in relative pose have a more pronounced impact on the attention scores.}
    \vspace{\belowfigcapmargin}
    \vspace{-2mm}
    \label{fig:roce_vis}
\end{figure}

Let
$\mathbf{R} = [\mathbf{R}_{t}, \mathbf{R}_{s}] \in \mathbb{C}^{2N \times (d_{head}/2)}$
be the RoPE rotation matrix encoding the {\em spatiotemporal location} of the video for the target and the source, respectively, with $\R_t = \R_s$.
Further let
$\mathbf{c} = [\mathbf{c}_{t}, \mathbf{c}_{s}] \in \mathbb{R}^{2N \times d}$
be the Pl\"ucker ray tokens encoding the {\em camera parameters} of the target and the source. Our goal is to further encode the camera information through relative positional embeddings to the {\em spatial} part, a method we call RoCE, as shown in~\cref{fig:overview}-(c). Specifically, at each DiTBlock, we have
\begin{align}
    \phib_{\q\kb} &= [\mathbf{0}, \text{MLP}_{\q\kb}(\mathbf{c})]  \in \mathbb{R}^{2N \times (d_{head}/2)}, \\
    \phib_{\vb\ob} &= [\mathbf{0}, \text{MLP}_{\vb\ob}(\mathbf{c})]  \in \mathbb{R}^{2N \times (d_{head}/2)},
\end{align}
where $\mathbf{0} \in \mathbb{R}^{2N \times (d_{head}/6)}$ is the all-zeros matrix to prevent it from affecting the temporal positions. Similar to RoPE, we construct the learnable camera rotation matrix by grouping and applying it to the query and the key vectors
\begin{align}
    \tilde{\mathbf{R}}_{\q\kb} = e^{i\phib_{\q\kb}} \in \mathbb{C}^{2N \times (d_{head}/2)}, \\
    \tilde{\mathbf{R}}_{\vb\ob} = e^{i\phib_{\vb\ob}} \in \mathbb{C}^{2N \times (d_{head}/2)},
\end{align}
and subsequently
\begin{align}
    \bar{\q}' = \bar{\q} \circ \mathbf{R} \circ \tilde{\mathbf{R}}_{\q\kb},\,\quad \bar{\mathbf{k}}' = \bar{\mathbf{k}} \circ \mathbf{R} \circ \tilde{\mathbf{R}}_{\q\kb}.
\end{align}
For the $(n,m)$-th query and key vectors, the attention matrix now reads
\begin{equation}
    \mathbf{A}{(n, m)}' = \text{Re}[\bar{\q'}_{n}(\bar{\mathbf{k}}_{m}'^{*} \circ e^{i(\theta_c (n-m) + \phib_{\q\kb}(n, c) - \phib_{\q\kb}(m, c)})].
\end{equation}
Notably, as shown in~\cref{fig:roce_vis}, we observe that the attention matrix formed purely by the RoPE phase shifts (\ie, $\text{Re}[e^{i(\phib_{\q\kb}(n, c) - \phib_{\q\kb}(m, c)}]$) indicates that RoCE behaves similarly to RoPE within each frame, but is more sensitive to relative poses, substantially suppressing attention between distant viewpoints. This enables the model to align distant frames and consistently preserve background regions at corresponding positions, even when they are far apart in time.

\vspace{\paramargin}
\paragraph{Geometry-aware attention.}
We extend geometry-aware attention frameworks~\cite{miyato2024gta, li2025cameras, kong2024eschernet}, which treat token-wise camera encoding and geometry-aware attention independently, by explicitly coupling these components via RoPE phase shifts. We inject geometric awareness directly into the value aggregation process by leveraging $SO(2)$ characteristics of phase shifts. We apply ${\tilde{\R}_{\vb\ob}}^{-1}$\footnote{With a slight abuse of notation, $\R^{-1}$ refers to the matrix that applies the complex rotation in the negative direction.} before the attention weighting, and apply ${\tilde{\R}_{\vb\ob}}$ after the value aggregation as:
\begin{align}
    \bar{\ob}' = \left(\A' \left(\underbrace{\bar{\vb} \circ {\tilde{\R}_{\vb\ob}}^{-1}}_{\bar{\vb}'} \right)\right) \circ \tilde{\R}_{\vb\ob}.
\end{align}
This enables an efficient implementation of a {\em learnable} geometry-aware attention similar to GTA~\cite{miyato2024gta} when fine-tuning video generative models, whereas previous methods require training from scratch on static scenes. Furthermore, the model better separates dynamic objects from static backgrounds, as tokens in static regions stay multi-view consistent, while tokens on moving objects break this pattern. This yields more geometrically consistent retakes.

\section{Experimental Results}
\label{sec:exp}

\subsection{Experimental Setup}
\label{sec:exp_setup}

\begin{figure*}[t!]
    \centering
    \setlength\tabcolsep{0.2pt}
    
    \begin{tabular}{@{}c@{\,}cccccc@{}}

        \raisebox{0.02\linewidth}{\rotatebox{90}{\scriptsize Camera}} & \multicolumn{3}{c}{\adjincludegraphics[clip,width=0.158\linewidth,height=0.0911\linewidth,trim={0 0 0 0}]{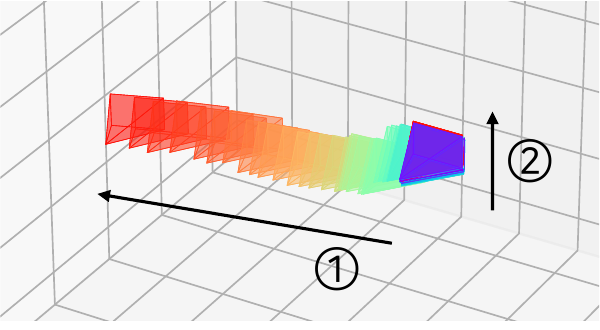}} &
        & \adjincludegraphics[clip,width=0.158\linewidth,height=0.0911\linewidth,trim={0 0 0 0}]{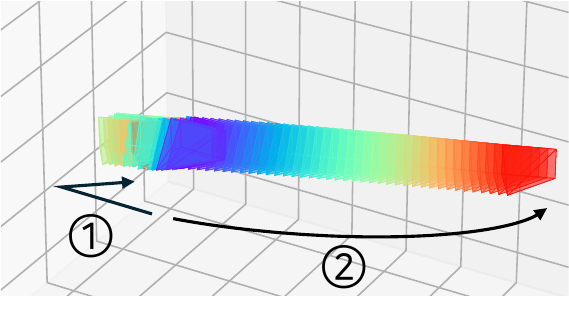} &  \\
    
        \raisebox{0.01\linewidth}{\rotatebox{90}{\scriptsize Input Video}} &
        \adjincludegraphics[clip,width=0.158\linewidth,trim={0 0 0 0}]{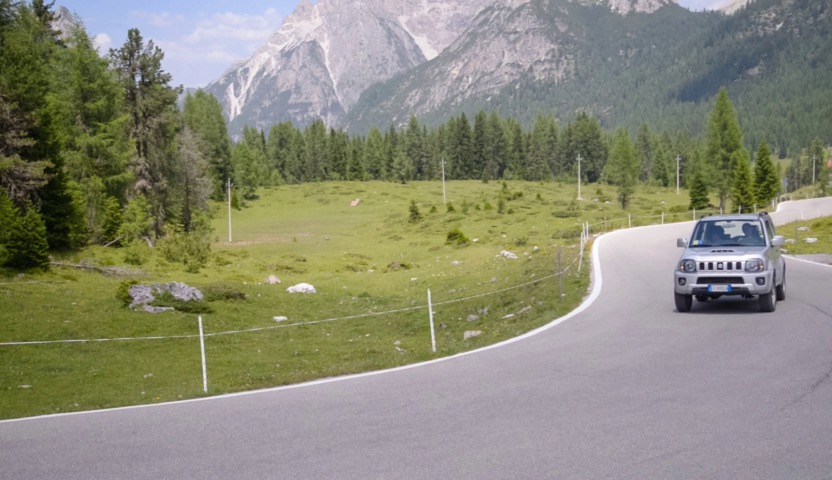} &
        \adjincludegraphics[clip,width=0.158\linewidth,trim={0 0 0 0}]{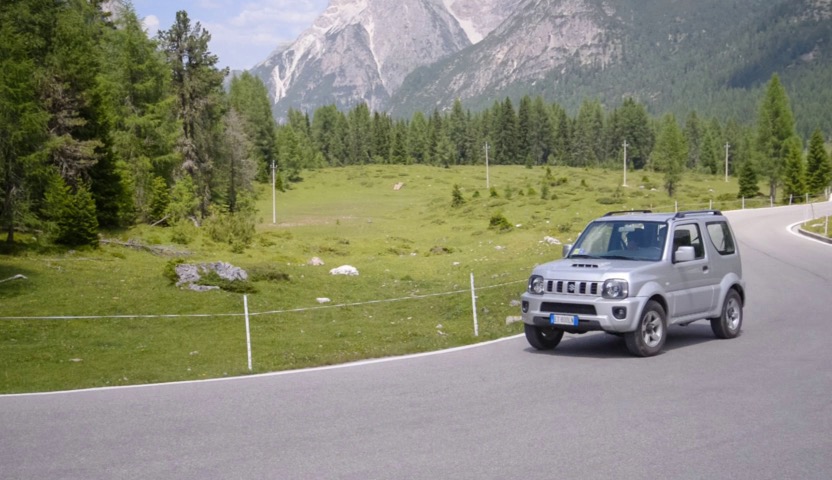} &
        \adjincludegraphics[clip,width=0.158\linewidth,trim={0 0 0 0}]{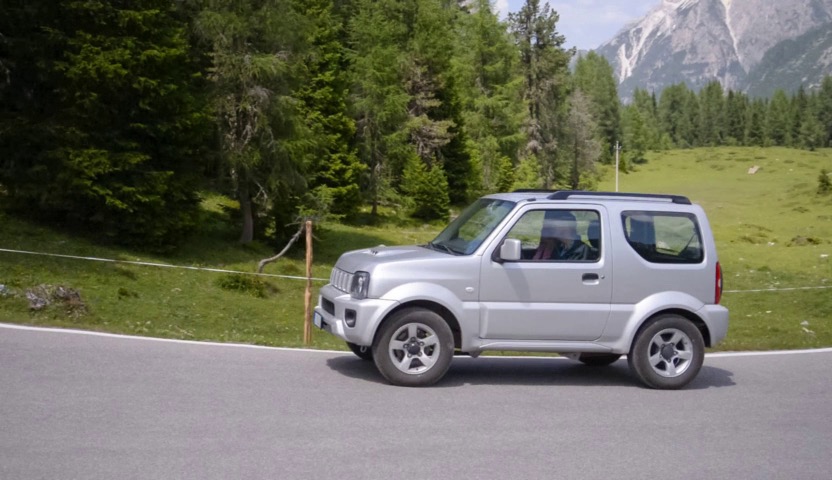} &
        \adjincludegraphics[clip,width=0.158\linewidth,trim={0 0 0 0}]{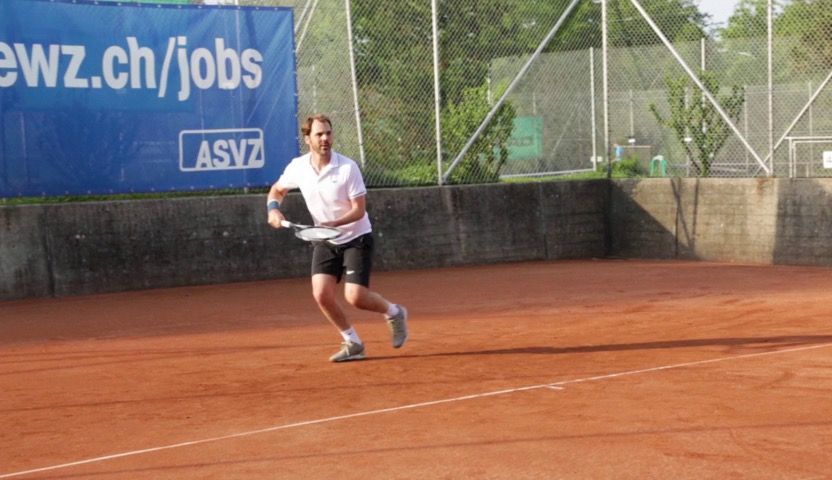} &
        \adjincludegraphics[clip,width=0.158\linewidth,trim={0 0 0 0}]{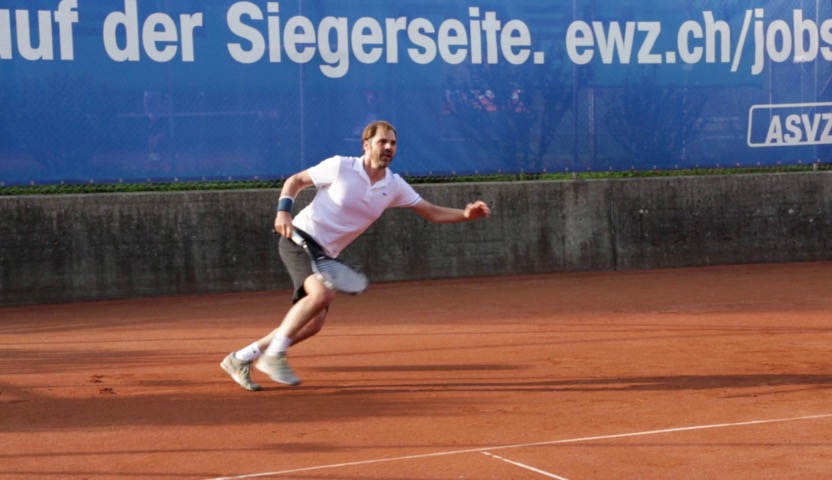} &
        \adjincludegraphics[clip,width=0.158\linewidth,trim={0 0 0 0}]{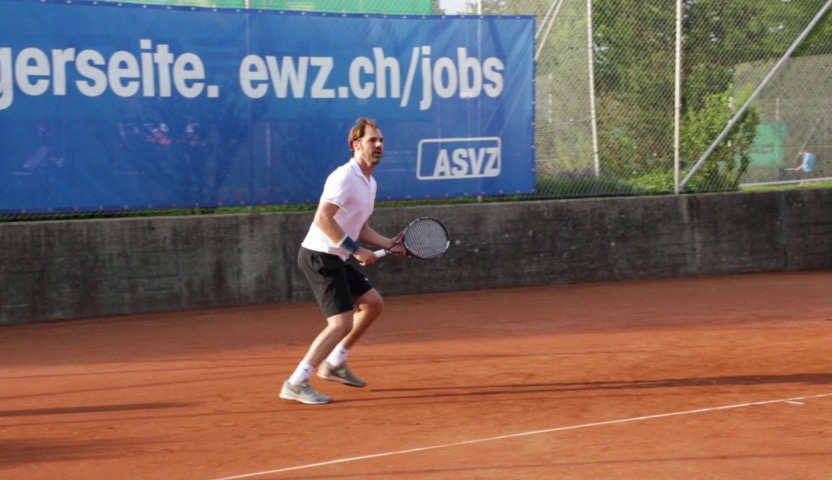} \\

        \raisebox{0.03\linewidth}{\rotatebox{90}{\scriptsize GCD}} &
        \adjincludegraphics[clip,width=0.158\linewidth, height=0.0911\linewidth, trim={0 0 0 0}]{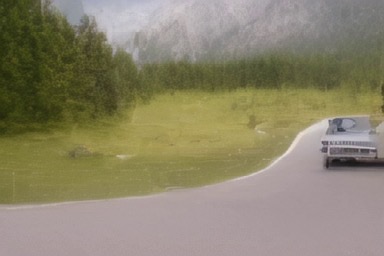} &
        \adjincludegraphics[clip,width=0.158\linewidth, height=0.0911\linewidth, trim={0 0 0 0}]{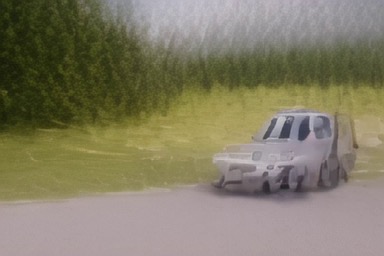} &
        \adjincludegraphics[clip,width=0.158\linewidth, height=0.0911\linewidth, trim={0 0 0 0}]{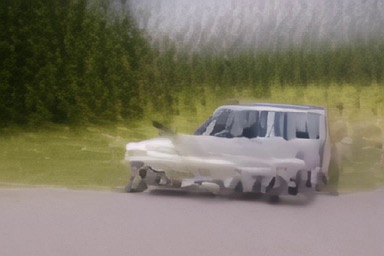} &
        \adjincludegraphics[clip,width=0.158\linewidth, height=0.0911\linewidth, trim={0 0 0 0}]{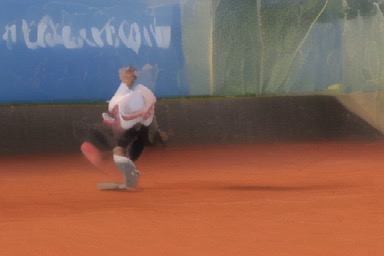} &
        \adjincludegraphics[clip,width=0.158\linewidth, height=0.0911\linewidth, trim={0 0 0 0}]{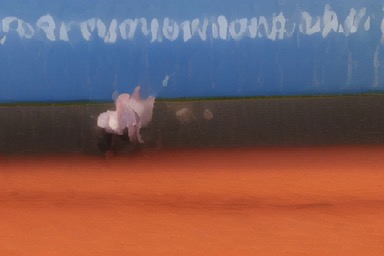} &
        \adjincludegraphics[clip,width=0.158\linewidth, height=0.0911\linewidth, trim={0 0 0 0}]{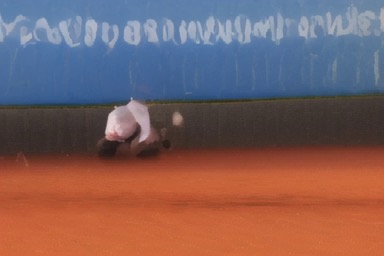} \\

        \raisebox{0.005\linewidth}{\rotatebox{90}{\scriptsize ReCamMaster}} &
        \adjincludegraphics[clip,width=0.158\linewidth,trim={0 0 0 0}]{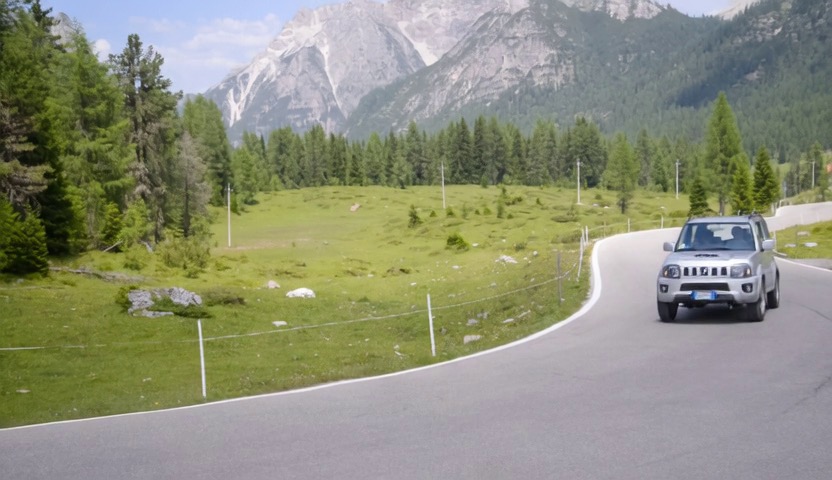} &
        \adjincludegraphics[clip,width=0.158\linewidth,trim={0 0 0 0}]{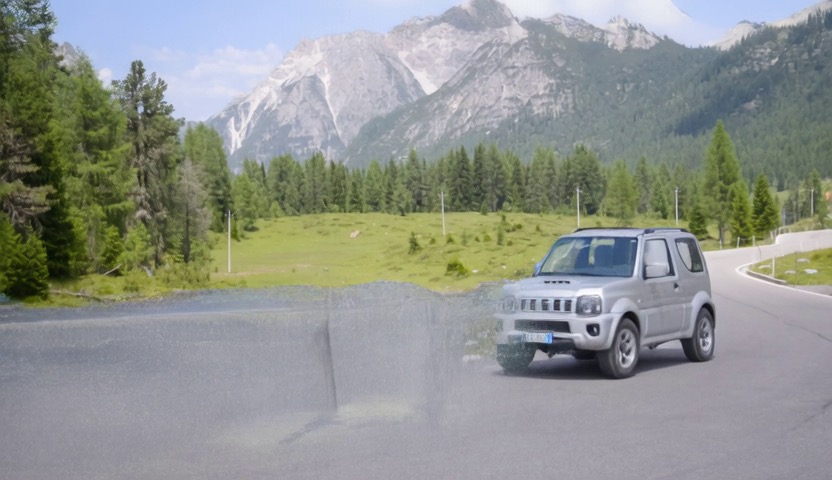} &
        \adjincludegraphics[clip,width=0.158\linewidth,trim={0 0 0 0}]{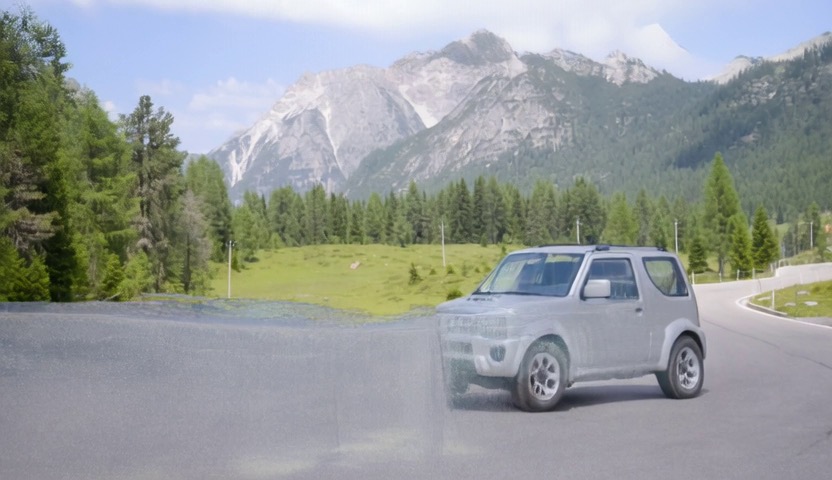} &
        \adjincludegraphics[clip,width=0.158\linewidth,trim={0 0 0 0}]{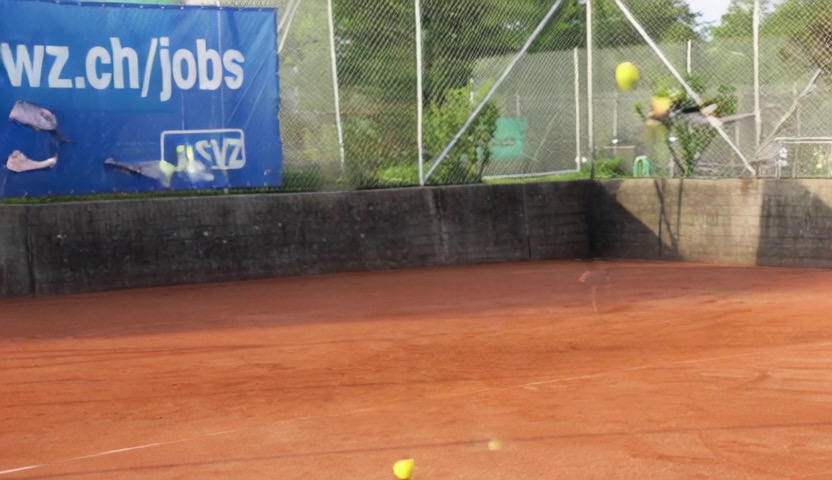} &
        \adjincludegraphics[clip,width=0.158\linewidth,trim={0 0 0 0}]{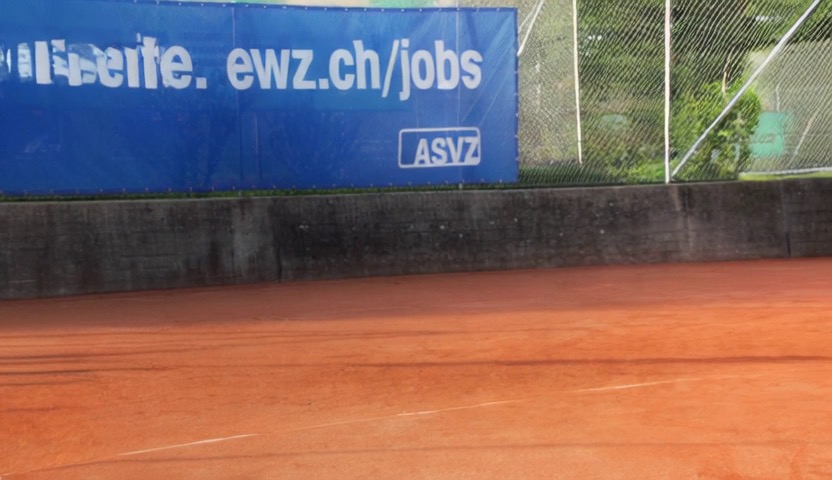} &
        \adjincludegraphics[clip,width=0.158\linewidth,trim={0 0 0 0}]{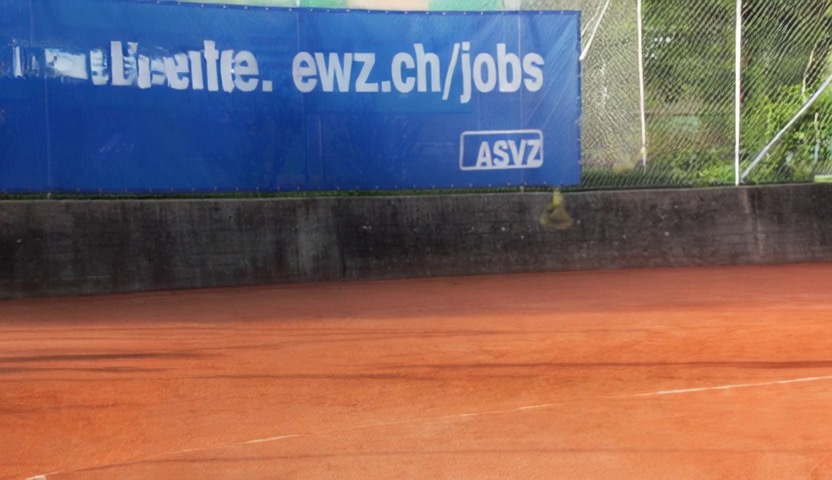} \\

        \raisebox{0.015\linewidth}{\rotatebox{90}{\makecell[c]{\scriptsize Trajectory\\[-4pt]\scriptsize Crafter}}} &
        \adjincludegraphics[clip,width=0.158\linewidth,trim={0 0 0 0}]{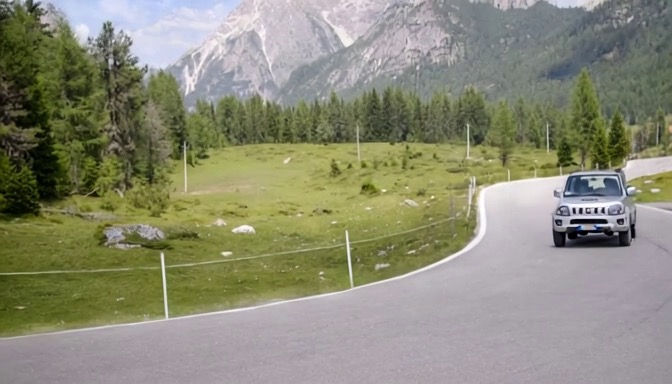} &
        \adjincludegraphics[clip,width=0.158\linewidth,trim={0 0 0 0}]{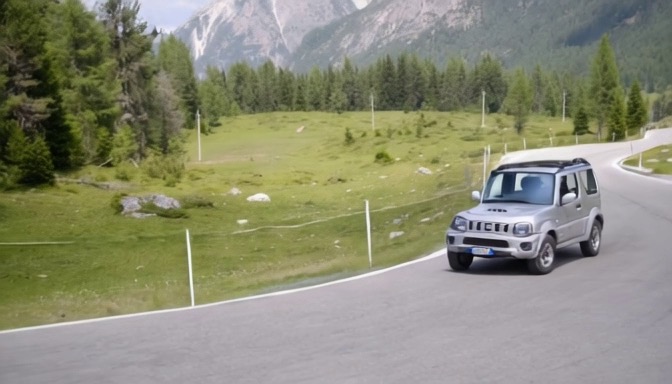} &
        \adjincludegraphics[clip,width=0.158\linewidth,trim={0 0 0 0}]{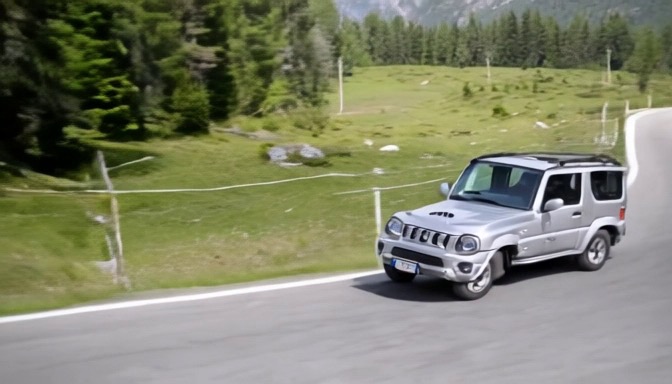} &
        \adjincludegraphics[clip,width=0.158\linewidth,trim={0 0 0 0}]{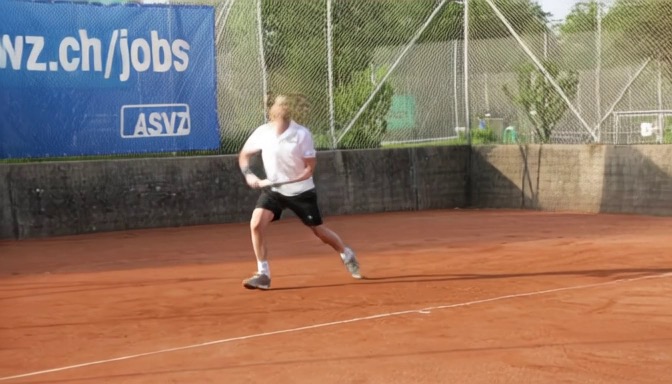} &
        \adjincludegraphics[clip,width=0.158\linewidth,trim={0 0 0 0}]{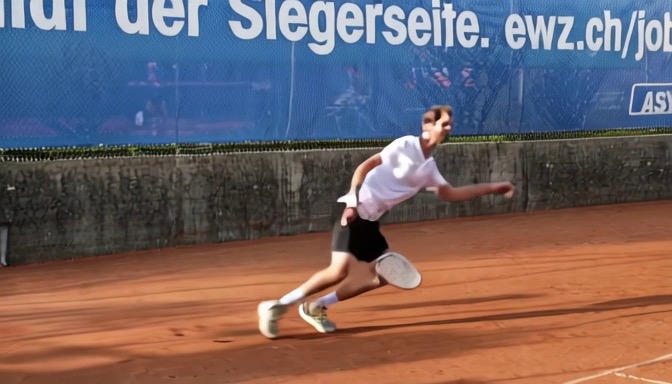} &
        \adjincludegraphics[clip,width=0.158\linewidth,trim={0 0 0 0}]{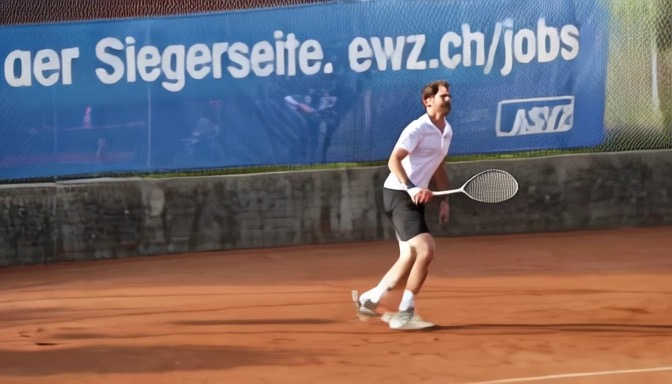} \\

        \raisebox{0.02\linewidth}{\rotatebox{90}{\makecell[c]{\scriptsize CogNVS}}} &
        \adjincludegraphics[clip,width=0.158\linewidth,trim={0 0 0 0}]{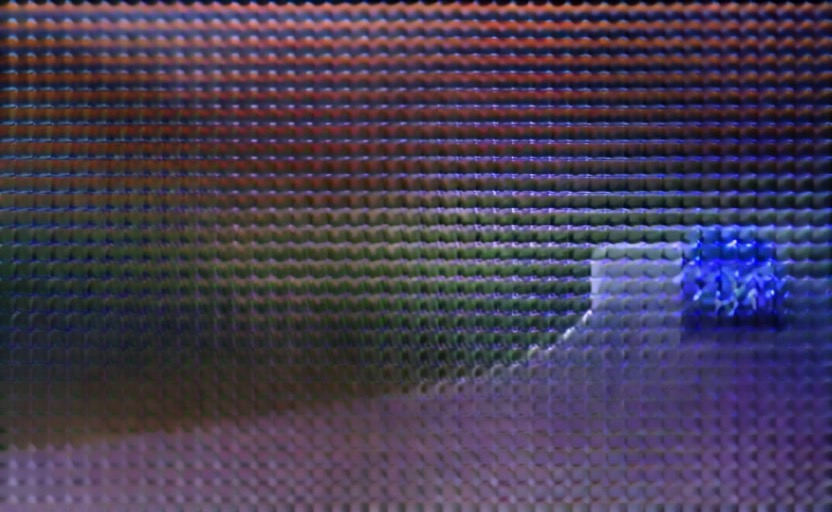} &
        \adjincludegraphics[clip,width=0.158\linewidth,trim={0 0 0 0}]{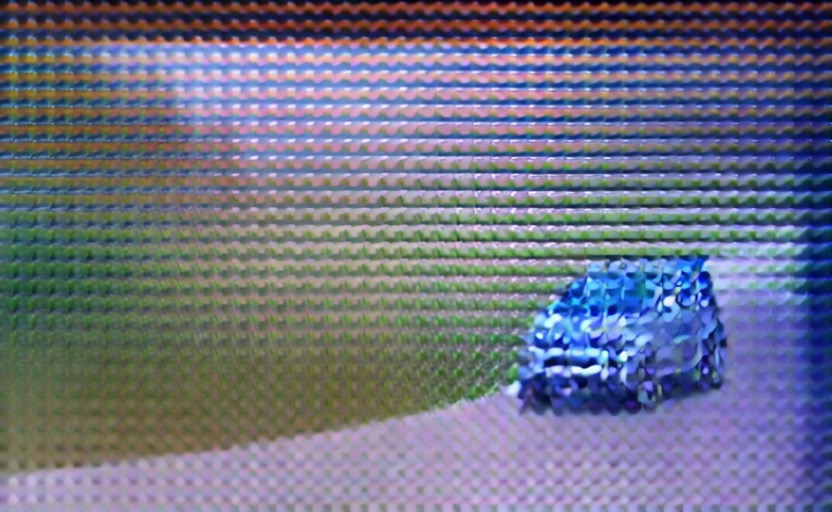} &
        \adjincludegraphics[clip,width=0.158\linewidth,trim={0 0 0 0}]{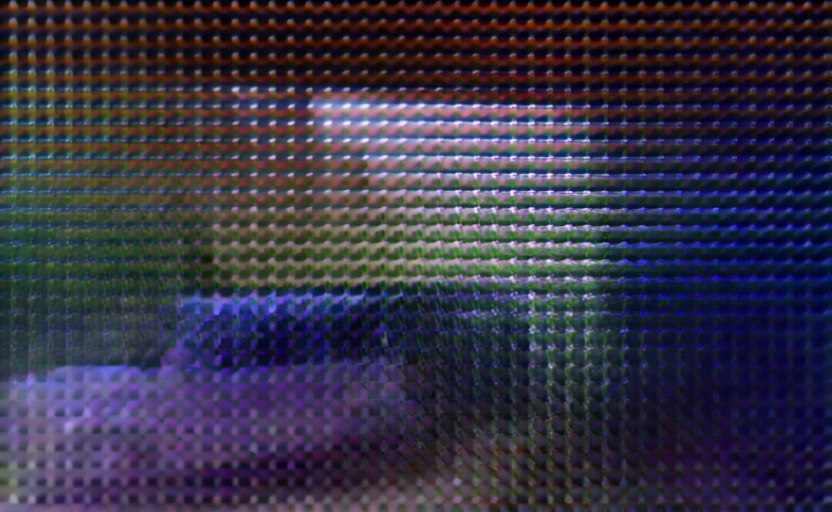} &
        \adjincludegraphics[clip,width=0.158\linewidth,trim={0 0 0 0}]{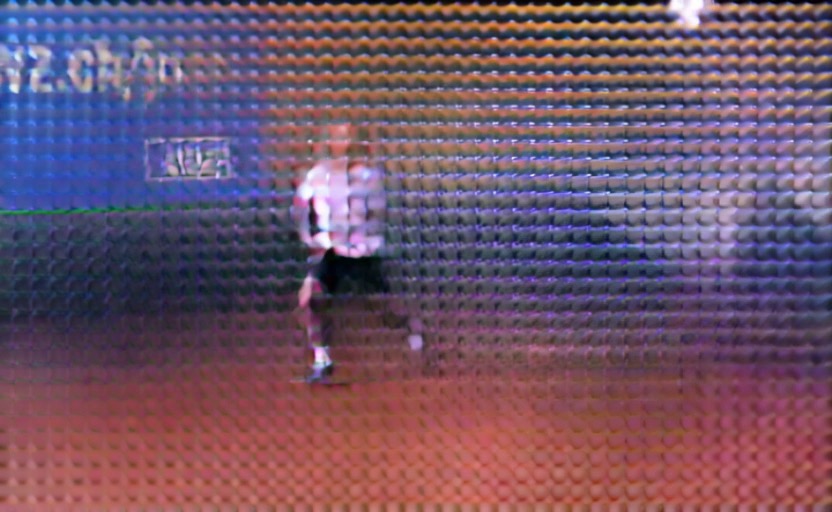} &
        \adjincludegraphics[clip,width=0.158\linewidth,trim={0 0 0 0}]{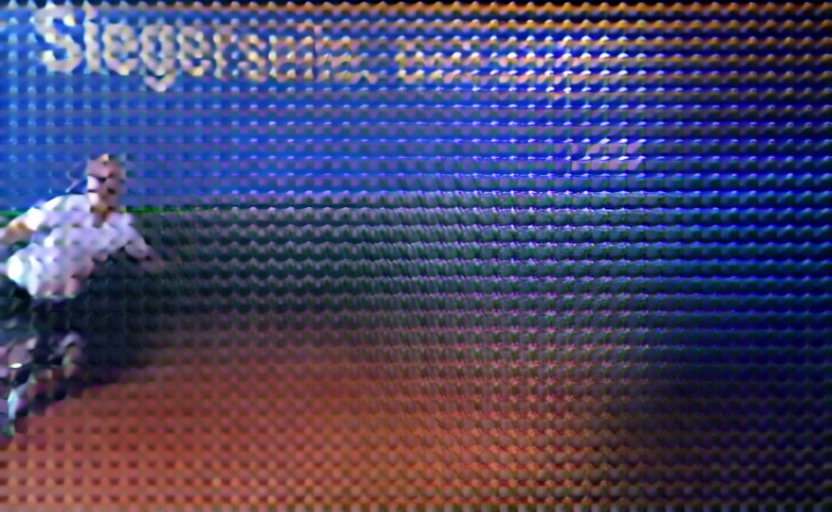} &
        \adjincludegraphics[clip,width=0.158\linewidth,trim={0 0 0 0}]{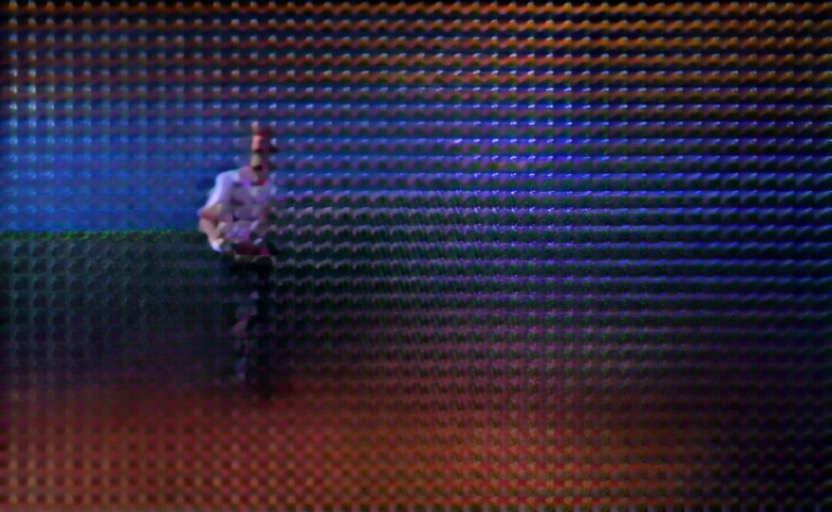} \\
        
        \raisebox{0.03\linewidth}{\rotatebox{90}{\scriptsize Ours}} &
        \adjincludegraphics[clip,width=0.158\linewidth,trim={0 0 0 0}]{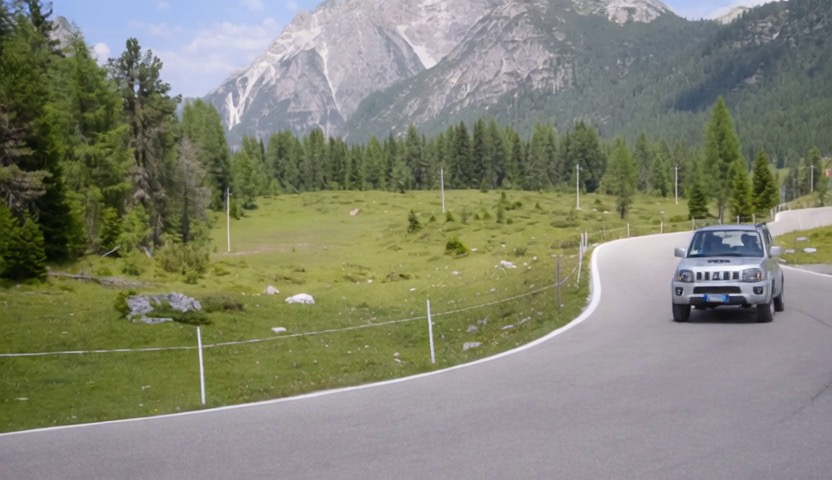} &
        \adjincludegraphics[clip,width=0.158\linewidth,trim={0 0 0 0}]{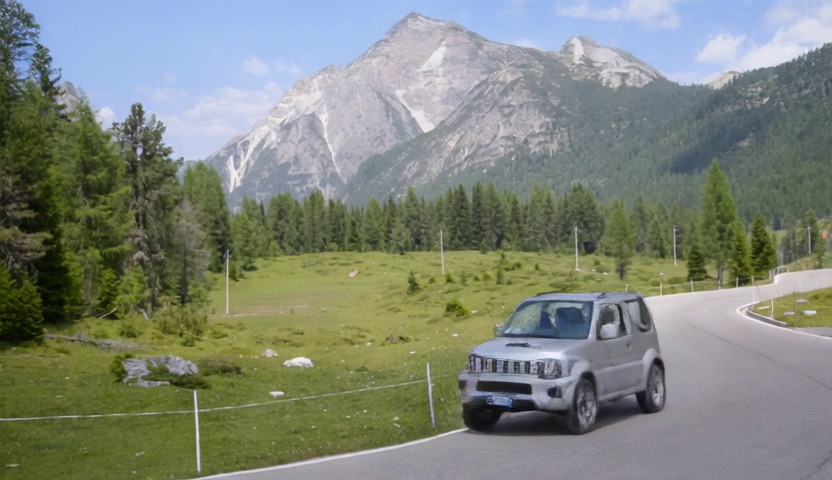} &
        \adjincludegraphics[clip,width=0.158\linewidth,trim={0 0 0 0}]{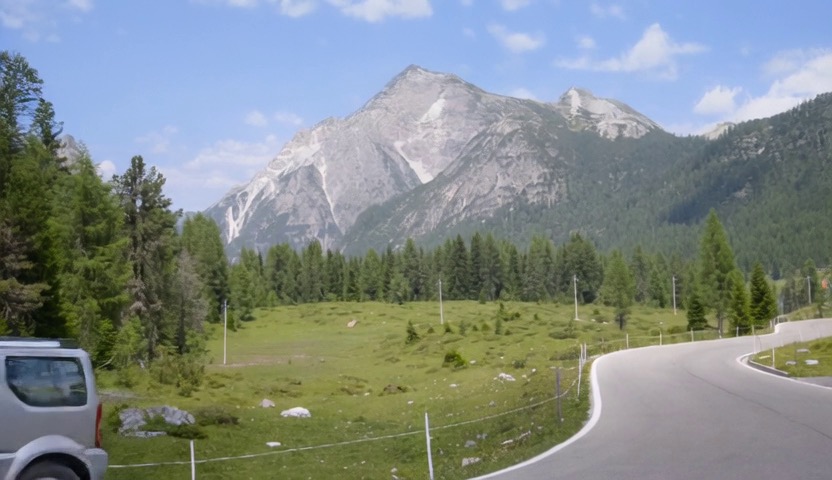} &
        \adjincludegraphics[clip,width=0.158\linewidth,trim={0 0 0 0}]{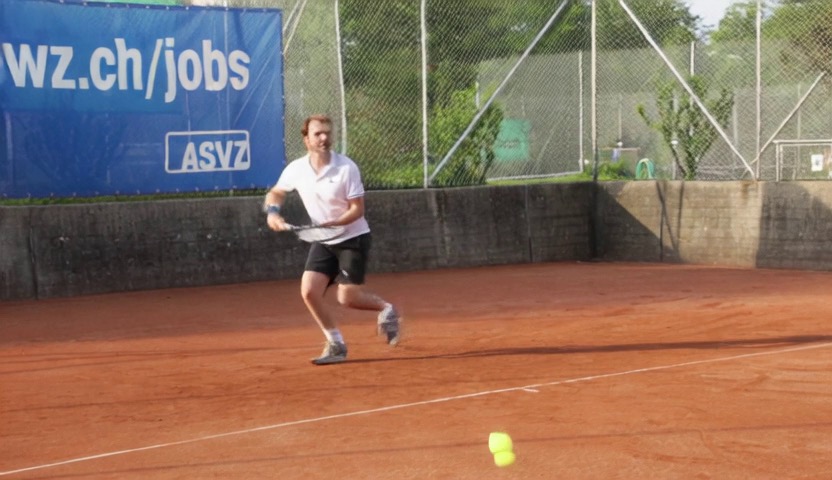} &
        \adjincludegraphics[clip,width=0.158\linewidth,trim={0 0 0 0}]{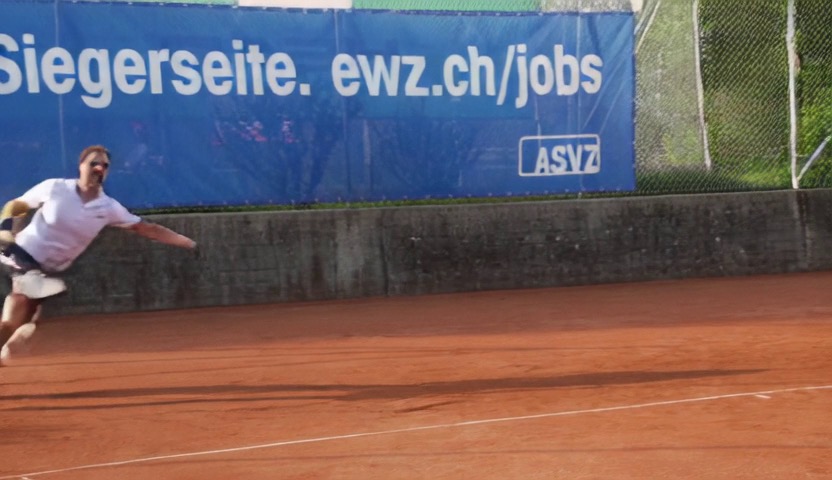} &
        \adjincludegraphics[clip,width=0.158\linewidth,trim={0 0 0 0}]{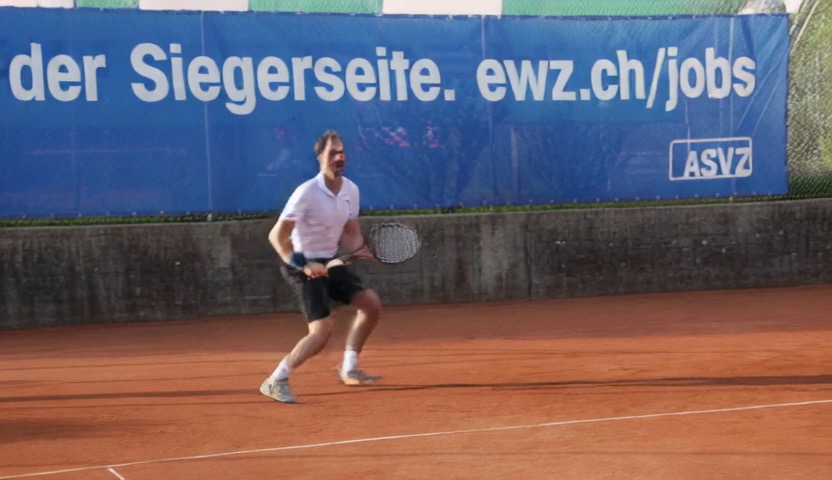} \\
        & \multicolumn{3}{c}{\normalsize \textit{Tilt Up}} &
        \multicolumn{3}{c}{\normalsize \textit{Arc Right}}  \\
    \end{tabular}
    \vspace{-1mm}
    \vspace{\abovefigcapmargin}
    \caption{\textbf{Qualitative results on the DAVIS dataset~\cite{pont20172017}.} ReDirector generates realistic video retakes (②) from dynamically captured input video (①), achieving better camera control, dynamic object localization, and background preservation.}
    \vspace{\belowfigcapmargin}
    \vspace{1mm}
    \label{fig:davis}
\end{figure*}

\begin{table*}[t!]
    \centering
    \setlength\tabcolsep{8pt}
    \resizebox{\linewidth}{!}{
    \begin{tabular}{lccccccccccc}
       \toprule
        \multirow{2}{*}[-1.33ex]{Method}  &\multicolumn{6}{c}{Visual quality$\uparrow$} & \multicolumn{2}{c}{Geometric Consistency} & \multicolumn{2}{c}{Camera Accuracy}  \\
        \arrayrulecolor{gray}\cmidrule(lr){2-7} \cmidrule(lr){8-9} \cmidrule(lr){10-11}
        & \makecell{Subject \\ Consistency}  & \makecell{Background \\ Consistency} &  \makecell{Aesthetic \\ Quality}  & \makecell{Imaging \\ Quality} &\makecell{Temporal \\ Flickering} & \makecell{Motion \\ Smoothness}  & Dyn-MEt3R$\uparrow$ & MEt3R$\downarrow$ & TransErr$\downarrow$ & RotErr$\downarrow$  \\
        \midrule
        GCD~\cite{van2024generative}  & 0.7117 & 0.8400 & 0.3998 & 0.4928 & 0.9526 & 0.9639 & 0.6898 & 0.4438 & 0.1062 & 22.853 \\
        ReCamMaster~\cite{Bai_2025_ICCV}  & \cellcolor{tabsecond} 0.9008 & \cellcolor{tabfirst} 0.9212 & \cellcolor{tabsecond} 0.5064 & \cellcolor{tabsecond} 0.6461 & \cellcolor{tabfirst} 0.9673 & \cellcolor{tabfirst} 0.9881 & \cellcolor{tabsecond} 0.7857 & \cellcolor{tabthird} 0.3472 & \cellcolor{tabsecond} 0.0292 & \cellcolor{tabsecond} 2.347 \\
        TrajectoryCrafter~\cite{Yu_2025_ICCV} & 0.8846 & \cellcolor{tabsecond} 0.9174 & \cellcolor{tabthird}  0.5046 & \cellcolor{tabthird}  0.6071 & 0.9364 & \cellcolor{tabthird} 0.9727 & \cellcolor{tabthird} 0.7338 & \cellcolor{tabsecond} 0.3272 & \cellcolor{tabthird} 0.0697 & \cellcolor{tabthird} 9.115 \\
        CogNVS~\cite{chen2025cognvs}  & \cellcolor{tabthird} 0.8929 & 0.9055 & 0.2160 & 0.4300 & \cellcolor{tabsecond} 0.9637 & 0.9721 & 0.6845 & 0.4036 & 0.0768 & 10.878 \\
        \arrayrulecolor{gray} \midrule
         \textbf{Ours (ReDirector)} & \cellcolor{tabfirst} 0.9043 & \cellcolor{tabthird} 0.9171 & \cellcolor{tabfirst} 0.5149 & \cellcolor{tabfirst} 0.6668 & \cellcolor{tabthird} 0.9548 & \cellcolor{tabsecond} 0.9867 & \cellcolor{tabfirst} 0.8477 & \cellcolor{tabfirst} 0.3073 & \cellcolor{tabfirst} 0.0165 & \cellcolor{tabfirst} 1.666 \\
       \arrayrulecolor{black}\bottomrule
    \end{tabular}
    }
    \vspace{\abovetabcapmargin}
    \caption{\textbf{Quantitative results on the DAVIS dataset~\cite{pont20172017}.} ReDirector significantly improves geometric consistency and camera control. Some visual quality metrics are lower because ReDirector explores larger scene scales under the same camera motion, whereas methods with milder camera movement naturally yield higher scores on metrics that favor consistent backgrounds and smoother motion.}
    \vspace{\belowtabcapmargin}
    \label{tab:davis_main}
\end{table*}

\paragraph{Implementation details.}
We fine-tune ReDirector on the MultiCamVideo dataset~\cite{Bai_2025_ICCV} for 20K steps at a resolution of $480 \times 832$ with a learning rate of 0.0001 and a batch size of 8. All training videos contain 81 frames. Training takes about 90 hours on eight RTX Pro 6000 Blackwell GPUs.

\vspace{\paramargin}
\paragraph{Baselines.}
We compare our method with state-of-the-art video retake generation approaches~\cite{van2024generative, Bai_2025_ICCV, Yu_2025_ICCV, chen2025cognvs}.
GCD~\cite{van2024generative} and ReCamMaster~\cite{Bai_2025_ICCV} implicitly generate video retakes conditioned on camera extrinsic parameters, and mainly focus on retakes of 14 and 81 frames, respectively. TrajectoryCrafter~\cite{Yu_2025_ICCV} and CogNVS~\cite{chen2025cognvs} are the explicit warping-based methods, which primarily focus on 49-frame retakes.

\vspace{\paramargin}
\paragraph{Evaluation protocol.}
We construct an evaluation set using 50 videos from the DAVIS dataset~\cite{pont20172017} and 10 target camera trajectories from ReCamMaster~\cite{Bai_2025_ICCV}, yielding 500 test cases. Note that video lengths range from a few dozen frames to about 100 frames, and the target trajectories include rotations and translations along multiple axes. We extract camera poses for the input videos using ViPE~\cite{huang2025vipe}. For evaluation, we use multiple metrics from VBench~\cite{huang2024vbench} to assess visual quality, Dyn-MEt3R~\cite{Park_2025_ICCV} to measure geometric consistency of the generated retakes, and per-frame MEt3R~\cite{Asim_2025_CVPR} to quantify consistency with the input video. We also report TransErr and RotErr~\cite{he2024cameractrl, zhang2024raydiffusion, Go_2025_CVPR, Go_2025_ICCV}, which measure errors in relative translation and rotation for every frame pair by estimating input camera poses using ViPE.

\subsection{Main Results}

\paragraph{Any-length video retake generation.}
\Cref{fig:davis} shows the qualitative results on the DAVIS dataset~\cite{pont20172017}. We observe that previous methods degrade when the input video length deviates from their assumed setting. Also, explicit approaches~\cite{Yu_2025_ICCV, chen2025cognvs} fail to faithfully preserve the input content, even though they condition on warped frames, while implicit methods~\cite{van2024generative, Bai_2025_ICCV} often produce blurry objects and sometimes cause dynamic objects to disappear. As the sequence progresses and the visibility of previously seen views decreases, especially under dynamic camera motion, they increasingly fail to localize dynamic objects and maintain consistent backgrounds. In contrast, ReDirector produces high-quality video retakes that remain stable even under dynamic camera motion, accurately localizes dynamic objects over time, and consistently preserves static backgrounds across distant viewpoints. It better preserves fine details, avoids warping artifacts, and maintains a coherent scene layout that closely follows the target camera trajectories. \Cref{tab:davis_main} further demonstrates the superiority of our method, with particularly large gains in geometric consistency and camera controllability. We attribute these improvements to the seamless integration of camera encoding with 3D RoPE and the incorporation of geometry-aware attention, which together enable more reliable multi-view reasoning and physically accurate video retakes.

\begin{figure*}[t!]
    \centering
    \setlength\tabcolsep{1pt}
    
    \begin{tabular}{@{}c@{\,}cccccc@{}}
        \raisebox{0.02\linewidth}{\rotatebox{90}{\scriptsize Camera}} & \multicolumn{2}{c}{\adjincludegraphics[clip,width=0.158\linewidth,height=0.0923\linewidth,trim={0 0 0 0}]{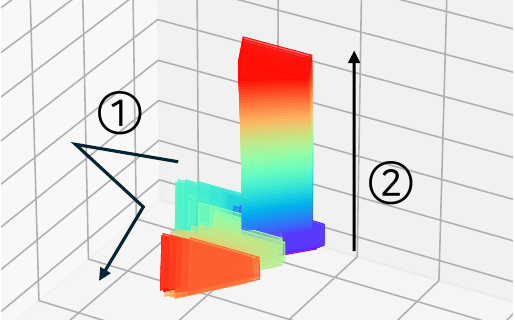}} & \multicolumn{2}{c}{\adjincludegraphics[clip,width=0.158\linewidth,height=0.0923\linewidth,trim={0 0 0 0}]{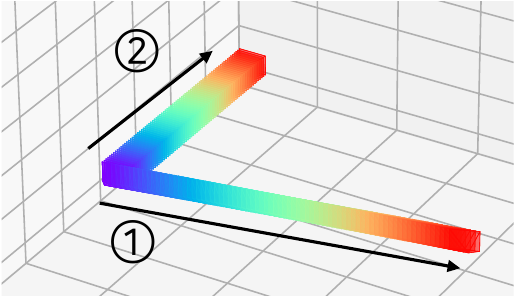}} & \multicolumn{2}{c}{\adjincludegraphics[clip,width=0.158\linewidth,height=0.0923\linewidth,trim={0 0 0 0}]{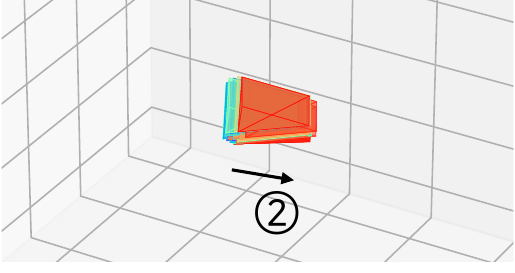}}  \\
    
        \raisebox{0.01\linewidth}{\rotatebox{90}{\scriptsize Input Video}} &
        \adjincludegraphics[clip,width=0.16\linewidth,trim={0 0 0 0}]{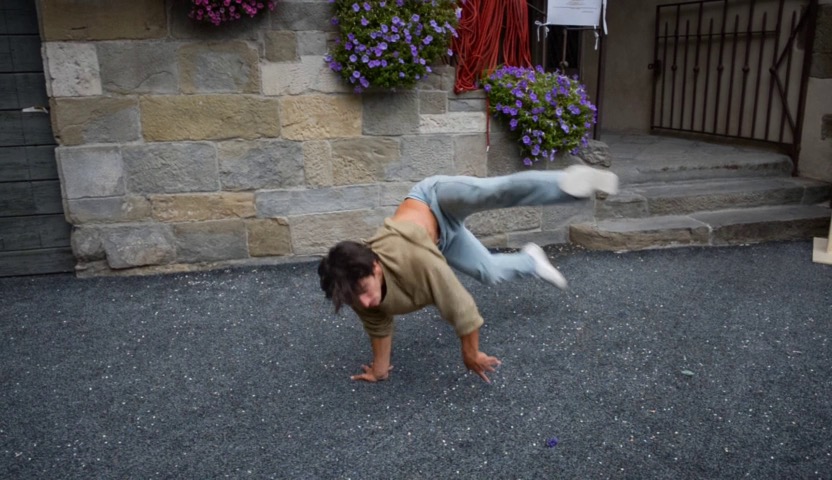} &
        \adjincludegraphics[clip,width=0.16\linewidth,trim={0 0 0 0}]{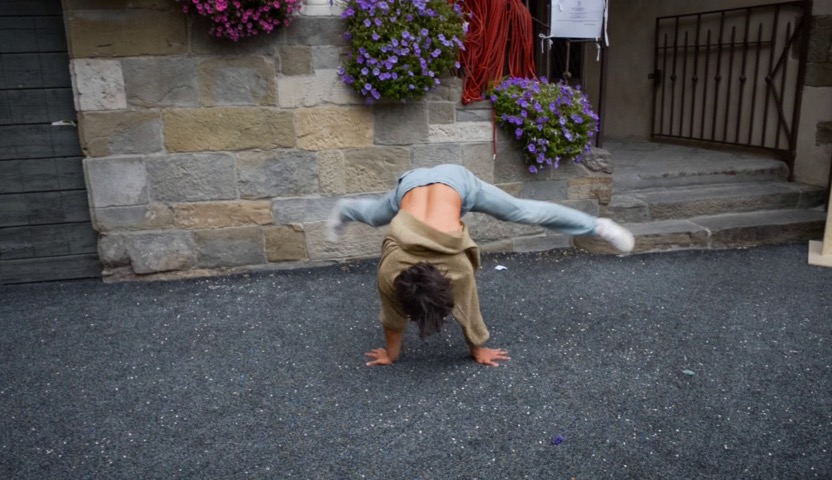} &
        \adjincludegraphics[clip,width=0.16\linewidth,trim={0 0 0 0}]{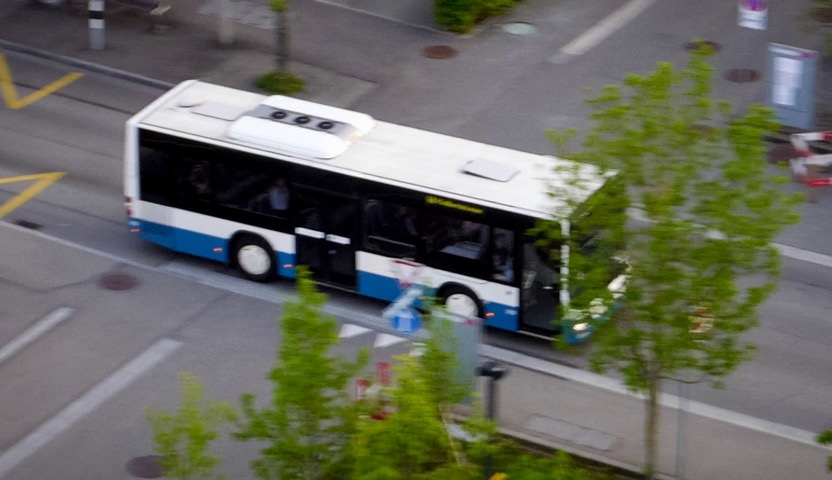} &
        \adjincludegraphics[clip,width=0.16\linewidth,trim={0 0 0 0}]{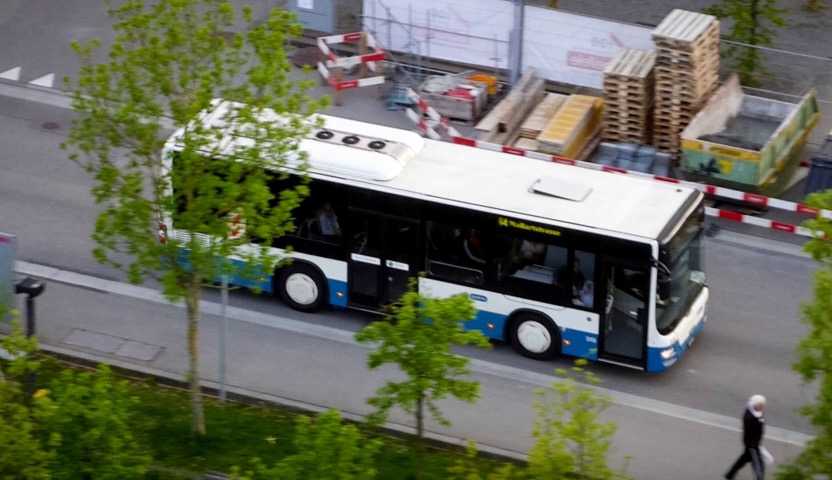} &
        \adjincludegraphics[clip,width=0.16\linewidth,trim={0 0 0 0}]{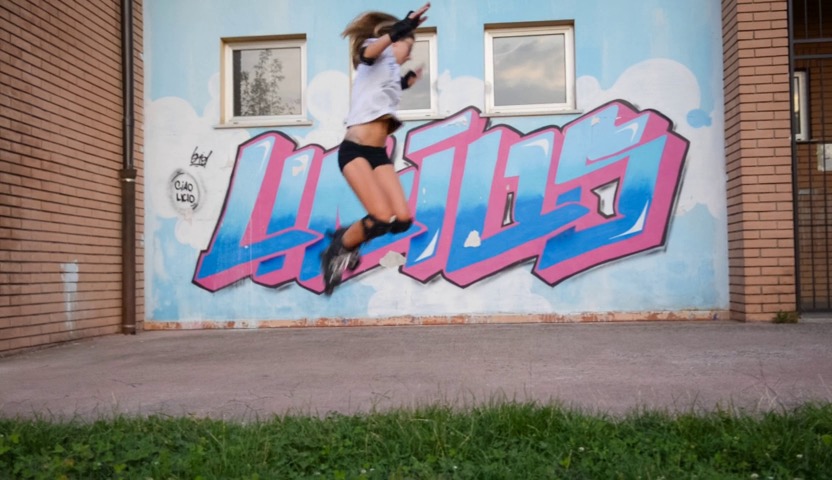} &
        \adjincludegraphics[clip,width=0.16\linewidth,trim={0 0 0 0}]{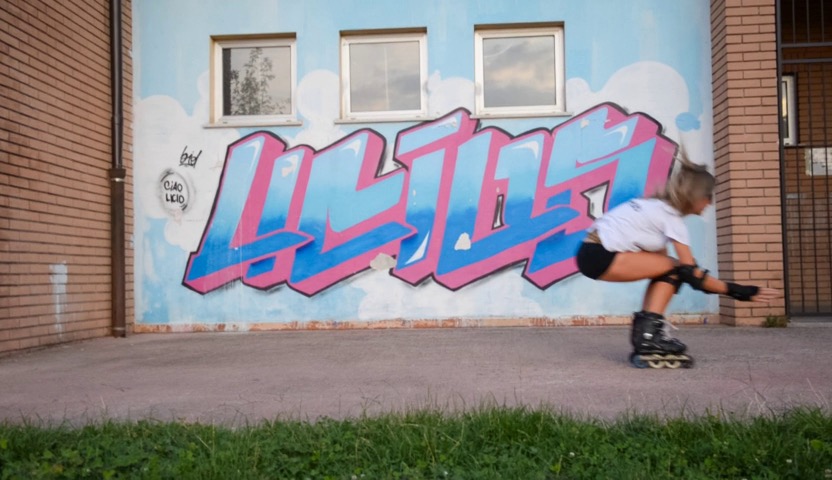} \\

        \raisebox{0.02\linewidth}{\rotatebox{90}{\scriptsize Addition}} &
        \adjincludegraphics[clip,width=0.16\linewidth,trim={0 0 0 0}]{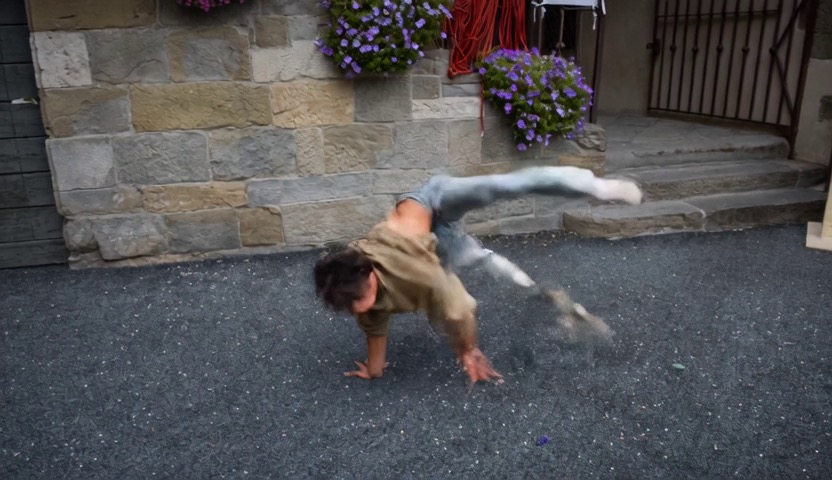} &
        \adjincludegraphics[clip,width=0.16\linewidth,trim={0 0 0 0}]{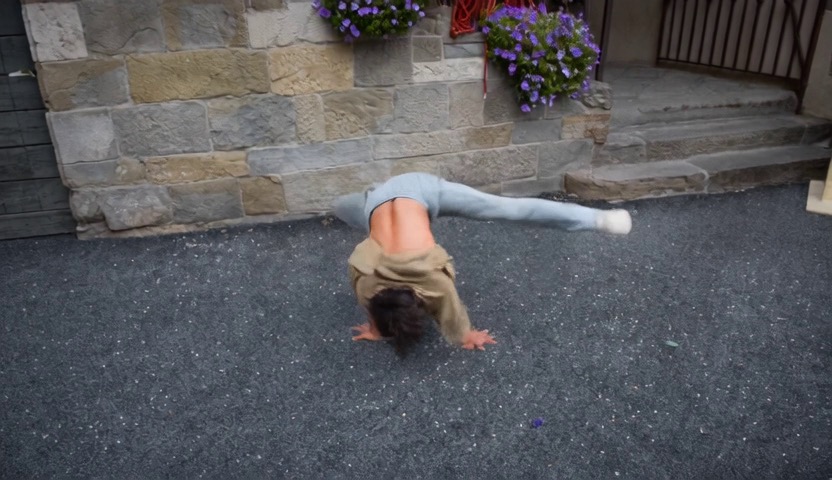} &
        \adjincludegraphics[clip,width=0.16\linewidth,trim={0 0 0 0}]{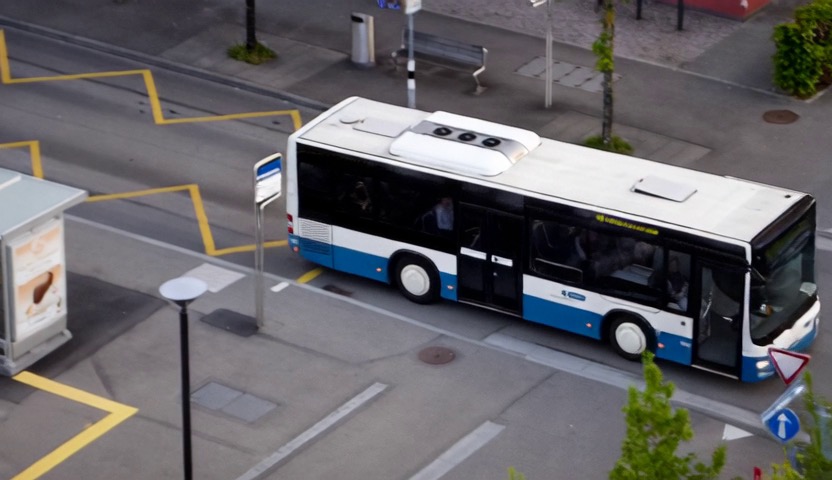} &
        \adjincludegraphics[clip,width=0.16\linewidth,trim={0 0 0 0}]{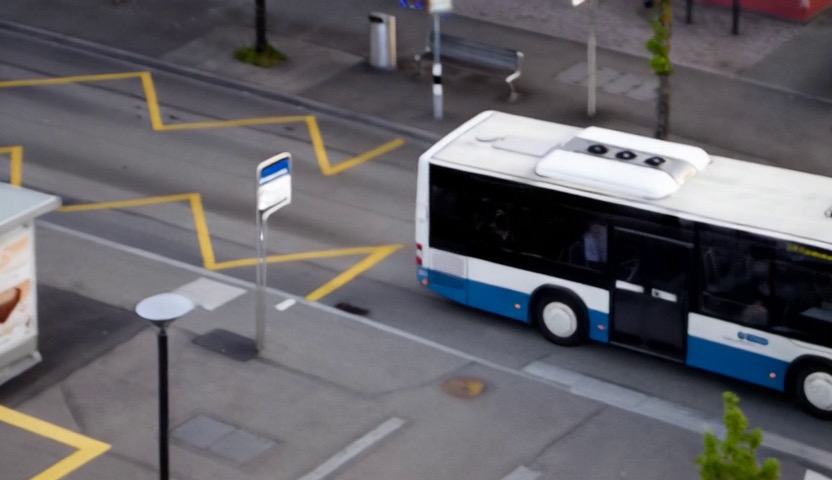} &
        \adjincludegraphics[clip,width=0.16\linewidth,trim={0 0 0 0}]{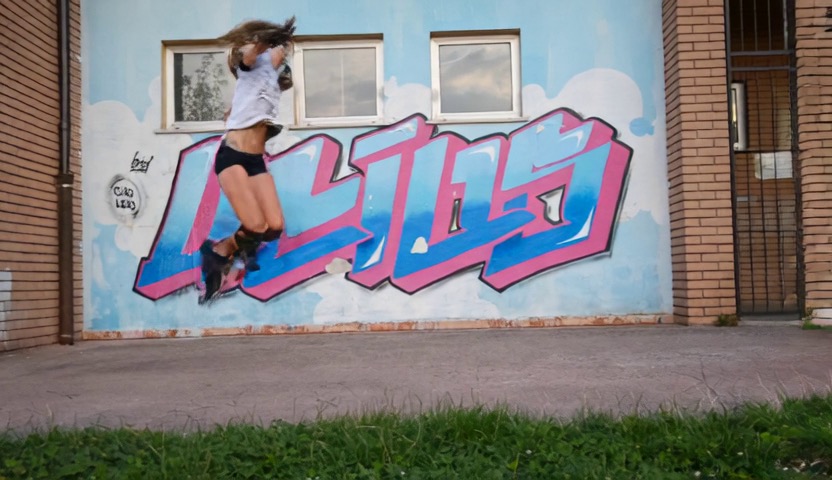} &
        \adjincludegraphics[clip,width=0.16\linewidth,trim={0 0 0 0}]{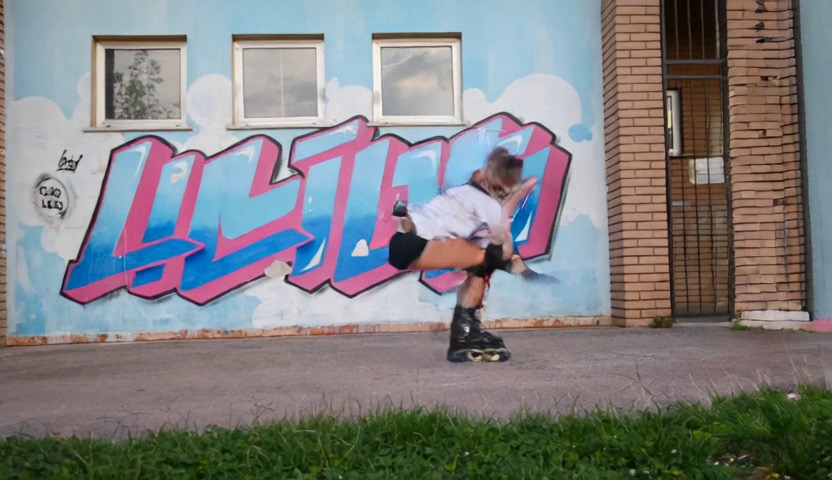} \\

        \raisebox{0.03\linewidth}{\rotatebox{90}{\scriptsize RoCE}} &
        \adjincludegraphics[clip,width=0.16\linewidth,trim={0 0 0 0}]{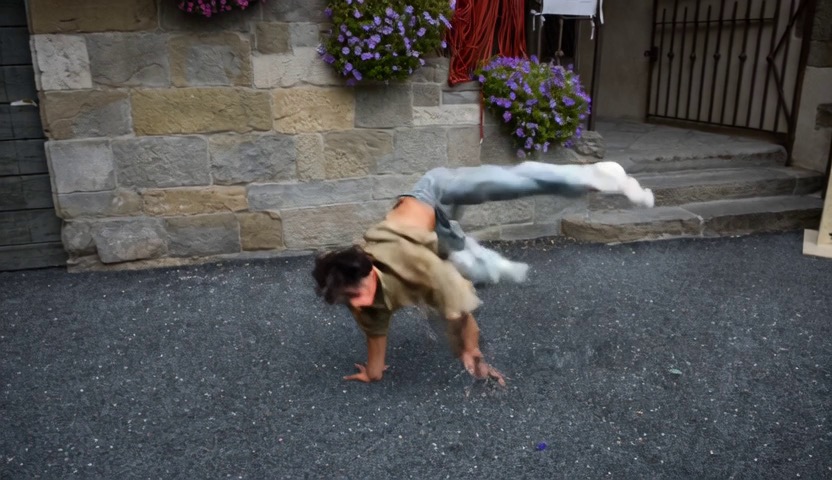} &
        \adjincludegraphics[clip,width=0.16\linewidth,trim={0 0 0 0}]{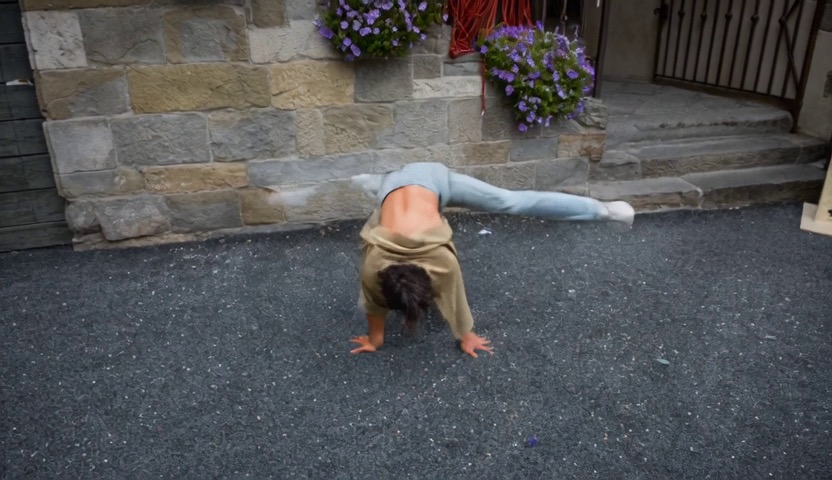} &
        \adjincludegraphics[clip,width=0.16\linewidth,trim={0 0 0 0}]{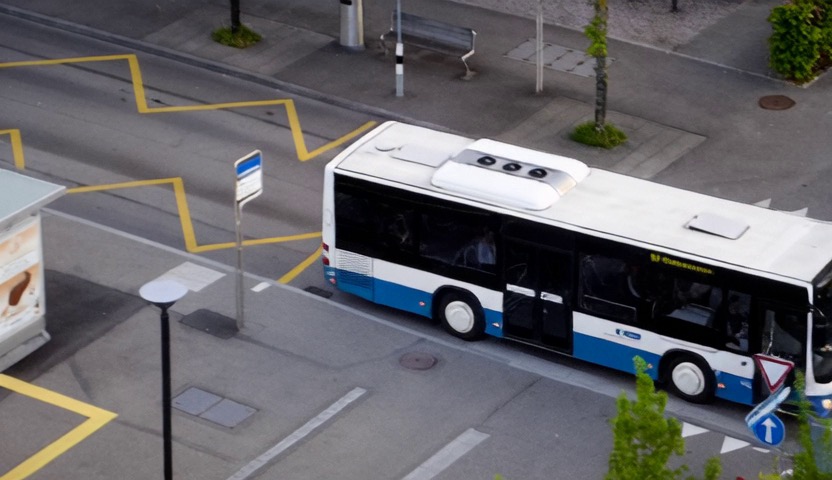} &
        \adjincludegraphics[clip,width=0.16\linewidth,trim={0 0 0 0}]{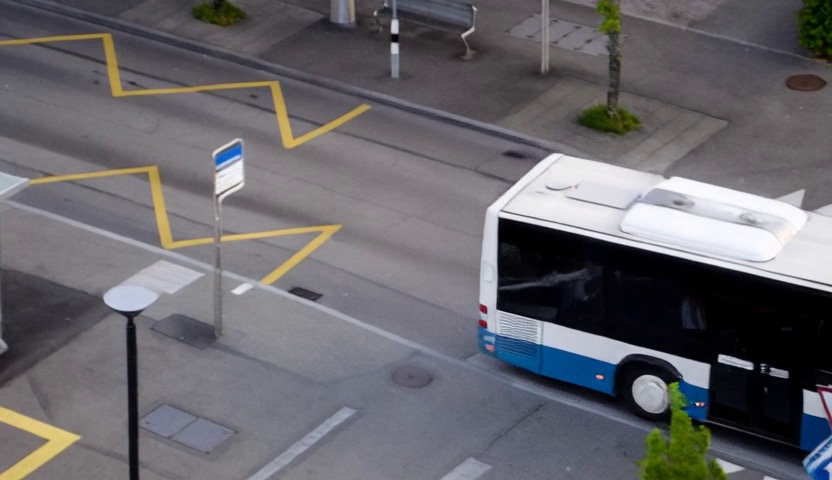} &
        \adjincludegraphics[clip,width=0.16\linewidth,trim={0 0 0 0}]{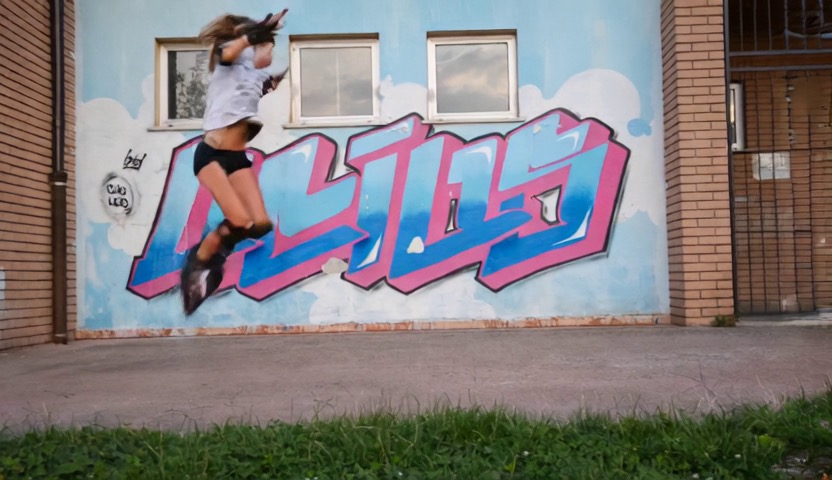} &
        \adjincludegraphics[clip,width=0.16\linewidth,trim={0 0 0 0}]{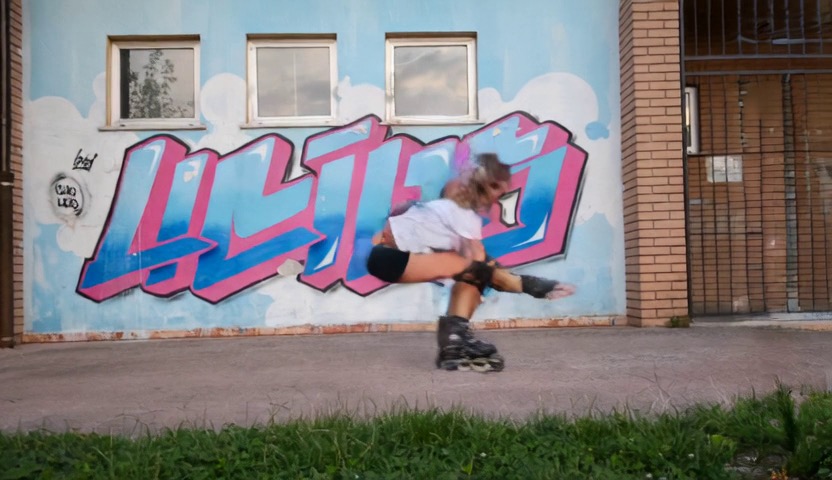} \\

        \raisebox{0.007\linewidth}{\rotatebox{90}{\scriptsize RoCE + GTA}} &
        \adjincludegraphics[clip,width=0.16\linewidth,trim={0 0 0 0}]{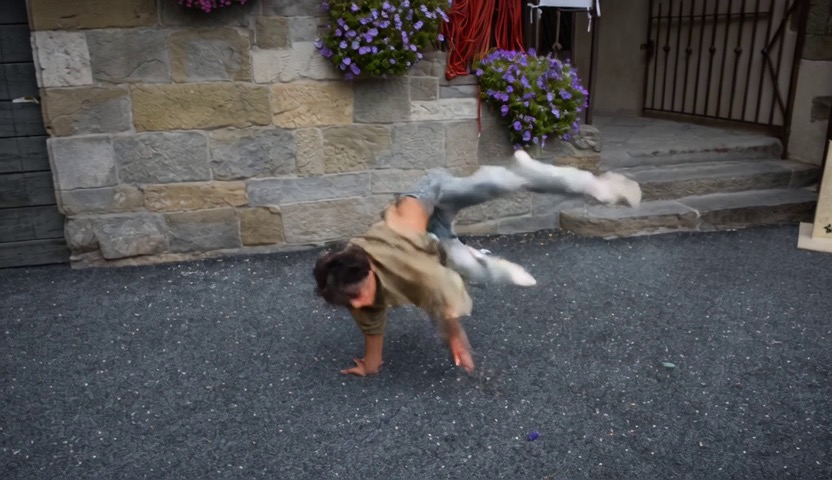} &
        \adjincludegraphics[clip,width=0.16\linewidth,trim={0 0 0 0}]{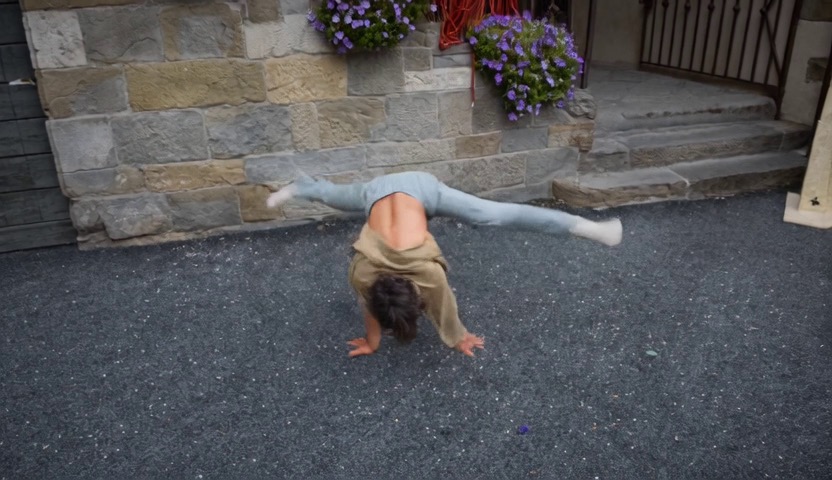} &
       \adjincludegraphics[clip,width=0.16\linewidth,trim={0 0 0 0}]{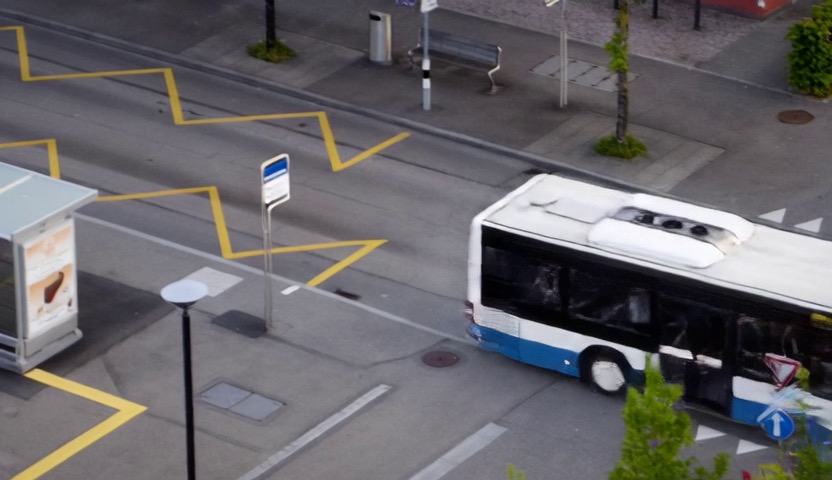} &
        \adjincludegraphics[clip,width=0.16\linewidth,trim={0 0 0 0}]{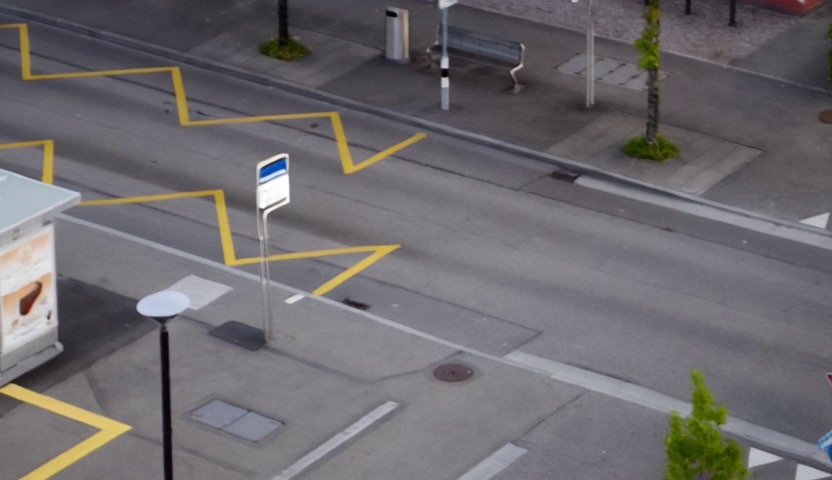} &
        \adjincludegraphics[clip,width=0.16\linewidth,trim={0 0 0 0}]{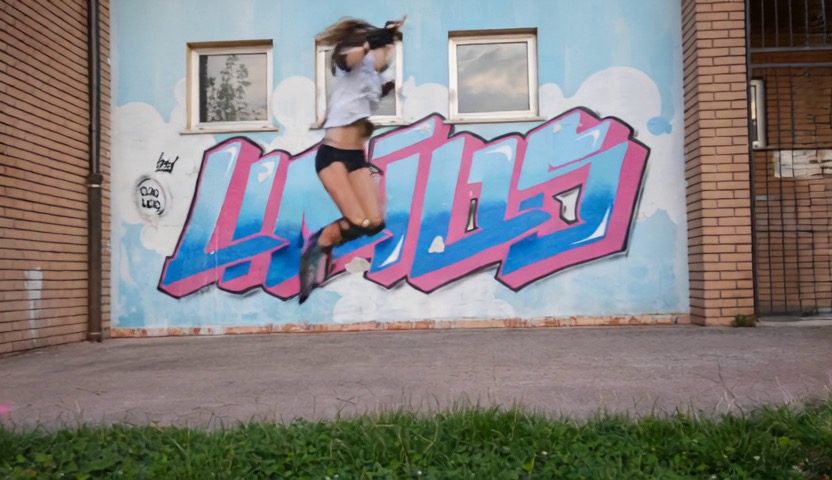} &
        \adjincludegraphics[clip,width=0.16\linewidth,trim={0 0 0 0}]{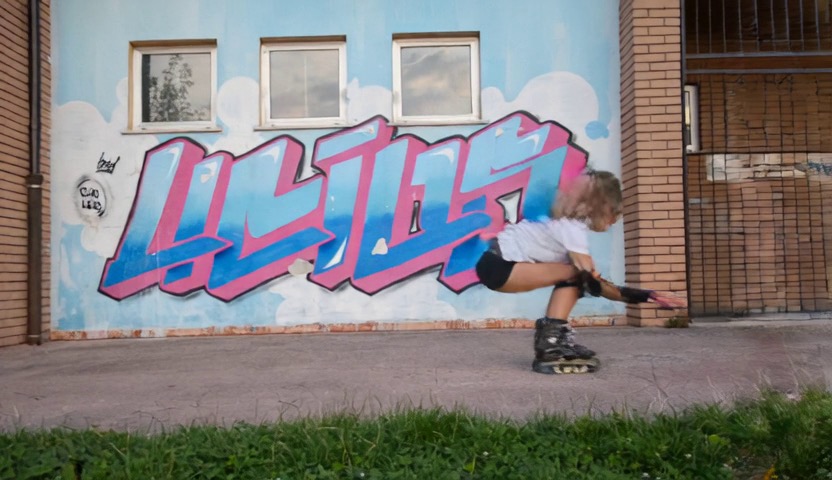} \\

        & \multicolumn{2}{c}{\normalsize \textit{Translate Up}} &
        \multicolumn{2}{c}{\normalsize \textit{Zoom In}} & \multicolumn{2}{c}{\normalsize \textit{Pan Right}} \\
        
    \end{tabular}
    \vspace{\abovefigcapmargin}
    \vspace{-1mm}
    \caption{\textbf{Qualitative ablation of conditioning type on DAVIS~\cite{pont20172017}.} Compared to simple additive conditioning, RoCE enhances geometric consistency, while the additional geometry-aware attention further boosts retake quality and achieves accurate dynamic object localization.}
    \vspace{\belowfigcapmargin}
    \vspace{2mm}
    \label{fig:abl_davis}
\end{figure*}

\begin{table*}[t!]
    \centering
    \setlength\tabcolsep{12pt}
    \resizebox{\linewidth}{!}{
    \begin{tabular}{lccccccc}
       \toprule
        \multirow{2}{*}{Method} & \multirow{2}{*}{Condition Type} &\multicolumn{2}{c}{Visual quality} & \multicolumn{2}{c}{Geometric Consistency} & \multicolumn{2}{c}{Camera Accuracy}  \\
        \arrayrulecolor{gray}\cmidrule(lr){3-4} \cmidrule(lr){5-6} \cmidrule(lr){7-8}
        & & \makecell{Aesthetic Quality$\uparrow$}  & \makecell{Imaging Quality$\uparrow$}  & Dyn-MEt3R$\uparrow$ &  MEt3R$\downarrow$ & TransErr$\downarrow$ & RotErr$\downarrow$  \\
        \midrule
         ReCamMaster~\cite{Bai_2025_ICCV} & Addition & 0.5064 & 0.6461 & 0.7857 & 0.3472 & 0.0292 & 2.347 \\
        \midrule
        Ours (ReDirector) & Addition & \cellcolor{tabthird} 0.5121 & \cellcolor{tabthird} 0.6643  & \cellcolor{tabsecond} 0.8378 & \cellcolor{tabsecond} 0.3159 & \cellcolor{tabthird} 0.0202 & \cellcolor{tabthird} 1.975 \\
       Ours (ReDirector) & RoCE & \cellcolor{tabsecond} 0.5139 & \cellcolor{tabsecond} 0.6656 & \cellcolor{tabthird} 0.8341 &\cellcolor{tabthird} 0.3164 & \cellcolor{tabsecond} 0.0193 & \cellcolor{tabsecond} 1.897 \\
        Ours (ReDirector) & RoCE + GTA~\cite{miyato2024gta} &  \cellcolor{tabfirst} 0.5149 & \cellcolor{tabfirst} 0.6668 & \cellcolor{tabfirst} 0.8477 & \cellcolor{tabfirst} 0.3073 & \cellcolor{tabfirst} 0.0165 & \cellcolor{tabfirst} 1.666 \\
       \arrayrulecolor{black}\bottomrule
    \end{tabular}
    }
    \vspace{\abovetabcapmargin}
    \caption{\textbf{Quantitative ablations of conditioning type on DAVIS~\cite{pont20172017}.} All components consistently improve the overall performance.}
    \vspace{\belowtabcapmargin}
    \vspace{-1mm}
    \label{tab:davis_abl}
\end{table*}

\vspace{\paramargin}
\paragraph{Ablation study.}
We conduct ablation studies to disentangle the contributions of each component in ReDirector. First, we correct the misuse of RoPE and camera encodings by introducing a shared 3D RoPE for input and target videos and employing token-wise encodings. Next, we replace simple additive conditioning with RoCE, a camera-conditioned RoPE phase shift in spatial axes. Finally, we incorporate geometry-aware attention on top of RoCE.

\Cref{fig:abl_davis} and \Cref{tab:davis_abl} summarize the results. Compared to ReCamMaster~\cite{Bai_2025_ICCV}, correcting RoPE and camera encodings improves all metrics, showing that tight alignment between input and target videos and proper integration of camera signals into positional encodings are crucial. Introducing RoCE yields further gains in visual quality and camera accuracy, indicating that phase shift in the complex domain is more effective than naïve addition for camera encoding. However, we observe a drop in geometric consistency metrics, suggesting that RoCE improves coarse alignment but still struggles to maintain fine-grained, multi-view consistent details. Finally, adding geometry-aware attention delivers the best performance across both geometric and visual metrics, confirming that the combination of aligned 3D RoPE, RoCE-based camera encoding, and geometry-aware attention is key to accurate camera control, better geometric consistency, and strong camera controllability.

\begin{figure*}[t!]
    \centering
    \setlength\tabcolsep{0.5pt}
    
    \begin{tabular}{cc:cc:cc:cc:cc:c}
        \adjincludegraphics[clip,width=0.088\linewidth,height=0.117\linewidth,trim={0 0 0 0}]{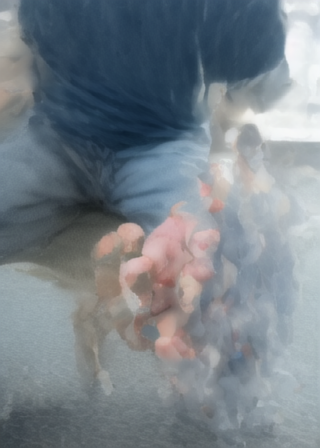} &
        \adjincludegraphics[clip,width=0.088\linewidth,height=0.117\linewidth,trim={0 0 0 0}]{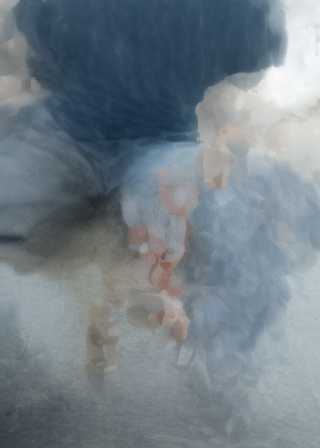} &
        \adjincludegraphics[clip,width=0.088\linewidth,height=0.117\linewidth,trim={0 0 0 0}]{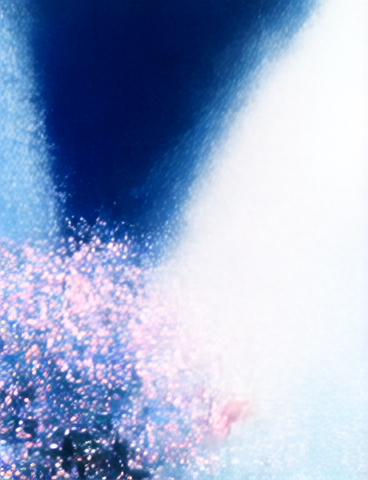} &
        \adjincludegraphics[clip,width=0.088\linewidth,height=0.117\linewidth,trim={0 0 0 0}]{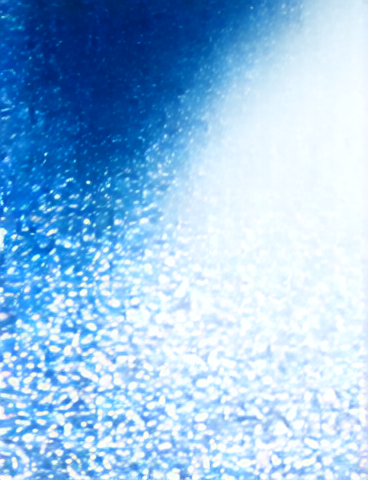} &
        \adjincludegraphics[clip,width=0.088\linewidth,height=0.117\linewidth,trim={0 0 0 0}]{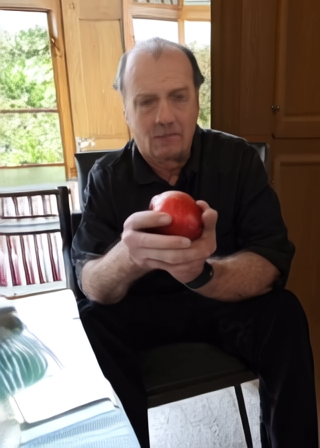} &
        \adjincludegraphics[clip,width=0.088\linewidth,height=0.117\linewidth,trim={0 0 0 0}]{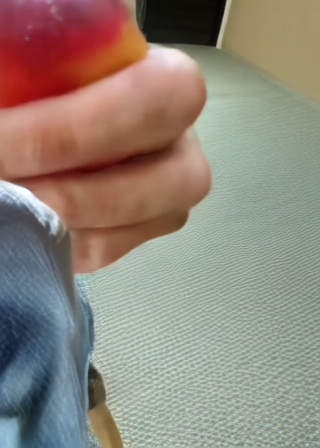} &
        \adjincludegraphics[clip,width=0.088\linewidth,height=0.117\linewidth,trim={0 0 0 0}]{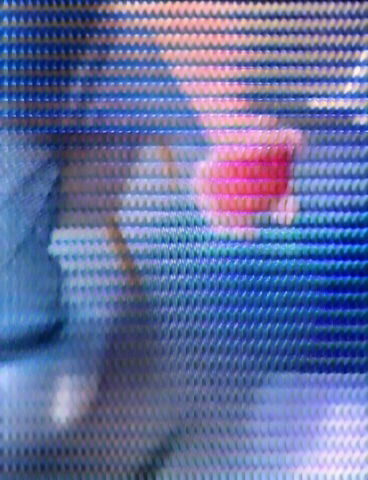} &
        \adjincludegraphics[clip,width=0.088\linewidth,height=0.117\linewidth,trim={0 0 0 0}]{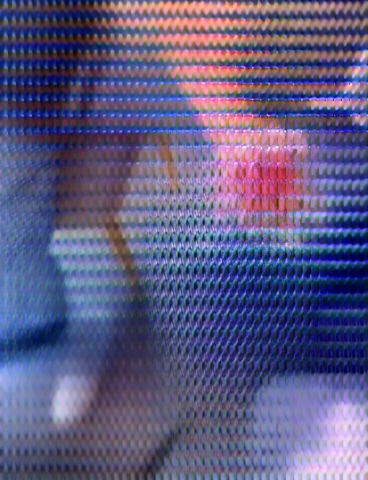} &
        \adjincludegraphics[clip,width=0.088\linewidth,height=0.117\linewidth,trim={0 0 0 0}]{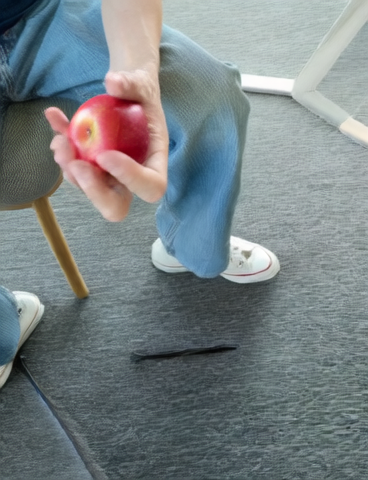} &
        \adjincludegraphics[clip,width=0.088\linewidth,height=0.117\linewidth,trim={0 0 0 0}]{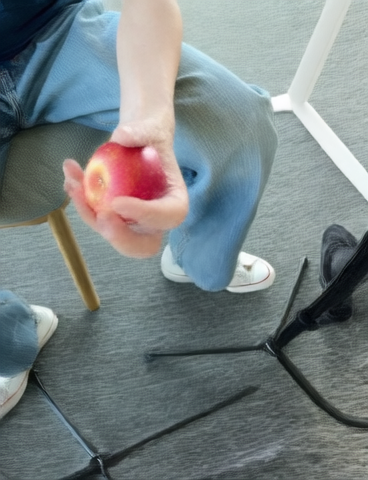} &
        \adjincludegraphics[clip,width=0.088\linewidth,height=0.117\linewidth,trim={0 0 0 0}]{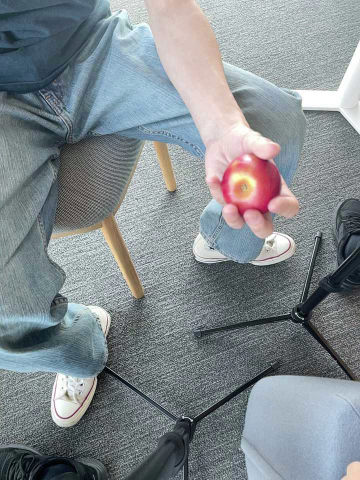} \\
        
        \adjincludegraphics[clip,width=0.088\linewidth,height=0.117\linewidth,trim={0 0 0 0}]{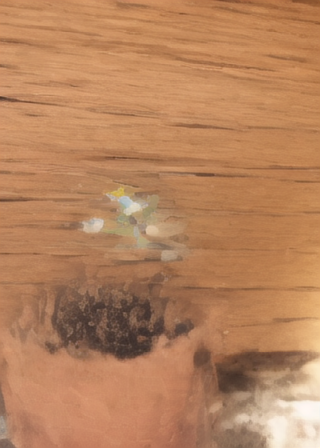} &
        \adjincludegraphics[clip,width=0.088\linewidth,height=0.117\linewidth,trim={0 0 0 0}]{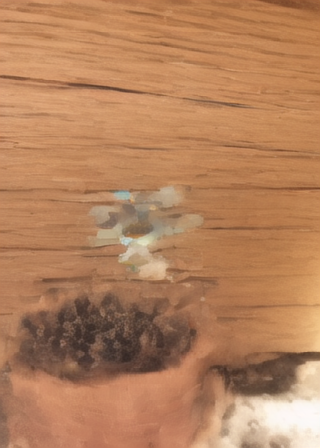} &
        \adjincludegraphics[clip,width=0.088\linewidth,height=0.117\linewidth,trim={0 0 0 0}]{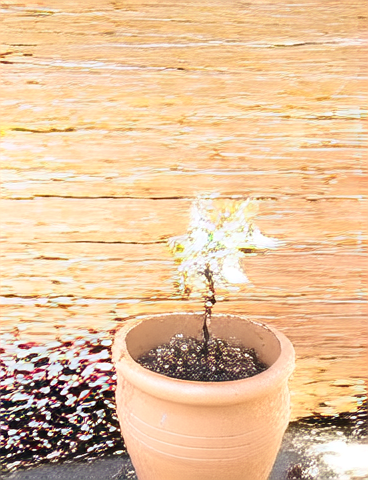} &
        \adjincludegraphics[clip,width=0.088\linewidth,height=0.117\linewidth,trim={0 0 0 0}]{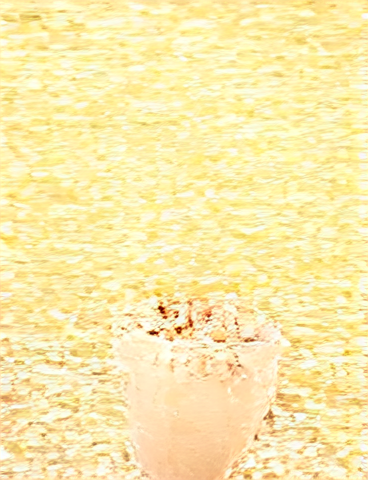} &
        \adjincludegraphics[clip,width=0.088\linewidth,height=0.117\linewidth,trim={0 0 0 0}]{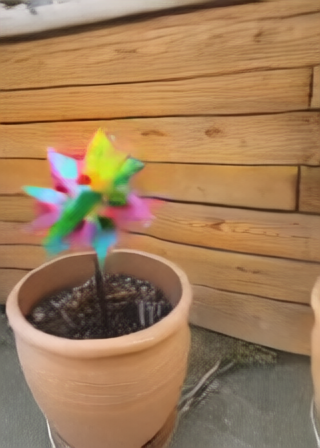} &
        \adjincludegraphics[clip,width=0.088\linewidth,height=0.117\linewidth,trim={0 0 0 0}]{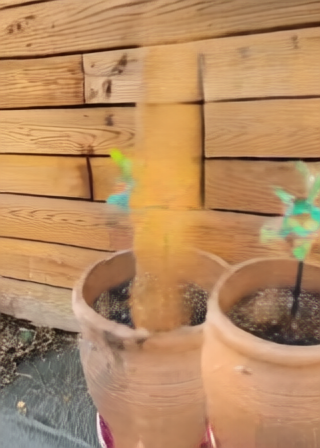} &
        \adjincludegraphics[clip,width=0.088\linewidth,height=0.117\linewidth,trim={0 0 0 0}]{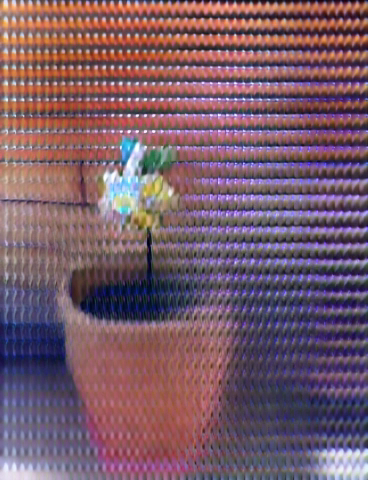} &
        \adjincludegraphics[clip,width=0.088\linewidth,height=0.117\linewidth,trim={0 0 0 0}]{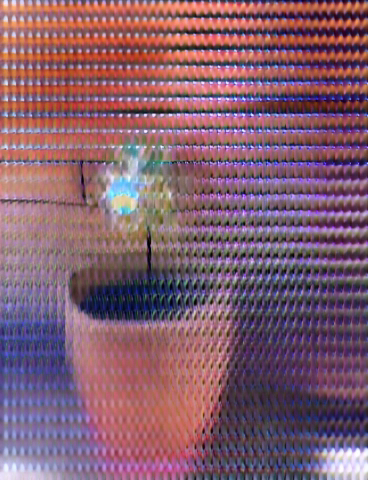} &
        \adjincludegraphics[clip,width=0.088\linewidth,height=0.117\linewidth,trim={0 0 0 0}]{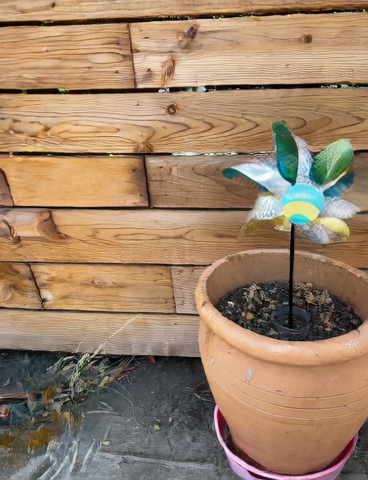} &
        \adjincludegraphics[clip,width=0.088\linewidth,height=0.117\linewidth,trim={0 0 0 0}]{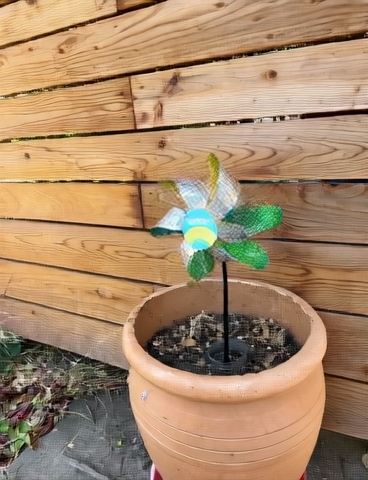} &
        \adjincludegraphics[clip,width=0.088\linewidth,height=0.117\linewidth,trim={0 0 0 0}]{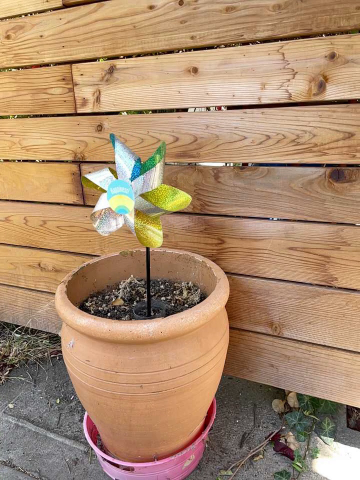} \\

        \adjincludegraphics[clip,width=0.088\linewidth,height=0.117\linewidth,trim={0 0 0 0}]{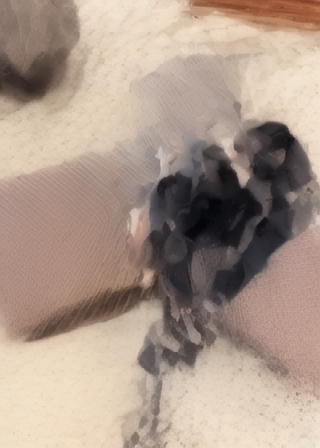} &
        \adjincludegraphics[clip,width=0.088\linewidth,height=0.117\linewidth,trim={0 0 0 0}]{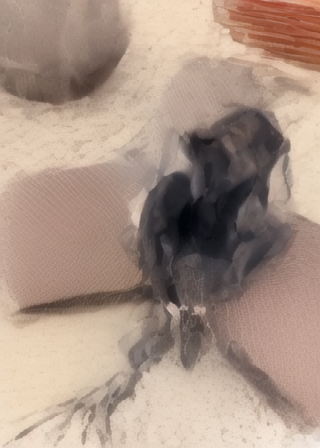} &
        \adjincludegraphics[clip,width=0.088\linewidth,height=0.117\linewidth,trim={0 0 0 0}]{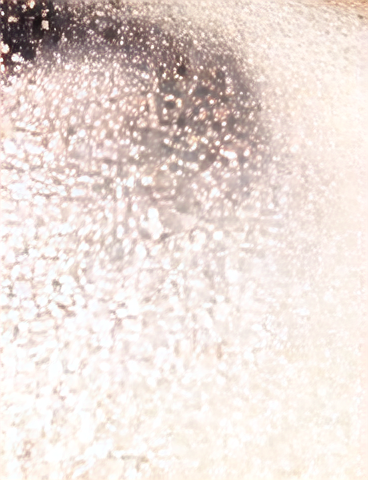} &
        \adjincludegraphics[clip,width=0.088\linewidth,height=0.117\linewidth,trim={0 0 0 0}]{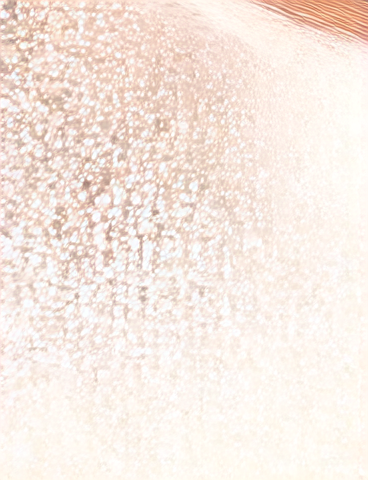} &
        \adjincludegraphics[clip,width=0.088\linewidth,height=0.117\linewidth,trim={0 0 0 0}]{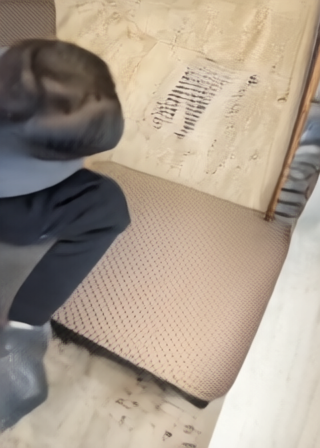} &
        \adjincludegraphics[clip,width=0.088\linewidth,height=0.117\linewidth,trim={0 0 0 0}]{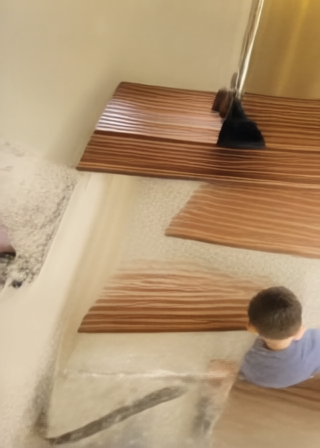} &
        \adjincludegraphics[clip,width=0.088\linewidth,height=0.117\linewidth,trim={0 0 0 0}]{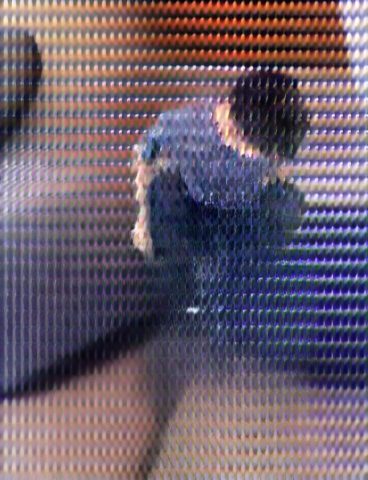} &
        \adjincludegraphics[clip,width=0.088\linewidth,height=0.117\linewidth,trim={0 0 0 0}]{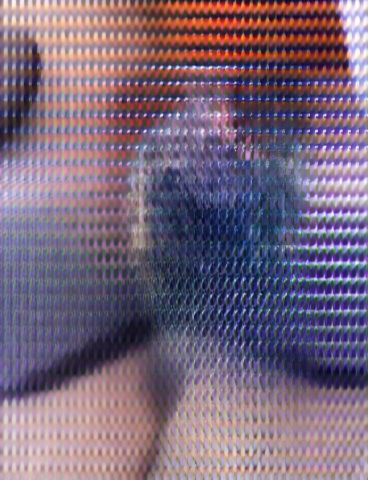} &
        \adjincludegraphics[clip,width=0.088\linewidth,height=0.117\linewidth,trim={0 0 0 0}]{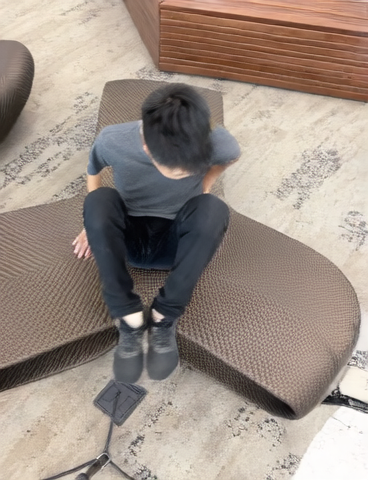} &
        \adjincludegraphics[clip,width=0.088\linewidth,height=0.117\linewidth,trim={0 0 0 0}]{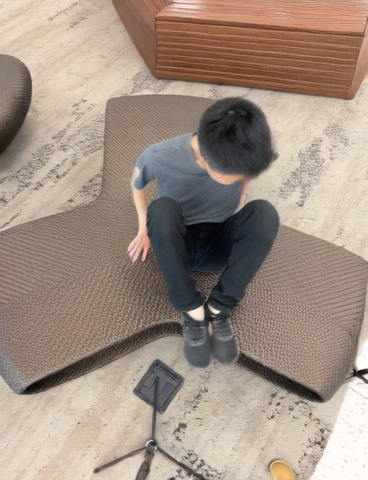} &
        \adjincludegraphics[clip,width=0.088\linewidth,height=0.117\linewidth,trim={0 0 0 0}]{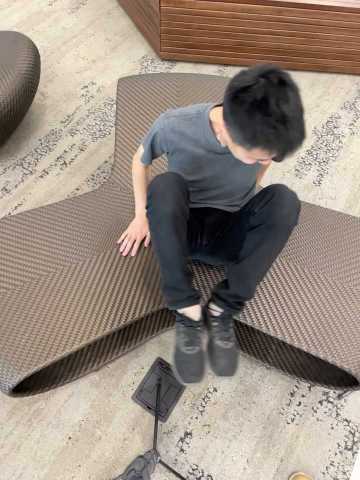} \\
         \multicolumn{2}{c}{\small GCD~\cite{van2024generative}} & \multicolumn{2}{c}{\small ReCamMaster~\cite{Bai_2025_ICCV}} & \multicolumn{2}{c}{\small TrajectoryCrafter~\cite{xiao2025trajectory}} & \multicolumn{2}{c}{\small CogNVS~\cite{chen2025cognvs}} & \multicolumn{2}{c}{\small Ours} & \multicolumn{1}{c}{\small GT}
    \end{tabular}
    \vspace{\abovefigcapmargin}
    \caption{\textbf{Qualitative results on the iPhone dataset~\cite{gao2022dynamic}.} For each method, the left and right columns are retakes with 161 and 241 frames, respectively. All previous methods degrade severely on long videos, whereas our method remains robust and produces high-quality retakes.}
    \vspace{\belowfigcapmargin}
    \label{fig:iphone}
\end{figure*}

\begin{table}[t!]
    \centering
    \setlength\tabcolsep{3pt}
    \resizebox{\linewidth}{!}{
    \begin{tabular}{lccccccc}
       \toprule
        \multirow{2}{*}{Method}  & \multicolumn{2}{c}{81 frame} & \multicolumn{2}{c}{161 frame} & \multicolumn{2}{c}{241 frame}  \\
        \arrayrulecolor{gray}\cmidrule(lr){2-3} \cmidrule(lr){4-5} \cmidrule(lr){6-7}
        & PSNR$\uparrow$ & LPIPS$\downarrow$ & PSNR$\uparrow$ & LPIPS$\downarrow$ & PSNR$\uparrow$ & LPIPS$\downarrow$  \\
       \arrayrulecolor{black} \midrule
        GCD~\cite{van2024generative}  & 9.56 & \cellcolor{tabthird} 0.697 & 9.63 & 0.758 & 9.41 & 0.765 \\
        ReCamMaster~\cite{Bai_2025_ICCV}  & \cellcolor{tabsecond} 10.69 & \cellcolor{tabsecond} 0.678 & \cellcolor{tabthird} 10.03 & 0.762 & \cellcolor{tabthird} 10.37 & 0.772 \\
        TrajectoryCrafter~\cite{Yu_2025_ICCV}  & 8.79 & 0.733 & 9.58 & \cellcolor{tabthird} 0.744 &  10.35 & \cellcolor{tabthird} 0.741 \\
        CogNVS$^\dagger$~\cite{chen2025cognvs}  & \cellcolor{tabthird} 10.56 & 0.720 & \cellcolor{tabsecond} 10.63 & \cellcolor{tabsecond} 0.741 & \cellcolor{tabsecond} 10.81 & \cellcolor{tabsecond} 0.720 \\
        \arrayrulecolor{gray}\midrule
        Ours (ReDirector) & \cellcolor{tabfirst} 10.82 & \cellcolor{tabfirst} 0.655 & \cellcolor{tabfirst} 11.56 & \cellcolor{tabfirst} 0.631 & \cellcolor{tabfirst} 11.85 & \cellcolor{tabfirst} 0.611 \\
       \arrayrulecolor{black}\bottomrule
    \end{tabular}
    }
    \vspace{-1mm}
    \vspace{\abovetabcapmargin}
    \caption{\textbf{Quantitative results on iPhone dataset~\cite{gao2022dynamic}.} CogNVS$^\dagger$ uses MoSca~\cite{lei2024mosca} and LiDAR depths to align scene scales.}
    \vspace{\belowtabcapmargin}
    \vspace{-3mm}
    \label{tab:iphone_main}
\end{table}

\vspace{\paramargin}
\paragraph{Generalization to out-of-distribution conditions.}
Our method seamlessly integrates the camera control signal into RoPE phase shifts, and geometry-aware attention further enhances the multi-view consistency. To assess effectiveness, we evaluate the dynamic novel view synthesis performance~\cite{lei2024mosca, gao2021dynamic, li2023dynibar, park2024point}, confirming generalization to out-of-distribution trajectories, video lengths, and resolutions.

\Cref{fig:iphone} shows that GCD~\cite{van2024generative} and ReCamMaster~\cite{Bai_2025_ICCV} often produce blurry, speckled artifacts and fail at novel view synthesis on long video sequences.
TrajectoryCrafter~\cite{Yu_2025_ICCV} struggles to infer scene scale without LiDAR-based metric depth alignment, due to its reliance on hand-crafted heuristics in the warping process.
CogNVS~\cite{chen2025cognvs}, despite using a strong 4D reconstruction method~\cite{lei2024mosca} and LiDAR-based metric depth alignment, frequently fails to refine and can even degrade video quality. In contrast, our method remains robust on long sequences and produces realistic novel views without ground-truth metric depth or external geometry models. As longer input videos naturally cover larger portions of the scene, our approach exploits this to reconstruct additional regions (\eg, the black stand in the first example) and to recover scene scales that align more closely with the ground truth. \Cref{tab:iphone_main} further shows
that our method achieves the best novel view synthesis performance, with consistent improvements as input video length increases.
 
\begin{table}[t!]
    \centering
    \setlength\tabcolsep{8pt}
    \resizebox{\linewidth}{!}{
    \begin{tabular}{ccccc}
       \toprule
        \multirow{2}{*}{Train Iter} & \multicolumn{2}{c}{Geometric Consistency} & \multicolumn{2}{c}{Camera Accuracy}  \\
        \arrayrulecolor{gray}\cmidrule(lr){2-3} \cmidrule(lr){4-5}
        & Dyn-MEt3R$\uparrow$ & MEt3R$\downarrow$ & TransErr$\downarrow$ & RotErr$\downarrow$  \\
        \midrule
        20K & \cellcolor{tabsecond} 0.8477 & \cellcolor{tabsecond} 0.3073 & \cellcolor{tabsecond} 0.0165 & \cellcolor{tabsecond} 1.666 \\
        50K & \cellcolor{tabfirst} 0.8491 & \cellcolor{tabfirst} 0.2954 & \cellcolor{tabfirst} 0.0154 & \cellcolor{tabfirst} 1.521 \\
       \arrayrulecolor{black}\bottomrule
    \end{tabular}
    }
    \vspace{\abovetabcapmargin}
    \caption{\textbf{Quantitative ablations of training iterations on the DAVIS dataset~\cite{pont20172017}.} We observe that additional training consistently improves the geometric consistency and camera accuracy.}
    \label{tab:further}
    \vspace{\belowtabcapmargin}
\end{table}

\vspace{\paramargin}
\paragraph{Effect of additional training iterations.}
We investigate the effect of additional training iterations in \Cref{tab:further}. The results show steady gains in camera accuracy and geometric consistency, suggesting that further training progressively strengthens the model to internalize multi-view geometry purely from data, without explicit inductive biases. 
\section{Conclusion}
\label{sec:conclusion}

In this paper, we have introduced ReDirector, a camera-controlled video retake generation method for dynamically captured variable-length input videos. We first rectify a prevalent misuse of RoPE and camera encodings in prior work. Specifically, we apply the same RoPE for input and target videos, and use Plücker rays as camera pose representations, enabling length-agnostic conditioning and encoding token-level multi-view relationships. Then, we introduce RoCE, a camera-conditioned RoPE phase shift with geometry-aware attention, which improves geometric consistency, better dynamic object localization, and static background preservation, even when the input video contains dynamic camera motion.
Extensive experiments verify the effectiveness of our method in camera control, geometric consistency, and visual quality, with strong generalization to out-of-distribution trajectories, lengths, and resolutions.

{
    \small
    \bibliographystyle{ieeenat_fullname}
    \bibliography{main}
}

\clearpage \appendix

\maketitlesupplementary

\begin{table}[t!]
    \centering
    \setlength\tabcolsep{5pt}
    \resizebox{\linewidth}{!}{
    \begin{tabular}{lcccc}
       \toprule
        \multirow{2}{*}{Camera Pose} & \multicolumn{2}{c}{Geometric Consistency} & \multicolumn{2}{c}{Camera Accuracy}  \\
        \arrayrulecolor{gray}\cmidrule(lr){2-3} \cmidrule(lr){4-5}
        & Dyn-MEt3R$\uparrow$ & MEt3R$\downarrow$ & TransErr$\downarrow$ & RotErr$\downarrow$  \\
        \midrule
        MegaSaM~\cite{li2025megasam} & \cellcolor{tabsecond} 0.8357 & \cellcolor{tabfirst} 0.3041 & \cellcolor{tabsecond} 0.0210 & \cellcolor{tabsecond} 1.942  \\
        ViPE~\cite{huang2025vipe} & \cellcolor{tabfirst} 0.8477 & \cellcolor{tabsecond} 0.3073 & \cellcolor{tabfirst} 0.0165 & \cellcolor{tabfirst} 1.666 \\
       \arrayrulecolor{black}\bottomrule
    \end{tabular}
    }
    \vspace{\abovetabcapmargin}
    \caption{\textbf{Effect of input trajectories on the DAVIS dataset~\cite{pont20172017}.} ViPE~\cite{huang2025vipe} and MegaSaM~\cite{li2025megasam} approximately take 30 seconds and 3 minutes for a hundred-length video, respectively.}
    \label{tab:abl_pose}
    \vspace{\belowtabcapmargin}
\end{table}

\section{Evaluation Details}

For all experiments, we set the inference steps to 50 and the CFG scale~\cite{ho2022classifier} to 5. We use BLIP~\cite{li2022blip} for text captioning.

\vspace{\paramargin}
\paragraph{DAVIS dataset.} We use the camera intrinsics estimated by ViPE~\cite{huang2025vipe} when converting the target camera trajectories into Plücker rays. Since the 10 test camera trajectories in ReCamMaster~\cite{Bai_2025_ICCV} are originally defined over 81 frames, we reparameterize them as per-frame updates. Specifically, for each trajectory, we first accumulate the original 81-frame updates into a single total rotation and translation. Then, given an input video with $F$ frames, we divide this total camera motion by $F-1$ to obtain constant per-frame updates. This per-frame parameterization allows us to apply the same camera patterns to videos of arbitrary length in our experimental setting. The resulting per-frame camera trajectories are detailed below:

\vspace{2mm}
\begin{itemize}[itemsep=0.3em]
    \item \textbf{Pan right/left:} The camera rotates by $\frac{20}{F-1}$ degrees about the yaw axis per frame, with no translation.
    \item \textbf{Tilt up/down:} The camera rotates by $\frac{10}{F-1}$ degrees about the pitch axis per frame, with no translation.
    \item \textbf{Zoom in/out:} The camera translates by $\frac{2}{F-1}$(m) along the z-axis per frame, with no rotation.
    \item \textbf{Translate up/down:} For each frame, the camera tilts down/up by $\frac{14}{F-1}$ degrees and translates $\frac{1}{F-1}$(m) along the y-axis (\ie, up/down), respectively; in both cases, it additionally moves $\frac{0.12}{F-1}$(m) outward (\ie, zoom out).
    \item \textbf{Arc left/right:} For each frame, the camera pans right/left by $\frac{30}{F-1}$ degrees and translates $\frac{2}{F-1}$(m) along the x-axis (\ie, left/right), respectively; in both cases, it additionally moves $\frac{0.01}{F-1}$(m) inward (\ie, zoom in).
\end{itemize}
\vspace{2mm}

\vspace{\paramargin}
\paragraph{iPhone dataset.} We do not use any video pose estimation methods or other external models. We rely solely on the provided camera trajectories for each input video and the specified viewpoint for each test case. Even without external models and LiDAR depth, ReDirector achieves the best performance in dynamic novel view synthesis compared to previous video retake generation methods.

\begin{figure}[t!]
    \centering
    \includegraphics[width=\linewidth]{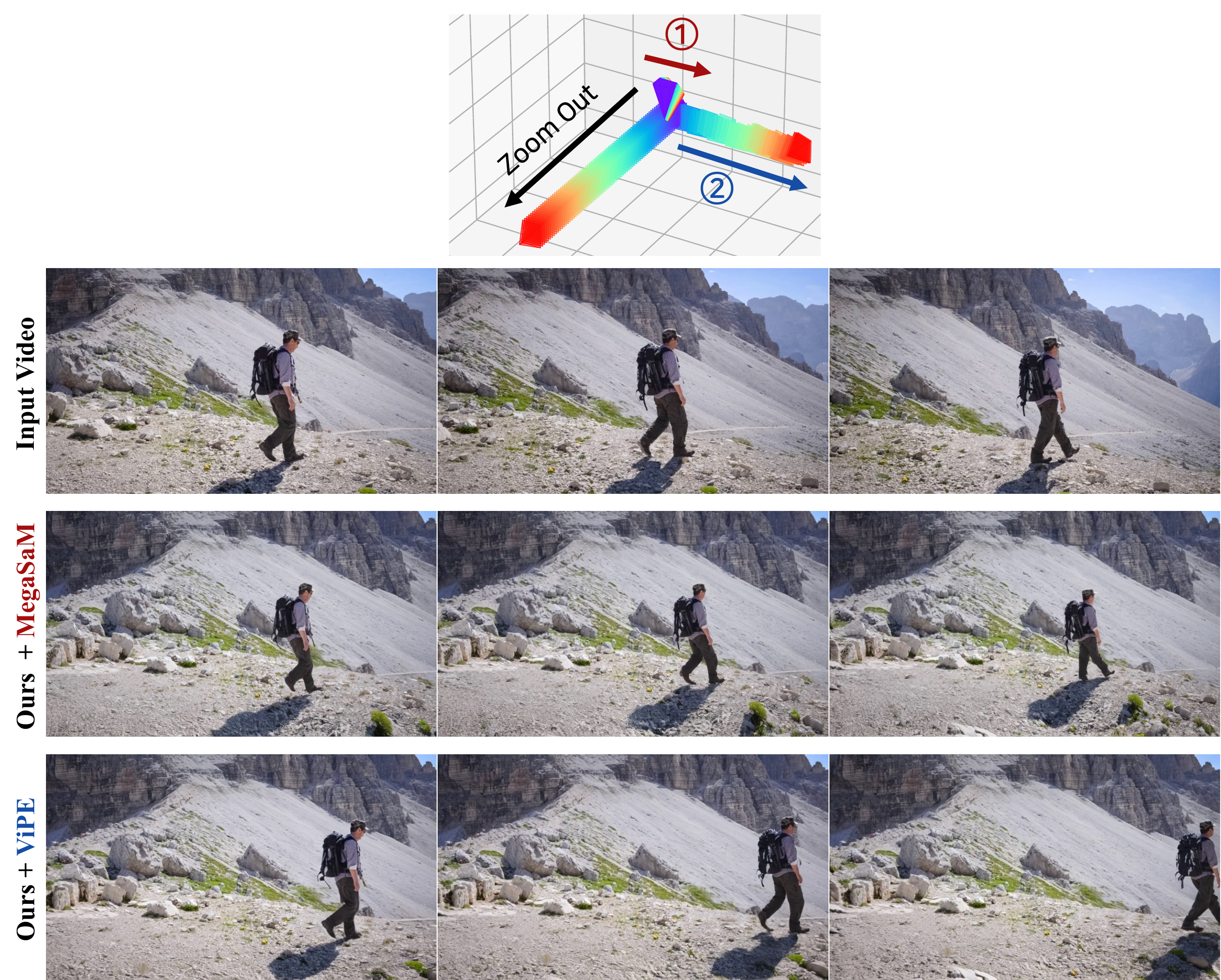}
    \vspace{\abovefigcapmargin}
    \vspace{-4.5mm}
    \caption{\textbf{Qualitative ablation on input trajectories.}}
    \vspace{\belowfigcapmargin}
    \label{fig:megasam}
\end{figure}

\begin{figure}[t!]
    \centering
    \setlength\tabcolsep{0.1pt}
    
    \begin{tabular}{@{}c@{\,}ccc@{}}
        \raisebox{0.015\linewidth}{\rotatebox{90}{\footnotesize Input Video}} &
        \adjincludegraphics[clip,width=0.32\linewidth,trim={0 0 0 0}]{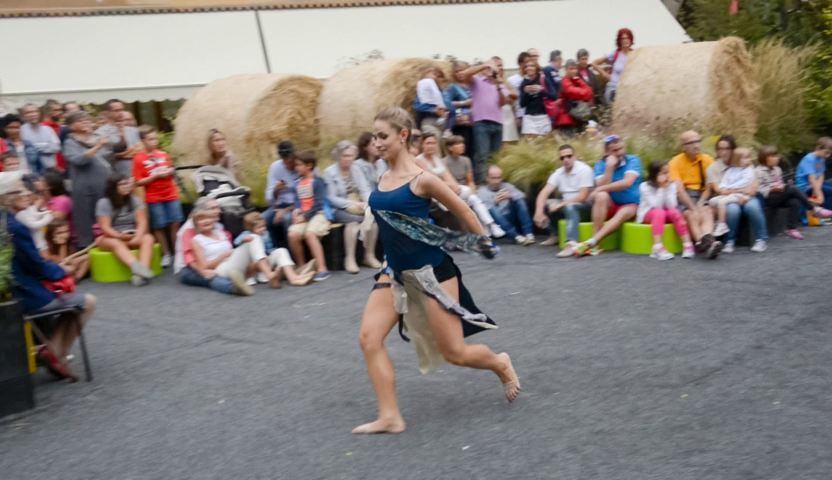} &
        \adjincludegraphics[clip,width=0.32\linewidth,trim={0 0 0 0}]{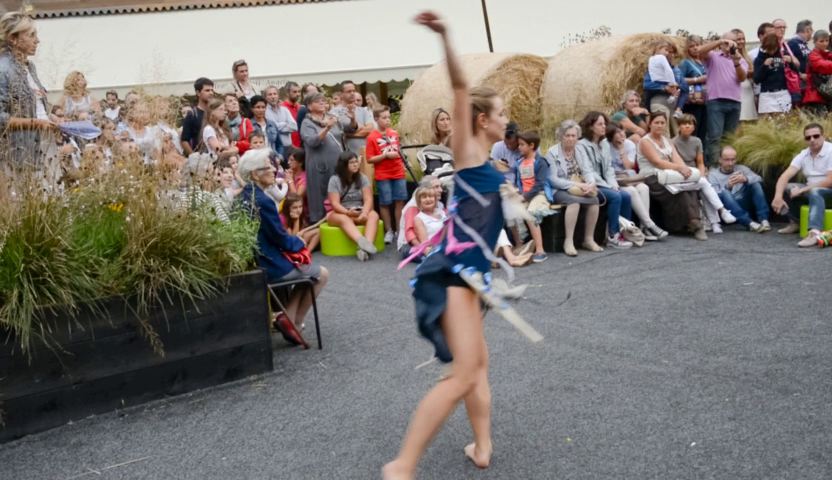} &
        \adjincludegraphics[clip,width=0.32\linewidth,trim={0 0 0 0}]{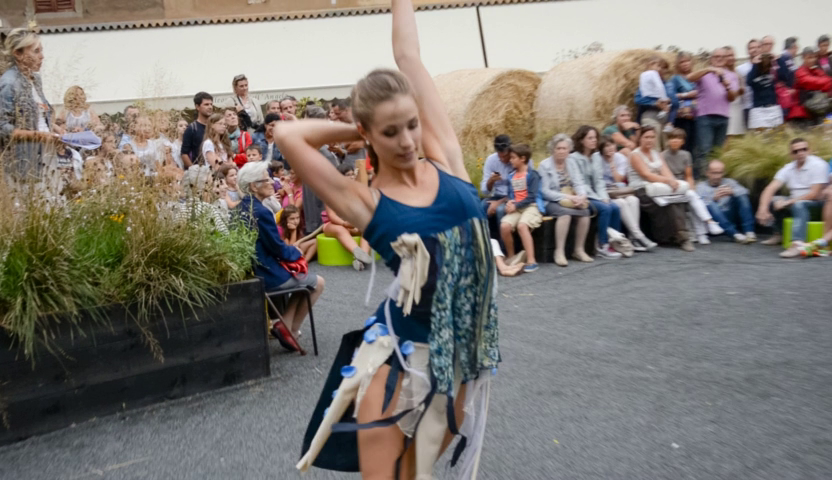} \\

        \raisebox{0.015\linewidth}{\rotatebox{90}{\footnotesize Ours (20K)}} &
        \adjincludegraphics[clip,width=0.32\linewidth,trim={0 0 0 0}]{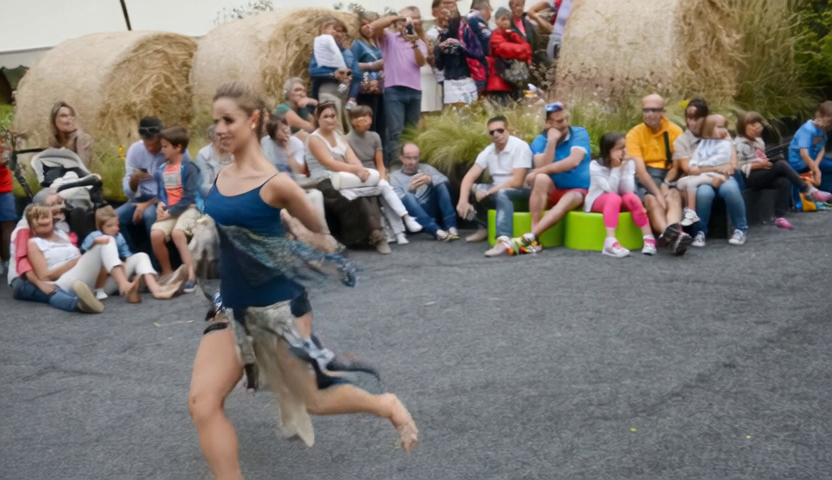} &
        \adjincludegraphics[clip,width=0.32\linewidth,trim={0 0 0 0}]{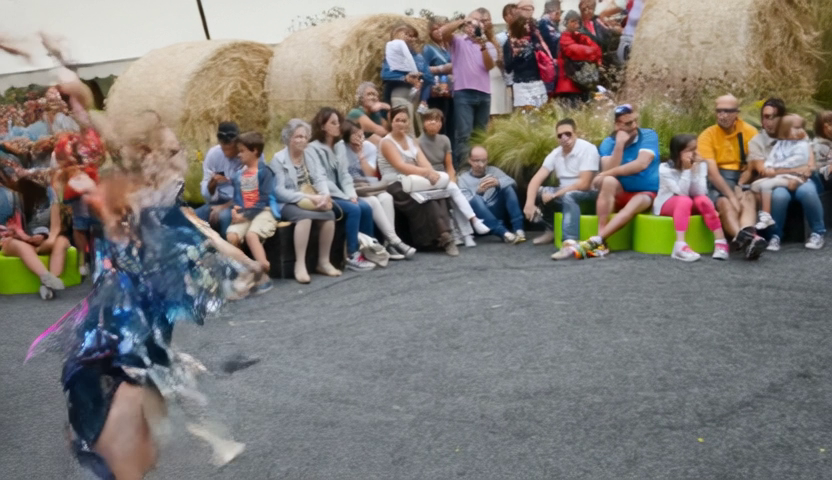} &
        \adjincludegraphics[clip,width=0.32\linewidth,trim={0 0 0 0}]{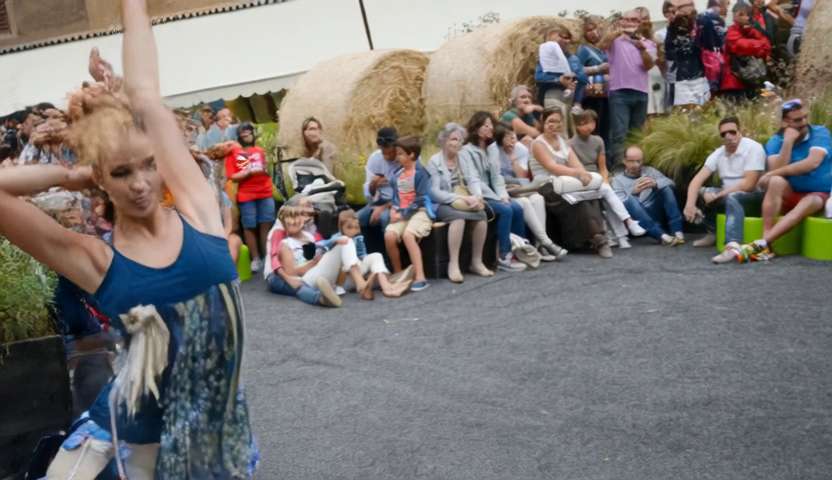} \\

        \raisebox{0.015\linewidth}{\rotatebox{90}{\footnotesize Ours (50K)}} &
        \adjincludegraphics[clip,width=0.32\linewidth,trim={0 0 0 0}]{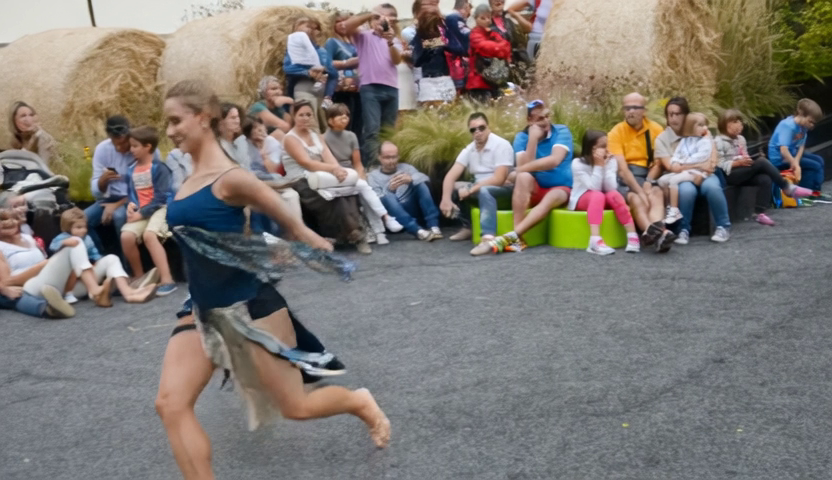} &
        \adjincludegraphics[clip,width=0.32\linewidth,trim={0 0 0 0}]{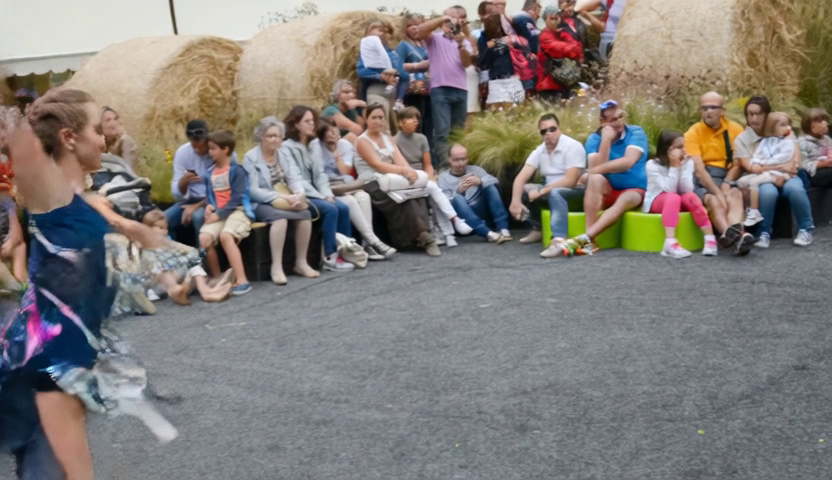} &
        \adjincludegraphics[clip,width=0.32\linewidth,trim={0 0 0 0}]{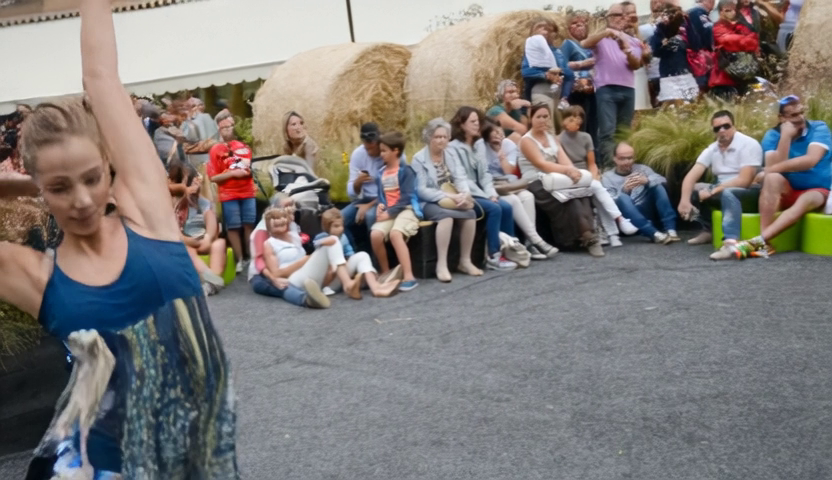} \\
    \end{tabular}
    \vspace{-2mm}
    \vspace{\abovefigcapmargin}
    \caption{\textbf{Qualitative ablation on training iterations.}}
    \vspace{\belowfigcapmargin}
    \vspace{-0.5mm}
    \label{fig:more_training}
\end{figure}

\section{Additional Results}
\label{sec:more_results}

\subsection{Ablative Results}

\paragraph{Effect of input trajectories.}
We evaluate the impact of input camera trajectories on retake performance by replacing ViPE with MegaSaM~\cite{li2025megasam}. ViPE predicts in metric scale, whereas MegaSaM operates in a scale-ambiguous space and provides more accurate camera poses with an optimization process. Because the target camera trajectories are also defined in metric scale, \Cref{tab:abl_pose} shows that metrically aligned poses yield better performance than more accurate but scale-indefinite poses. Consistently, \Cref{fig:megasam} demonstrates that poses from ViPE produce a geometrically accurate video retake, whereas poses from MegaSaM cause the model to fail to properly localize the person.

\vspace{\paramargin}
\paragraph{Effect of further training.}
As shown in~\cref{fig:more_training}, we qualitatively verify that additional training improves both geometric consistency and dynamic object localization.

\begin{figure*}[t!]
    \centering
    \setlength\tabcolsep{0.2pt}
    \renewcommand{\arraystretch}{0.1}
    \begin{tabular}{@{}c@{\,}cccccc@{}}

        \raisebox{0.02\linewidth}{\rotatebox{90}{\scriptsize Camera}} & \multicolumn{3}{c}{\adjincludegraphics[clip,width=0.158\linewidth,height=0.0911\linewidth,trim={0 0 0 0}]{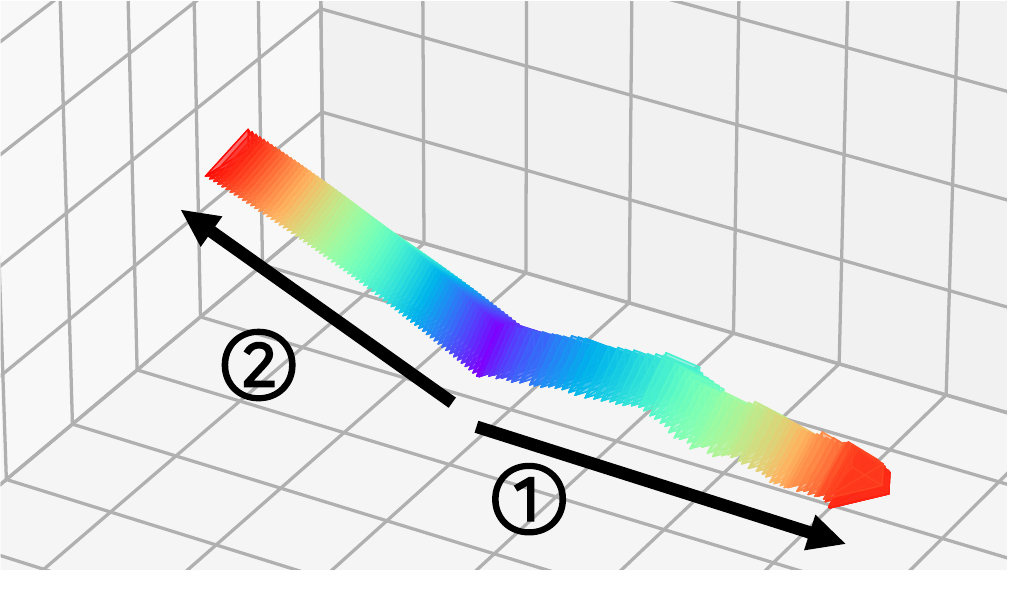}} &
        & \adjincludegraphics[clip,width=0.158\linewidth,height=0.0911\linewidth,trim={0 0 0 0}]{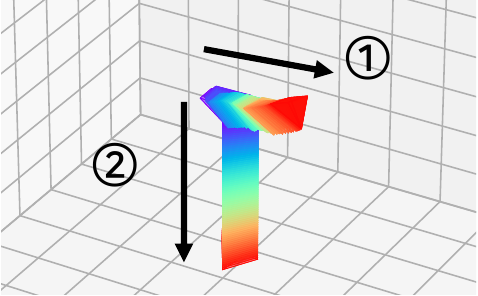} &  \\
    
        \raisebox{0.01\linewidth}{\rotatebox{90}{\scriptsize Input Video}} &
        \adjincludegraphics[clip,width=0.158\linewidth,trim={0 0 0 0}]{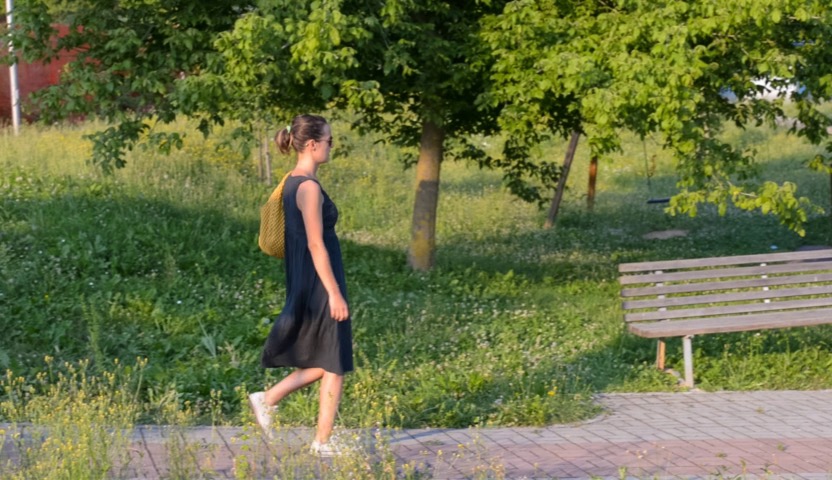} &
        \adjincludegraphics[clip,width=0.158\linewidth,trim={0 0 0 0}]{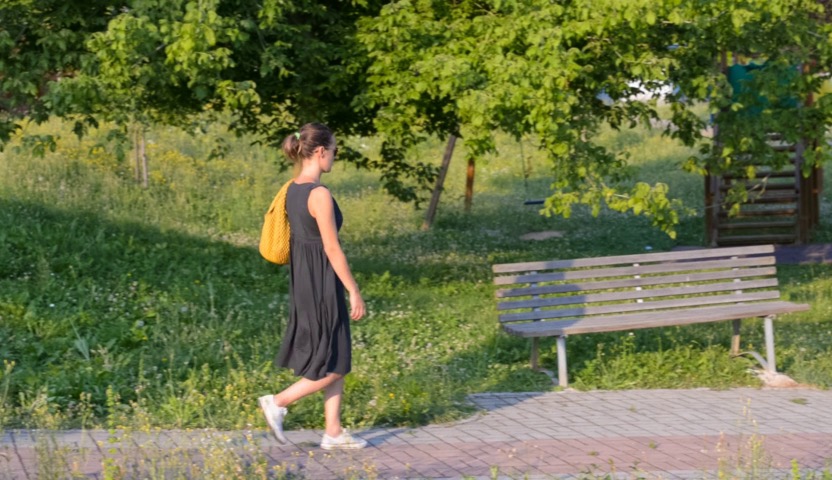} &
        \adjincludegraphics[clip,width=0.158\linewidth,trim={0 0 0 0}]{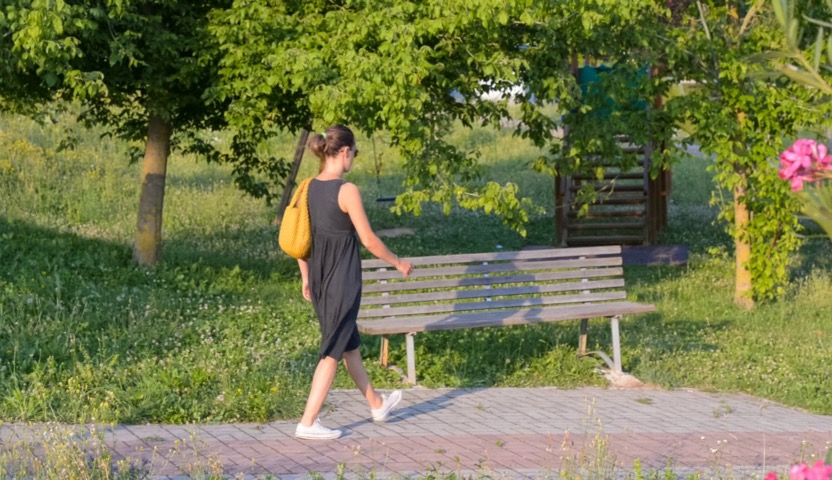} &
        \adjincludegraphics[clip,width=0.158\linewidth,trim={0 0 0 0}]{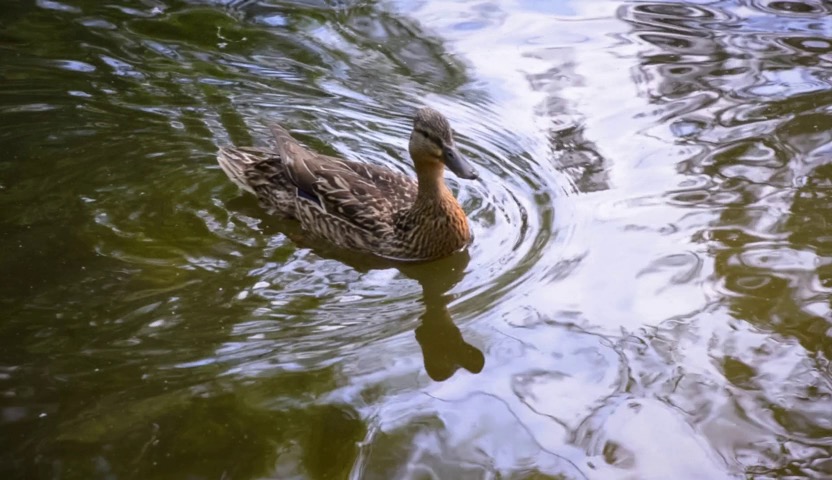} &
        \adjincludegraphics[clip,width=0.158\linewidth,trim={0 0 0 0}]{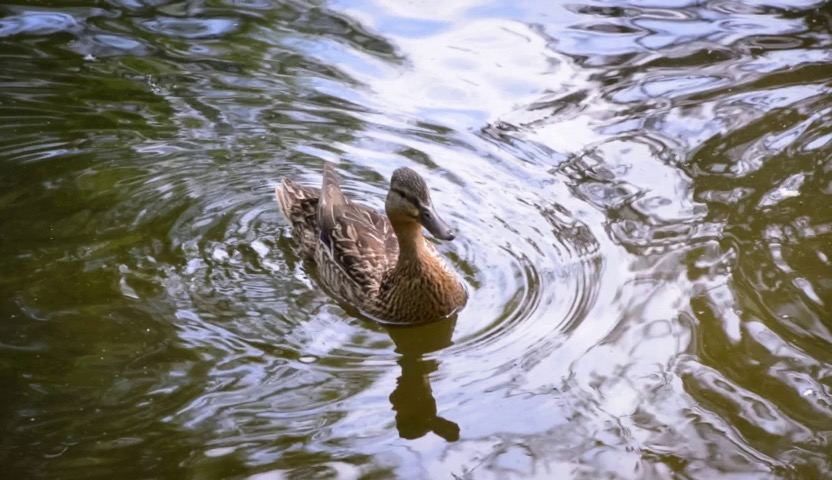} &
        \adjincludegraphics[clip,width=0.158\linewidth,trim={0 0 0 0}]{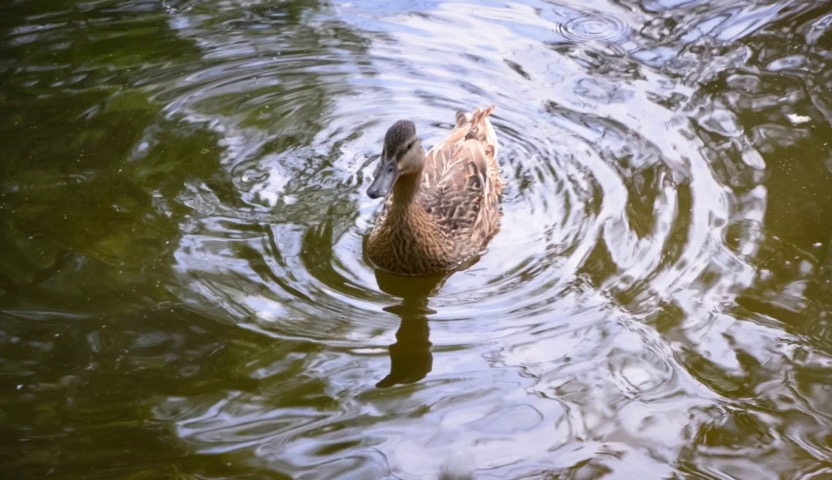} \\

        \raisebox{0.03\linewidth}{\rotatebox{90}{\scriptsize GCD}} &
        \adjincludegraphics[clip,width=0.158\linewidth, height=0.0911\linewidth, trim={0 0 0 0}]{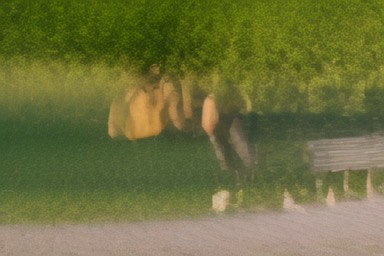} &
        \adjincludegraphics[clip,width=0.158\linewidth, height=0.0911\linewidth, trim={0 0 0 0}]{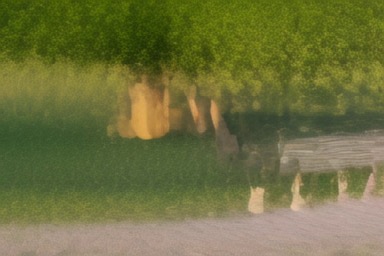} &
        \adjincludegraphics[clip,width=0.158\linewidth, height=0.0911\linewidth, trim={0 0 0 0}]{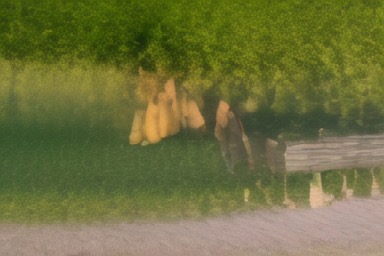} &
        \adjincludegraphics[clip,width=0.158\linewidth, height=0.0911\linewidth, trim={0 0 0 0}]{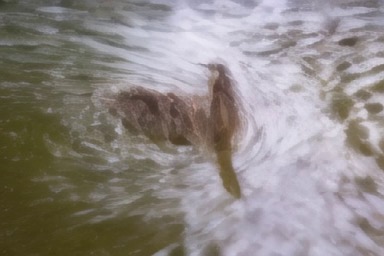} &
        \adjincludegraphics[clip,width=0.158\linewidth, height=0.0911\linewidth, trim={0 0 0 0}]{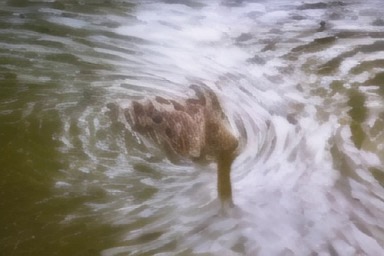} &
        \adjincludegraphics[clip,width=0.158\linewidth, height=0.0911\linewidth, trim={0 0 0 0}]{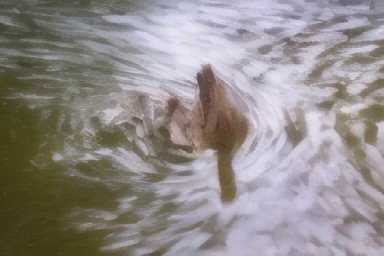} \\

        \raisebox{0.005\linewidth}{\rotatebox{90}{\scriptsize ReCamMaster}} &
        \adjincludegraphics[clip,width=0.158\linewidth,trim={0 0 0 0}]{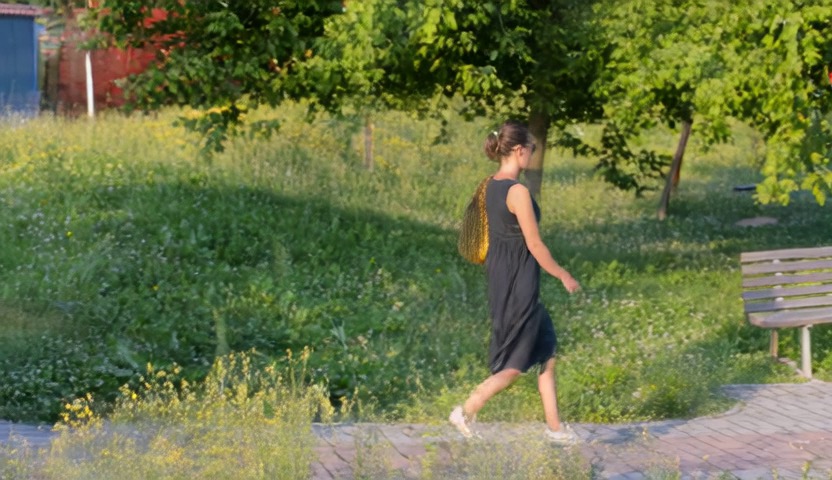} &
        \adjincludegraphics[clip,width=0.158\linewidth,trim={0 0 0 0}]{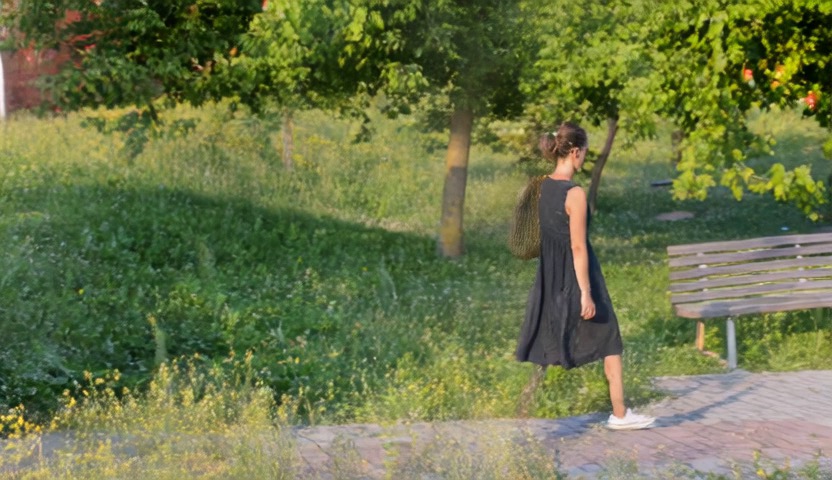} &
        \adjincludegraphics[clip,width=0.158\linewidth,trim={0 0 0 0}]{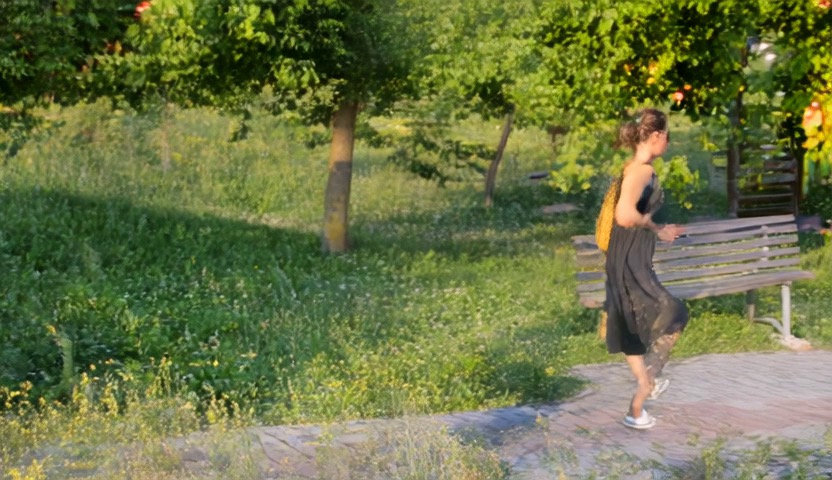} &
        \adjincludegraphics[clip,width=0.158\linewidth,trim={0 0 0 0}]{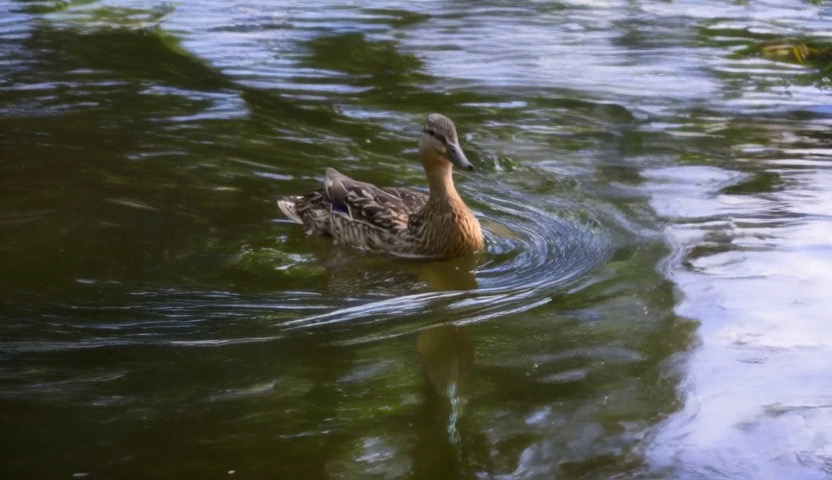} &
        \adjincludegraphics[clip,width=0.158\linewidth,trim={0 0 0 0}]{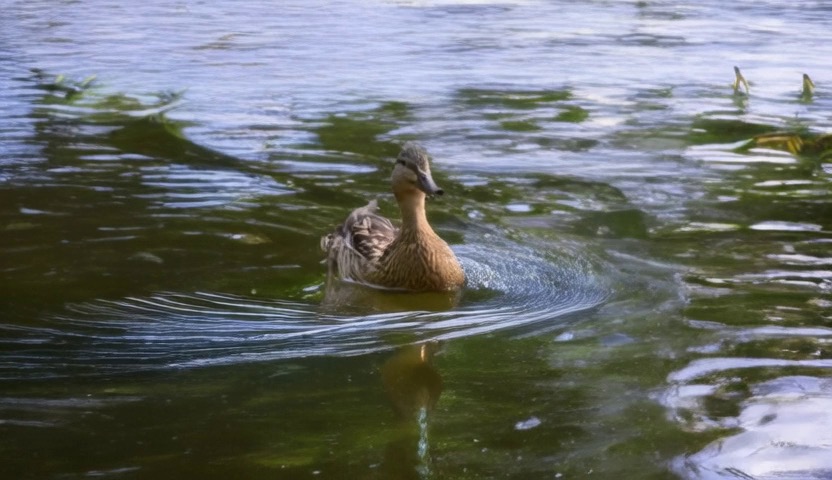} &
        \adjincludegraphics[clip,width=0.158\linewidth,trim={0 0 0 0}]{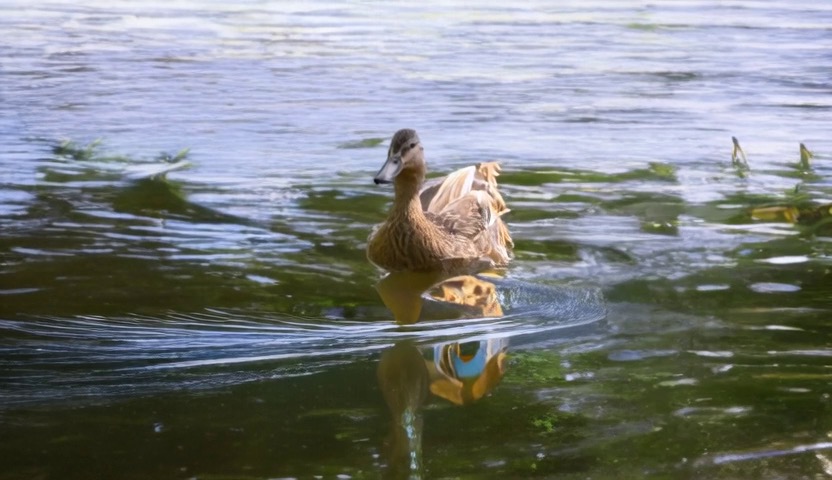} \\

        \raisebox{0.015\linewidth}{\rotatebox{90}{\makecell[c]{\scriptsize Trajectory\\[-4pt]\scriptsize Crafter}}} &
        \adjincludegraphics[clip,width=0.158\linewidth,trim={0 0 0 0}]{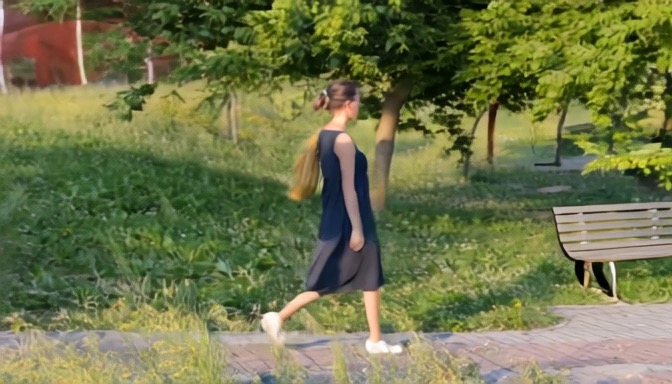} &
        \adjincludegraphics[clip,width=0.158\linewidth,trim={0 0 0 0}]{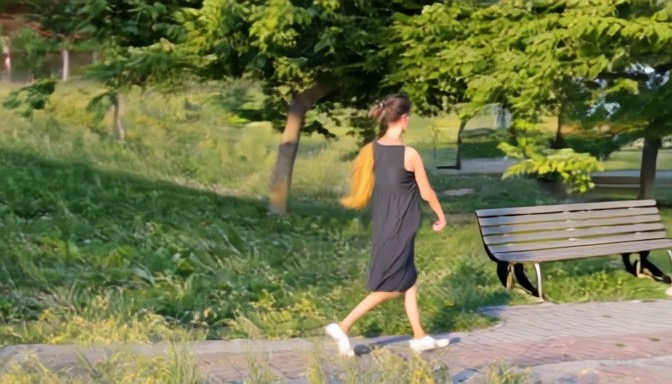} &
        \adjincludegraphics[clip,width=0.158\linewidth,trim={0 0 0 0}]{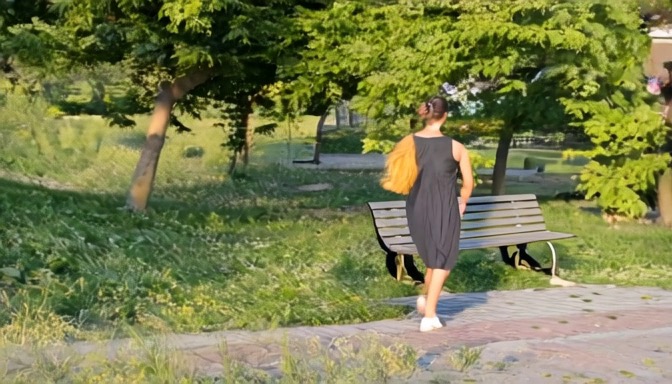} &
        \adjincludegraphics[clip,width=0.158\linewidth,trim={0 0 0 0}]{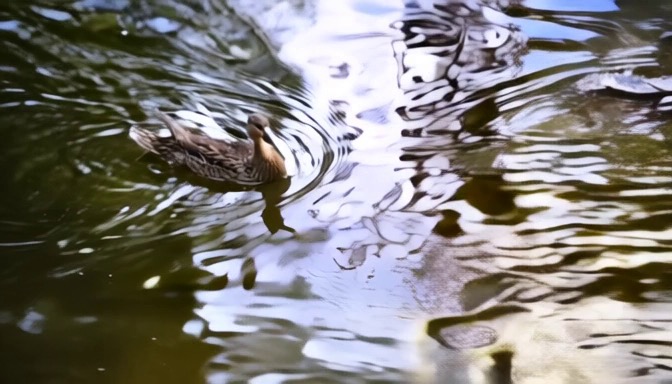} &
        \adjincludegraphics[clip,width=0.158\linewidth,trim={0 0 0 0}]{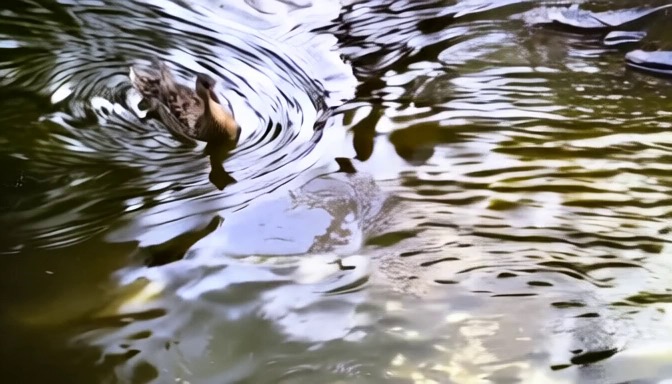} &
        \adjincludegraphics[clip,width=0.158\linewidth,trim={0 0 0 0}]{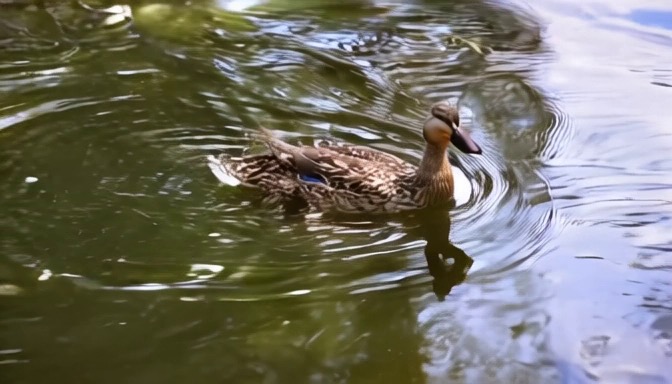} \\

        \raisebox{0.02\linewidth}{\rotatebox{90}{\makecell[c]{\scriptsize CogNVS}}} &
        \adjincludegraphics[clip,width=0.158\linewidth,trim={0 0 0 0}]{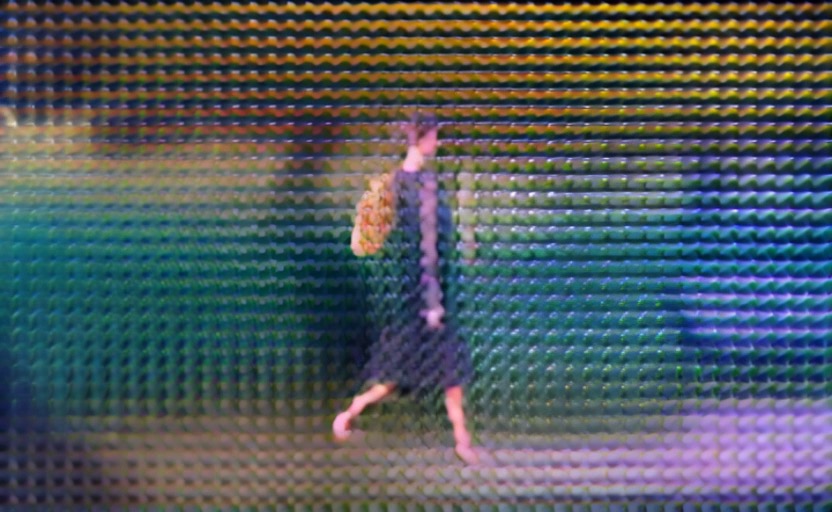} &
        \adjincludegraphics[clip,width=0.158\linewidth,trim={0 0 0 0}]{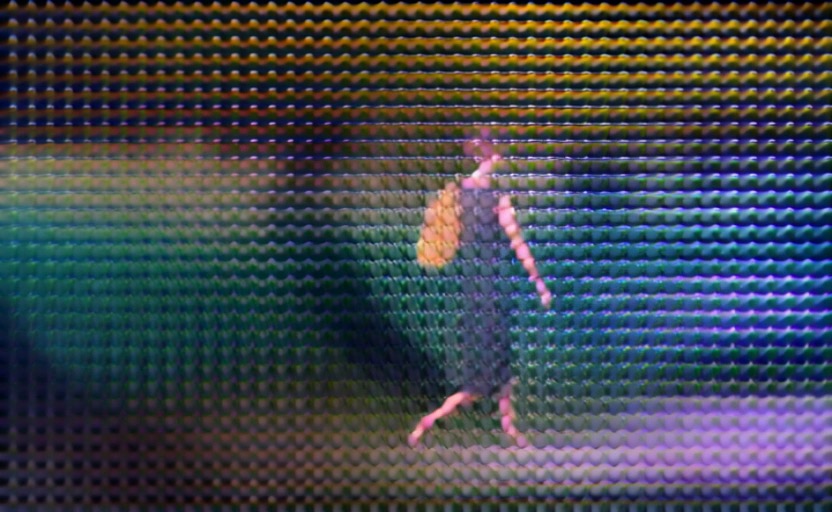} &
        \adjincludegraphics[clip,width=0.158\linewidth,trim={0 0 0 0}]{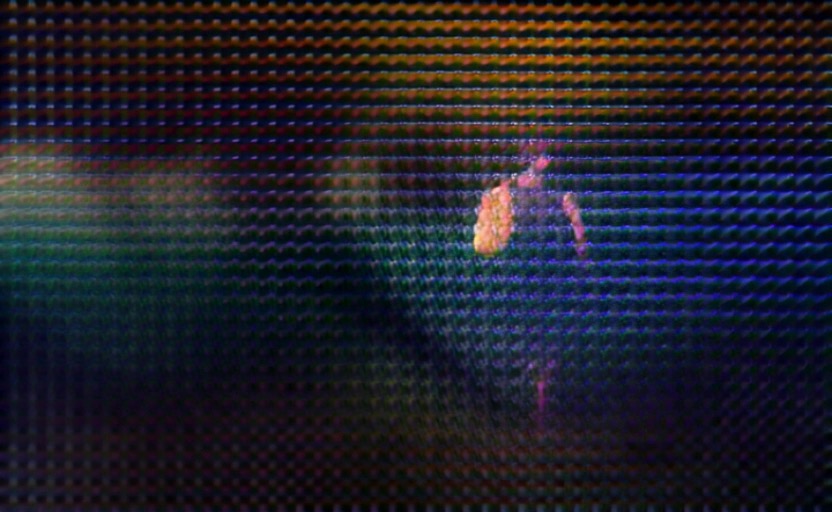} &
        \adjincludegraphics[clip,width=0.158\linewidth,trim={0 0 0 0}]{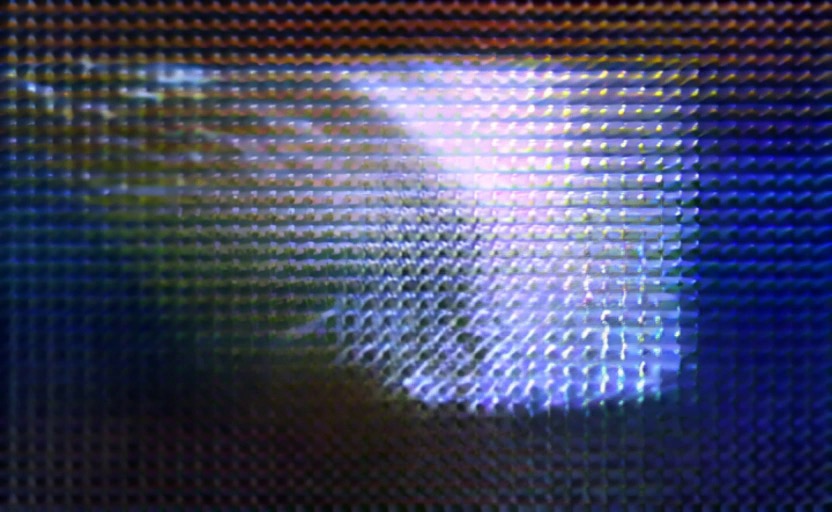} &
        \adjincludegraphics[clip,width=0.158\linewidth,trim={0 0 0 0}]{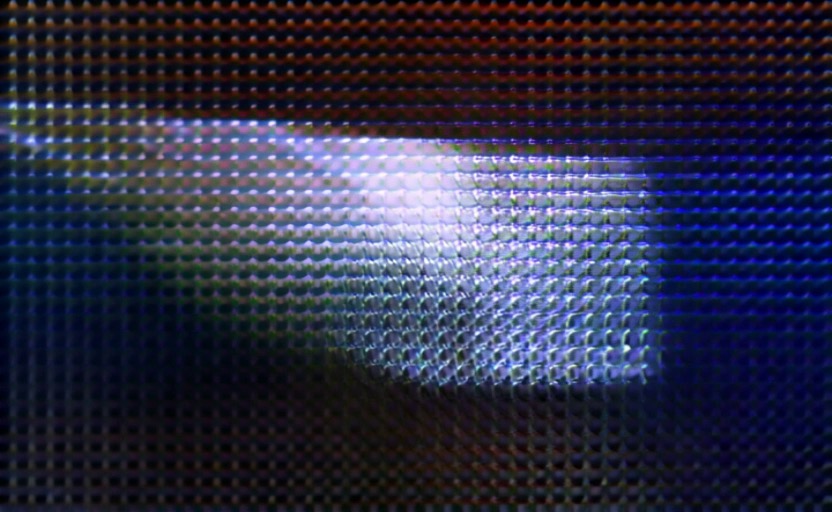} &
        \adjincludegraphics[clip,width=0.158\linewidth,trim={0 0 0 0}]{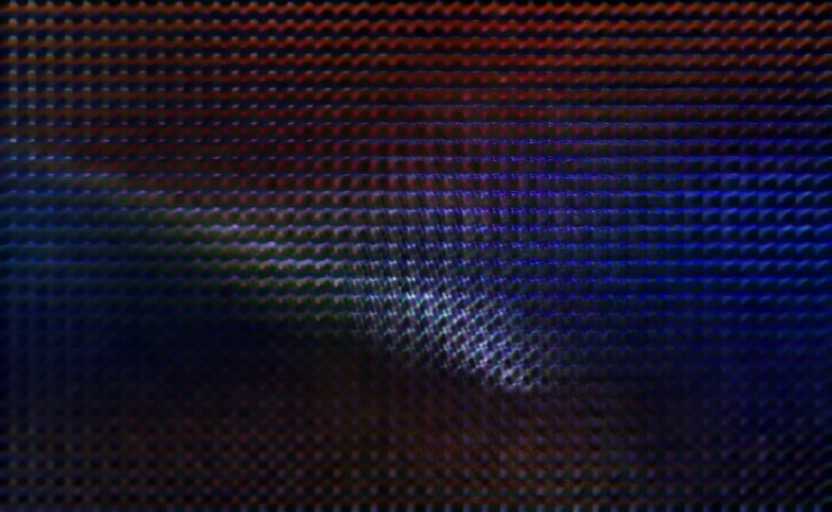} \\
        
        \raisebox{0.03\linewidth}{\rotatebox{90}{\scriptsize Ours}} &
        \adjincludegraphics[clip,width=0.158\linewidth,trim={0 0 0 0}]{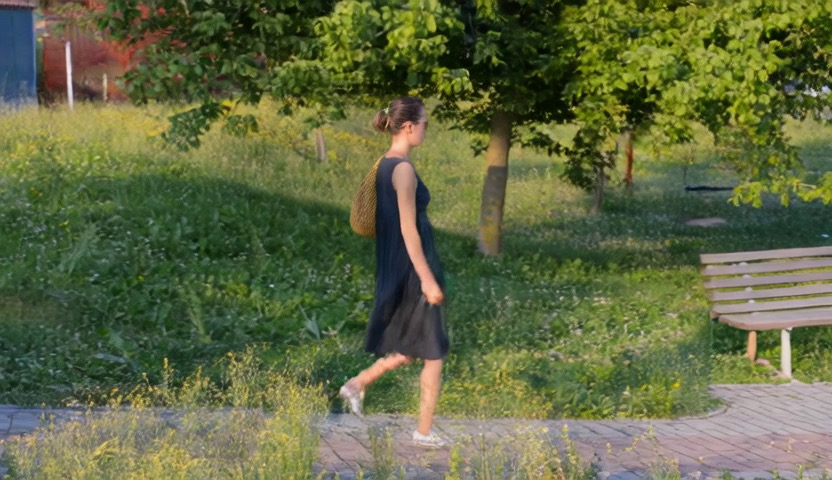} &
        \adjincludegraphics[clip,width=0.158\linewidth,trim={0 0 0 0}]{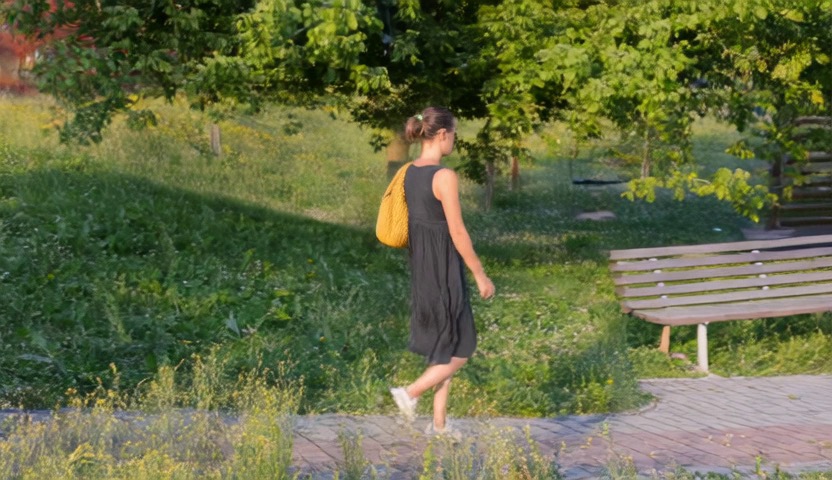} &
        \adjincludegraphics[clip,width=0.158\linewidth,trim={0 0 0 0}]{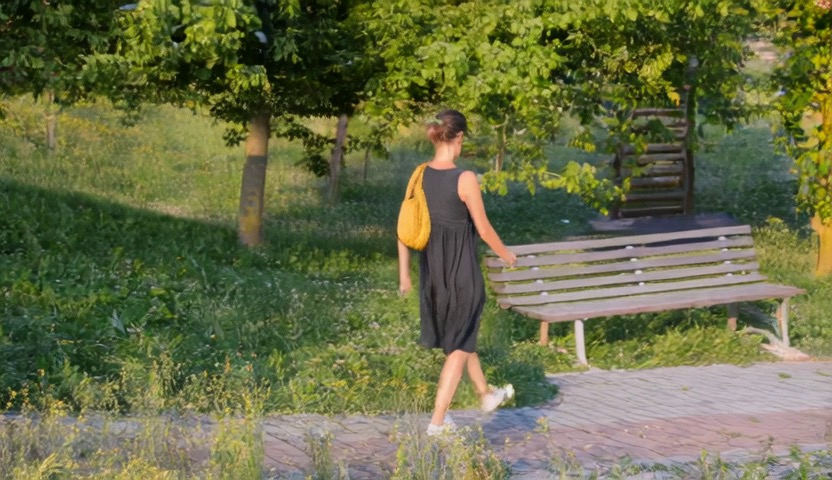} &
        \adjincludegraphics[clip,width=0.158\linewidth,trim={0 0 0 0}]{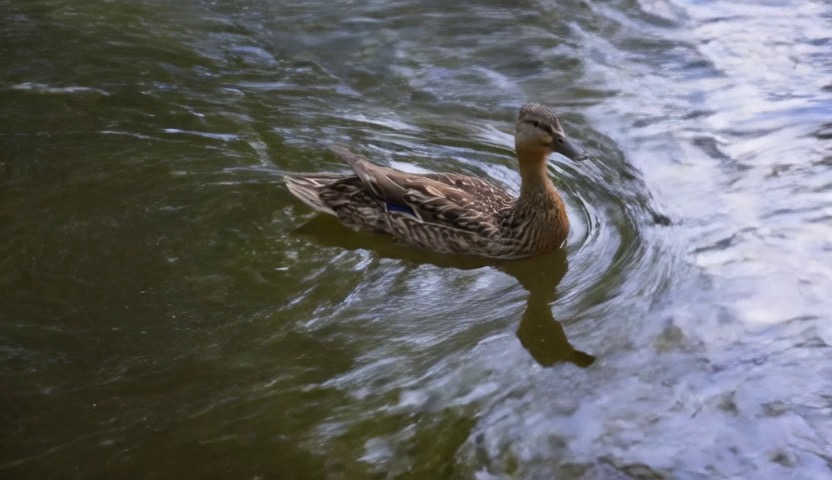} &
        \adjincludegraphics[clip,width=0.158\linewidth,trim={0 0 0 0}]{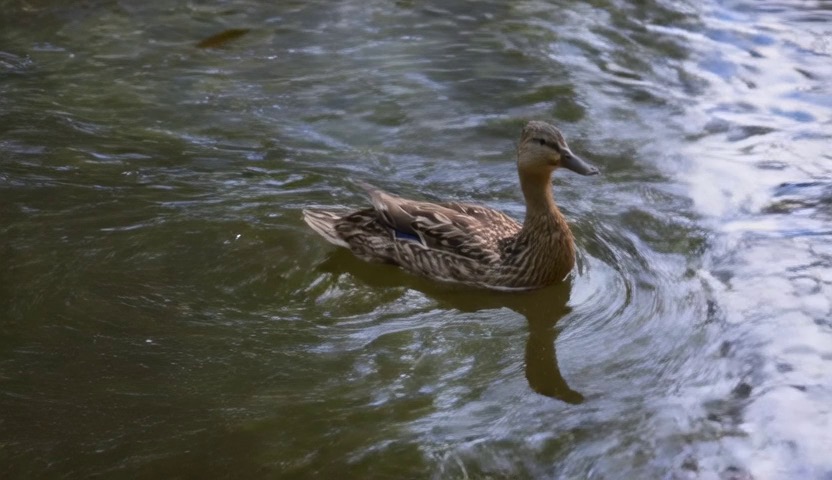} &
        \adjincludegraphics[clip,width=0.158\linewidth,trim={0 0 0 0}]{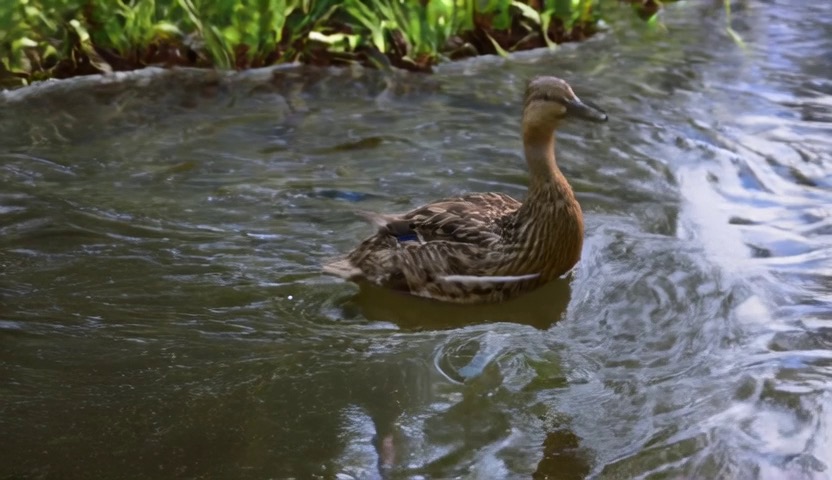} \\
        \addlinespace[0.5ex]
        & \multicolumn{3}{c}{\normalsize \textit{Arc Left}} &
        \multicolumn{3}{c}{\normalsize \textit{Translate Down}}  \\
    \end{tabular}
    \vspace{\abovefigcapmargin}
    \caption{\textbf{Additional qualitative results on the DAVIS dataset~\cite{pont20172017}.}}
    \vspace{\belowfigcapmargin}
    \label{fig:additional_davis}
\end{figure*}

\subsection{More Results and Applications}

We provide additional qualitative comparisons in~\cref{fig:additional_davis}. In the first example, previous methods produce unrealistic retakes, and TrajectoryCrafter renders the person’s legs inconsistently with the input video. In the second example, we observe that previous methods either produce inconsistent water textures or fail to achieve precise camera control. In contrast, our method consistently preserves both dynamic objects and backgrounds, and faithfully follows the target camera trajectories even under textureless backgrounds.

\begin{figure*}[p]
    \centering
    \includegraphics[width=\linewidth]{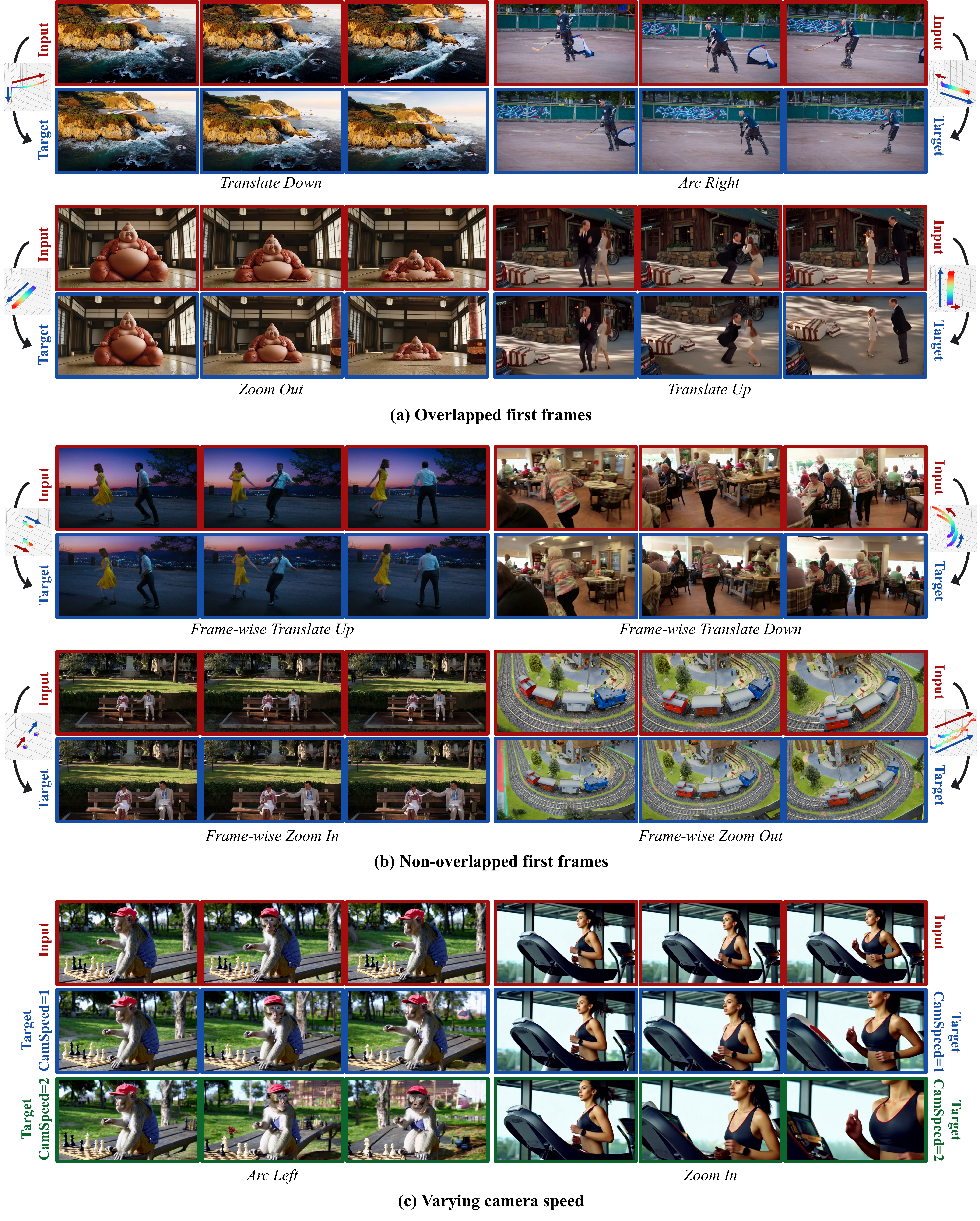}
    \caption{\textbf{Additional qualitative results and applications of ReDirector.}}
    \vspace{\belowfigcapmargin}
    \label{fig:additional_results}
\end{figure*}

We present additional video retakes generated by ReDirector in~\cref{fig:additional_results}-(a). The results confirm that our method generates high-quality, geometrically accurate, and temporally coherent video retakes across various trajectories and lengths. Although the training data consists of synchronized videos whose first frames are fully overlapped, we additionally include time-reversed versions of these videos during training. This enables our model to generate retakes with non-overlapping first frames, as shown in~\cref{fig:additional_results}-(b). Finally, since our method is well aligned with metrically scaled camera trajectories, we can stretch the target trajectories; \cref{fig:additional_results}-(c) shows video retakes when the camera trajectories are scaled by a factor of two.

\begin{figure}[t!]
    \centering
    \setlength\tabcolsep{0.5pt}
    
    \begin{tabular}{@{}c@{\,}cc@{}}
        \raisebox{0.02\linewidth}{\rotatebox{90}{\normalsize Input Video}} &
        \adjincludegraphics[clip,width=0.43\linewidth,trim={0 0 0 0}]{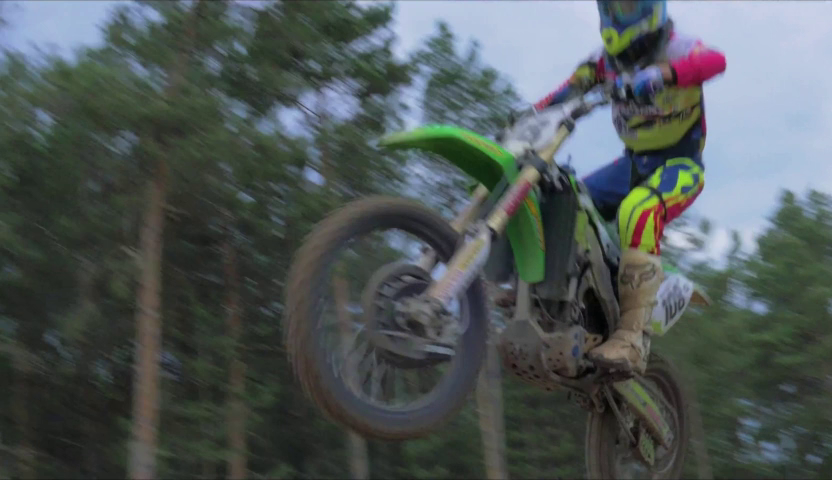} &
        \adjincludegraphics[clip,width=0.43\linewidth,trim={0 0 0 0}]{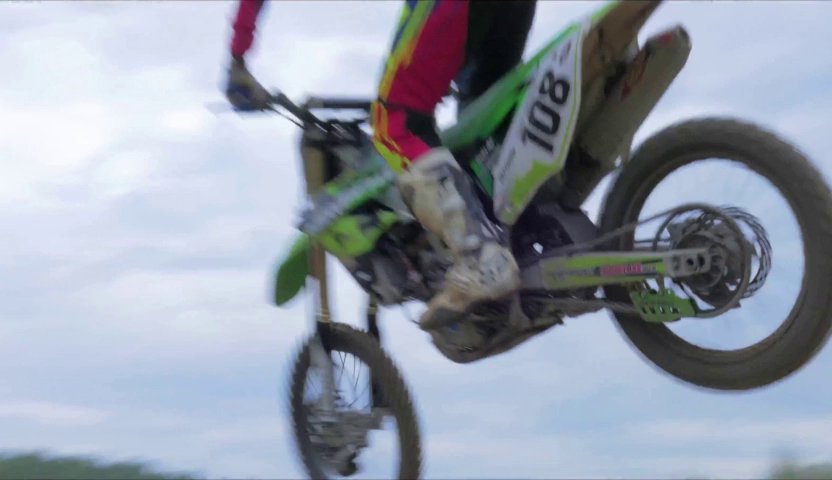} \\

        \raisebox{0.08\linewidth}{\rotatebox{90}{\normalsize Ours}} &
        \adjincludegraphics[clip,width=0.43\linewidth,trim={0 0 0 0}]{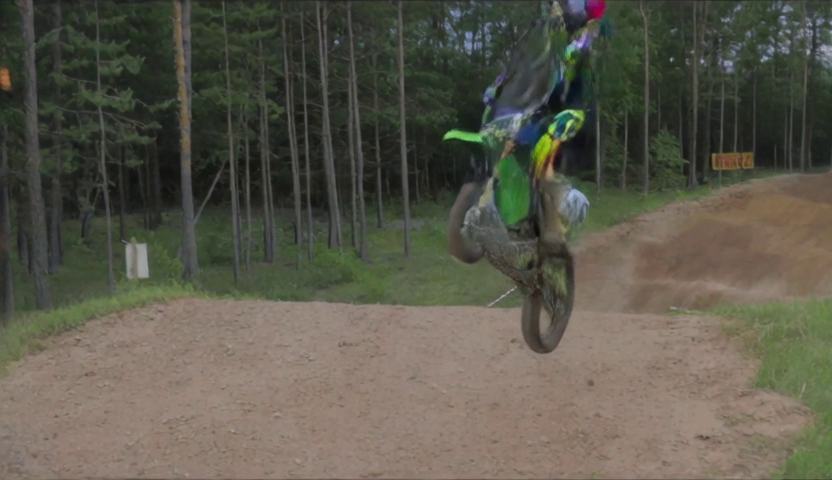} &
        \adjincludegraphics[clip,width=0.43\linewidth,trim={0 0 0 0}]{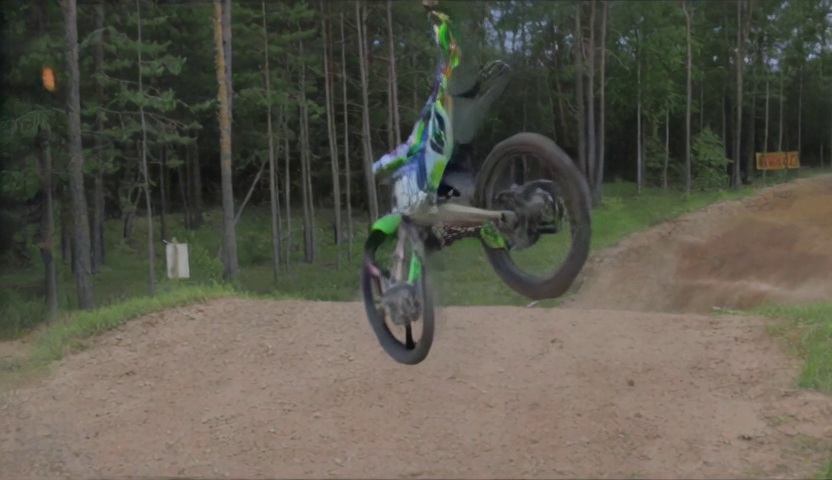} \\
    \end{tabular}
    \vspace{\abovefigcapmargin}
    \caption{\textbf{Failure case.}}
    \vspace{\belowfigcapmargin}
    \label{fig:failure}
\end{figure}

\section{Limitations and Discussions}

As shown in~\cref{fig:failure}, our method still struggles in challenging scenarios with extreme object and camera motion, particularly when large dynamic objects dominate the frame, making it difficult to reliably infer multi-view relationships. A promising future direction is to integrate our framework with next-frame video generation models~\cite{zhou2025learning, yu2025cam} in the context of world models, or to combine it with external geometry models~\cite{hu2025depthcrafter, chen2025video, karaev2024cotracker, xiao2024spatialtracker, nam2025generating} to further improve geometric robustness and fidelity.

\end{document}